\pdfoutput=1
\documentclass{article}

\PassOptionsToPackage{numbers}{natbib}
\usepackage[eandd, final]{neurips_2026}

\usepackage[utf8]{inputenc} % allow utf-8 input
\usepackage[T1]{fontenc}    % use 8-bit T1 fonts
\usepackage{hyperref}       % hyperlinks
\usepackage{url}            % simple URL typesetting
\usepackage{booktabs}       % professional-quality tables
\usepackage{amsfonts}       % blackboard math symbols
\usepackage{nicefrac}       % compact symbols for 1/2, etc.
\usepackage{microtype}      % microtypography
\usepackage{xcolor}         % colors
\usepackage{graphicx}
\usepackage{enumitem}
\usepackage{array}
\usepackage{siunitx}
\usepackage{tabularx}
\usepackage{colortbl}
\usepackage{pifont}
\usepackage{multirow, makecell}
\usepackage{wrapfig}
\definecolor{okgreen}{HTML}{1B8A3A}
\definecolor{nored}{HTML}{C0392B}
\newcommand{\yes}{\textcolor{okgreen}{\ding{51}}}
\newcommand{\no}{\textcolor{nored}{\ding{55}}}
\usepackage{caption}
\usepackage[most]{tcolorbox}
\usepackage[capitalise,noabbrev]{cleveref}
\definecolor{bandExt}{HTML}{E8EEF7}   % light blue  — External Memory band
\definecolor{bandInt}{HTML}{F4ECE4}   % light amber — Internal Memory band
\newcommand{\classband}[2]{\rowcolor{#1}\multicolumn{7}{@{}l@{}}{\,\textbf{#2}}}
\newcommand{\famcell}[1]{\textsc{\footnotesize #1}}
\newcolumntype{Y}{>{\raggedright\arraybackslash\hspace{0pt}}X}

\definecolor{bandQual}{HTML}{ECF2E8}   % light green — Quality band
\newcommand{\metricband}[2]{\rowcolor{#1}\multicolumn{4}{@{}l@{}}{\,\textbf{#2}}}

\definecolor{famPerf}{HTML}{1F3A68}
\definecolor{famEff}{HTML}{8B5A1C}
\definecolor{famQual}{HTML}{3D5A2E}

\usepackage{xspace}
\usepackage{fontawesome5}
\newcommand{\qwens}{\textsc{Qwen3-8B}\xspace}
\newcommand{\qwenl}{\textsc{Qwen3-32B-AWQ}\xspace}
\newcommand{\gemma}{\textsc{Gemma-4-26B-A4B-IT}\xspace}
\newcommand{\gemmashort}{\textsc{Gemma-4}\xspace}

\title{Harness the Memory: A Holistic Evaluation of Memory Substrates in Memory Agents}

\author{%
  \textbf{Wei-Chieh Huang$^{1}$ \hspace{0.25em} Weizhi Zhang$^{1,\dagger}$ \hspace{0.25em} Yuchen Wu$^{2}$ \hspace{0.25em} Yankai Chen$^{3,4,\dagger}$ \hspace{0.25em} Eric Hanchen Jiang$^{5}$} \\
  \textbf{Wooseong Yang$^{1}$ \hspace{0.25em} Yiwei Yang$^{2}$ \hspace{0.25em} Henry Peng Zou$^{1}$ \hspace{0.25em} Hanrong Zhang$^{1}$ \hspace{0.25em} Ying Nian Wu$^{5}$} \\
  \textbf{Haolun Wu$^{3}$ \enspace Kai-Wei Chang$^{5}$ \enspace Philip S.~Yu$^{1}$ \enspace Xue Liu$^{3,4}$ \enspace Aylin Caliskan$^{2,\dagger}$} \\[3pt]
  {\small $^{1}$University of Illinois Chicago \quad $^{2}$University of Washington \quad $^{3}$McGill University}\\[1pt]
  {\small $^{4}$MBZUAI \quad $^{5}$University of California, Los Angeles}\\[1pt]
  {\small $^{\dagger}$Corresponding authors}
}


\begin{document}

\setlength{\tabcolsep}{0pt}% author block: no column padding
\maketitle
\setlength{\tabcolsep}{6pt}

\begin{abstract}
% Memory is becoming a core infrastructure for long-horizon LLM agents, yet existing evaluations provide limited guidance on which memory substrate should be used under different operating regimes. We present a controlled evaluation of memory substrates for memory-augmented agents, covering external memories such as dense and sparse indices, text records, structural stores, hierarchical stores, and refinement-based memories, as well as internal memories such as parametric updates and activation caches. Across three backbone models and four benchmarks spanning user-centric question answering and agent-centric decision-making, we instrument performance, efficiency, and memory-quality metrics under a unified harness. Our results show that no single substrate dominates across regimes: broad retrieval benefits long-context factual QA, while excessive retrieval can harm sequential decision-making by diluting attention over action-critical context. These findings motivate substrate routing as a necessary component of adaptive agent memory systems and provide empirical guidance for designing efficient, reliable, and regime-aware long-term memory for LLM agents.

% add anonyomous code link. (this a benchmark paper, it's expected and important to have the code & data), can add actual content later though)

Memory is becoming core infrastructure for long-horizon LLM agents, yet existing evaluations offer limited guidance on which memory substrate, namely the underlying medium in which memory is represented and stored, should be used under different operating regimes. We introduce a unified suite of 26 performance and efficiency metrics, including auxiliary-LLM costs attributed to the write, read, and management stages, and apply it identically to 11 memory methods in a controlled harness that varies only the memory substrate. The methods cover dense and sparse indices, text records, structural stores, hierarchical stores, refinement-based memories, parametric updates, and activation-compatible context mechanisms, and we evaluate them with three backbone models on four benchmark suites spanning user-centric question answering and agent-centric decision-making. Our results show that no single substrate consistently dominates: broad retrieval benefits long-context factual QA, while excessive retrieval can harm sequential decision-making by shifting attention away from action-critical context. Scalability introduces a further routing axis, as substrates that perform well at moderate history lengths can become costly or brittle at longer horizons. These regime-dependent reversals make substrate selection a design axis. Choosing a substrate per observable operating regime improves over the best fixed substrate on our benchmarks, while query-level routing leaves most of the oracle headroom open. The findings provide empirical guidance for designing efficient, reliable, and regime-aware long-term memory for LLM agents. Code is available at \url{https://github.com/Jasonya/SubMem}.
\end{abstract}

\section{Introduction}
\label{sec:intro}
Foundation model agents are rapidly advancing from single-turn assistants into long-horizon settings that demand persistent memory: coding assistants resolving issues and developing code in real repositories~\cite{zhang2024autocoderover,miao2025recode}, web agents navigating realistic websites~\cite{zhouwebarena}, personal companions tracking evolving user preferences across hundreds of sessions~\cite{zhong2024memorybank,zhang2026jarvis}, and scientific discovery agents refining hypotheses over iterative cycles~\cite{Ren2025TowardsSIA}. Despite spanning vastly different domains, these agents share a common challenge: writing new experience into a persistent store, reading relevant knowledge at decision time, and managing an ever-growing memory under finite compute and storage budgets \citep{xiong2025memory}. No single memory design excels at all three along the axes of accuracy, efficiency, and cost~\cite{li2025memos,zhou2025mem1}: a vector index writes cost-effectively but struggles with multi-hop retrieval \citep{gutierrez2024hipporag}; a knowledge graph reasons relationally but requires orders of magnitude more LLM calls to construct (Appendix~\ref{app:excluded-substrates}); a KV cache integrates tightly with the model but sacrifices scalability \citep{li2024snapkv}. This diversity of trade-offs motivates the long-term goal of a \textbf{universal, adaptive memory system} that dynamically selects or composes substrates, namely the underlying media in which memory is represented and stored, depending on the query, the task, and the deployment constraints.
 
%Designing such a system requires understanding when each substrate is preferable, and we argue this question is best framed through \emph{operational hardness}. Long-horizon memory management faces four recurring sources of difficulty: accumulated content must be compressed without losing signal, outdated facts must be detected and superseded, new sessions must efficiently reconstruct working state, and stored knowledge must remain inspectable for validation. Each memory substrate is structurally aligned to a different subset of these dimensions, which explains why no substrate dominates universally and why aggregate benchmarks that average across hardness dimensions mask critical deployment-specific failures. A universal memory system must therefore route queries to the substrate best aligned with the dominant hardness of the moment.

Designing such a system requires understanding when each substrate is preferable and,
critically, under what operating regime. Recent work on agent harnesses, the scaffolding around a model that decomposes tasks, manages context across sessions, and evaluates outputs, has shown that this scaffolding is often the primary lever for agent performance beyond raw model capability~\cite{pan2026natural, bui2026building}. The memory substrate is the infrastructure layer of any such harness: its design constrains every other harness decision, from write cost and read latency to whether stored knowledge is inspectable and how the system degrades as histories grow. Yet the same substrate can be the quality leader on one task and a strictly dominated choice
on another, so a universal harness should not commit to a single substrate at design time \citep{zhang2025memevolve}. Selecting between substrates per regime then becomes a design axis of the harness. Producing the empirical signal for that selection requires \emph{evaluating the harness itself}: isolating the memory substrate as a controlled variable while holding the rest of the scaffolding fixed, across regimes that stress fundamentally different memory operations.

\begin{figure}[t]
  \centering
  \includegraphics[width=\textwidth]{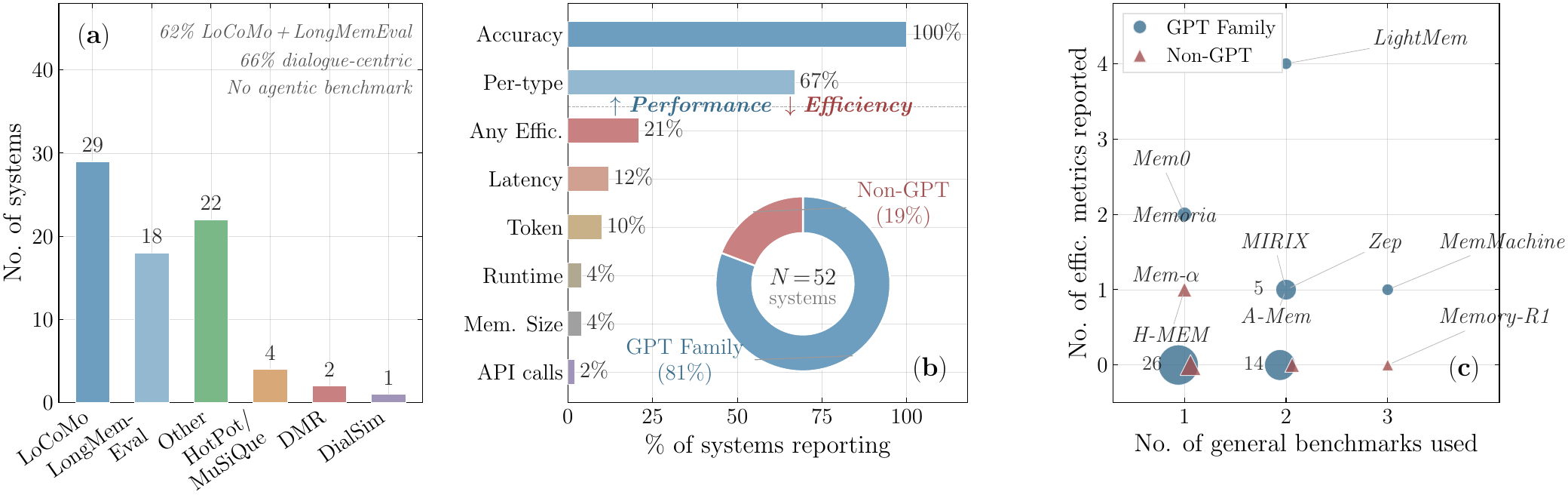}
  \vspace{-1em}
  \caption{%
    Evaluation landscape of 52 memory-augmented LLM systems (2023--2026).
    \textbf{(a)}~Benchmark adoption: 62\% of usage concentrates on
    LoCoMo and LongMemEval. \textbf{(b)}~Metric coverage: every system reports accuracy but
    only 21\% report any efficiency metric;
    % management cost and compression-style measurements are rarely reported in a standardized way
     81\% of systems use GPT-family backbones.
    \textbf{(c)}~Breadth and efficiency: half of
    systems use a single benchmark with no efficiency metrics; no
    system simultaneously achieves broad coverage and comprehensive
    efficiency reporting.%
  }
  \label{fig:landscape}
  \vspace{-1.8em}
\end{figure}

Yet current evaluations provide almost no such signal. As shown in Figure~\ref{fig:landscape}, a landscape analysis of 52 recent memory-augmented systems exposes three limitations that leave the path toward universal agent memory empirically ungrounded. First, evaluations are \emph{benchmark-concentrated}: 62\% of all benchmark and system pairs draw from just two dialogue-centric datasets, LoCoMo~\cite{maharana2024evaluating} and LongMemEval (LME)~\cite{wulongmemeval}, while agentic tasks remain almost unexamined (Figure~\ref{fig:landscape}\,a). Second, \emph{metric and backbone coverage is narrow}: every system reports accuracy, but only 21\% measure any efficiency dimension; management cost and compression ratio are unreported. Furthermore, 81\% of systems use the GPT-family as their primary backbone,
making it impossible to disentangle substrate effects from model-specific behavior (Figure~\ref{fig:landscape}\,b). Third, \emph{evaluation scope is shallow}: half of all systems test on a single benchmark with zero efficiency metrics, and no system simultaneously achieves broad coverage and comprehensive efficiency reporting (Figure~\ref{fig:landscape}\,c). The full per-system survey, inclusion criteria, and aggregate statistics are reported in
Appendix~\ref{app:landscape}. As a result, we know which system tops a given leaderboard in terms of accuracy, but not under what operating regime a different substrate would be preferable, precisely the selection signal that a regime-aware memory harness needs.

\begin{figure}[t]
  \centering
  \includegraphics[width=\textwidth]{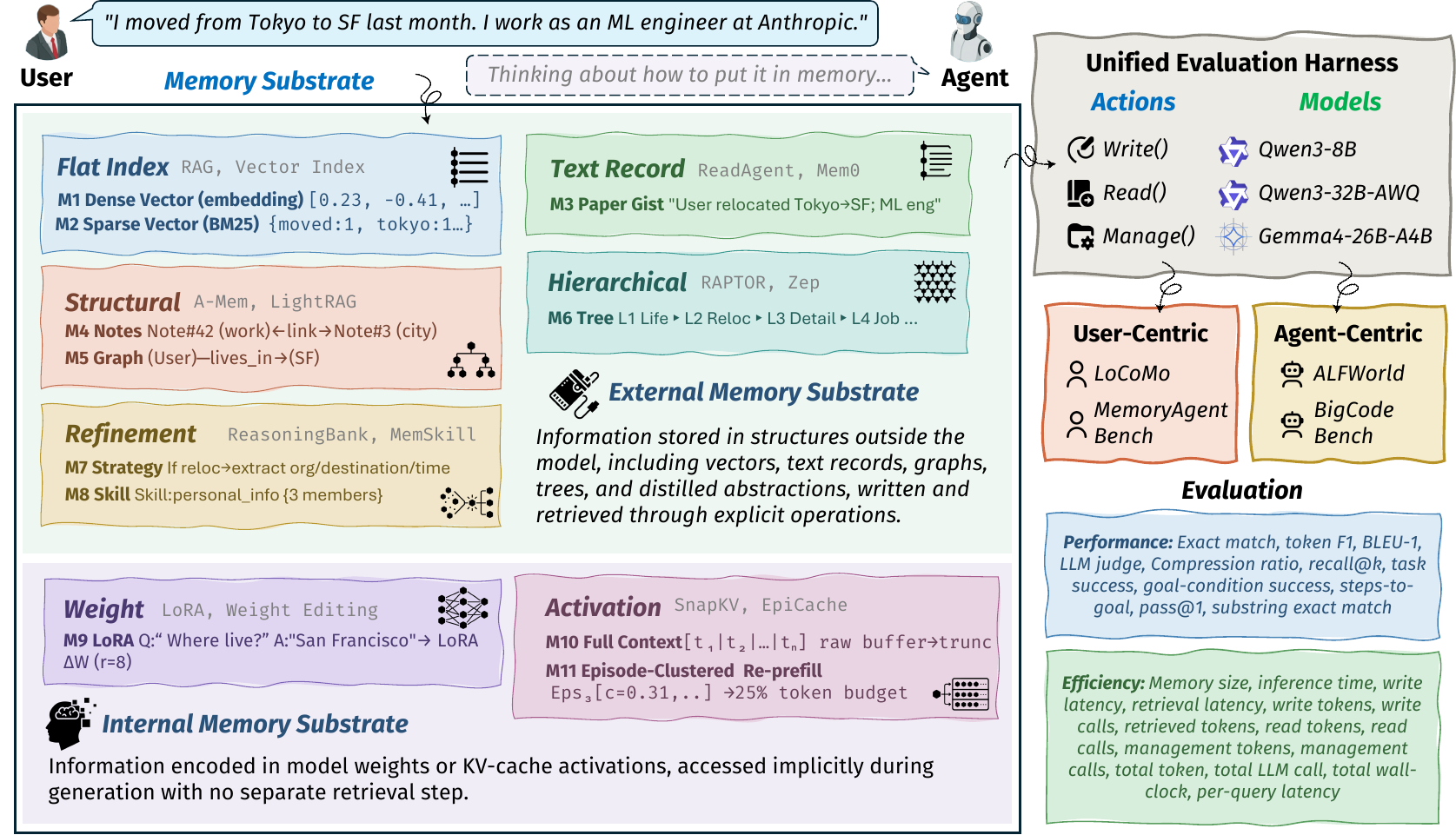}
  \caption{\textbf{Overview of the harness evaluation}. 11 memory methods (M1--M11) spanning seven substrate families are evaluated on
identical interaction histories across four benchmarks, instrumented
with 26 metrics in two families (performance and efficiency).}
  \label{fig:main_figure}
\end{figure}

We close this gap with a controlled harness evaluation that isolates the memory substrate, the storage-and-access layer of the harness defined in Appendix~\ref{app:substrate-def}, as the sole experimental variable. Adopting the taxonomy of \citep{huang2026rethinking}, we partition the design space into \textbf{external memory} (information in data structures outside the model) and \textbf{internal memory} (information encoded in model weights or activations), and implement 11 methods spanning all seven substrate families, as shown in Figure~\ref{fig:main_figure}. All substrates are evaluated on identical interaction histories under three backbones and four benchmarks covering both user-centric question answering and agent-centric decision-making.

The harness evaluation yields three findings. First, the optimal substrate \emph{reverses} between regimes. On \qwenl, structural graphs lead dialogue QA but are Pareto-dominated on agentic tasks, while refinement memories lead embodied planning but trail on QA (Section~\ref{sec:perf-latency}). No substrate leads on quality in both regimes, since the quality end of the performance--latency frontier moves from M5 on QA to M7 on ALFWorld and M2 on BigCodeBench-Hard. Second, retrieving more entries helps QA but hurts agentic decisions, and an attention probe on one backbone suggests why: as retrieval depth grows, the model's attention shifts away from the task context toward the retrieved block, which is exactly where the answer lives in QA but not in agentic tasks, where the policy needs to attend to the current observation instead (Section~\ref{sec:retrieval-scaling}). Third, a scalability study shows that the substrates whose quality grows most with history length, led by our harness approximation of skill bundles (M8), also pay the steepest latency growth, while only the flat indices stay cheap at long horizons (Section~\ref{sec:scalability}). Together, these findings yield a regime-conditioned design rule, \emph{trade read breadth for write depth} on agentic tasks (retrieve fewer entries per query and invest more in distilling and structuring memory at write time) while retrieving broadly on long-context QA, and a broader implication: no single substrate serves every regime, so substrate selection is a design axis of universal agent memory, whether it is realized as a \textbf{multi-substrate or multi-agent system}. Choosing a substrate per observable regime beats the best fixed substrate on our benchmarks, while finer query-level routing remains largely open (Appendix~\ref{app:routing}). %Specifically, our contributions are as follows:
Our contributions are summarized as follows:

\begin{itemize}[leftmargin=*,itemsep=2pt]
  \item \textbf{A unified metric suite for agent memory.} A suite of 26 performance and efficiency metrics applied identically to all 11 methods. It covers memory quality (compression ratio and the MAB LRU, TTL, and CR capabilities) and, to our knowledge, provides the first cost accounting that attributes auxiliary-LLM tokens and calls to the write, read, and management stages across substrate families.

  \item \textbf{A controlled, cross-regime harness evaluation.} A comparison that isolates the memory substrate across 11 methods and seven substrate families of the external and internal taxonomy, three backbones, and four benchmarks spanning user-centric and agent-centric regimes. It maps each method onto a unified performance--latency landscape and identifies which substrate families are dominated under which operating regime.

  \item \textbf{A diagnostic study of retrieval scaling and context scalability.} A
  retrieval-depth sweep paired with attention probing that diagnoses how retrieval breadth reshapes attention allocation across regimes, together with a scalability analysis that characterizes how each substrate family scales with input growth,
  yielding actionable design rules for regime-aware memory.
\end{itemize}

% \begin{enumerate}[leftmargin=*, nosep]

% \item \textbf{Controlled substrate-level comparison.} We present the first evaluation that isolates the effect of the memory substrate by holding fixed the base model, benchmark, evaluation prompt, and judge model across all 15 configurations. 

% \item \textbf{Multi-dimensional evaluation taxonomy.} We define and measure more than 20 metrics spanning task performance, computational efficiency, and memory quality. This fills critical blind spots identified in our landscape analysis: write cost, management cost, compression ratio, recency bias, and contradiction resolution are quantified for the first time across substrate families, revealing trade-off dimensions that accuracy-only evaluations cannot capture.

% \item \textbf{Actionable substrate characterization.} Our analysis yields concrete, finding-driven guidance for memory system design: write cost spans three orders of magnitude across substrates, faithfulness and accuracy are inversely correlated in fact-extraction substrates, and internal-memory substrates severely underperform external ones on memory-centric QA. These findings directly inform which substrate to select under given task, query-type, and deployment-cost constraints.

% \end{enumerate}

\section{Memory Substrates and Configuration}
\label{sec:substrates}

\definecolor{flatcolor}{HTML}{4A90D9}      % Flat — blue (M1, M2)
\definecolor{textcolor}{HTML}{E89B3C}      % Text — orange/tan (M3)
\definecolor{structcolor}{HTML}{C04B4B}    % Structural — red (M4, M5)
\definecolor{hiercolor}{HTML}{8B5CB8}      % Hierarchical — purple (M6)
\definecolor{refinecolor}{HTML}{D14F8C}    % Refinement — pink/magenta (M7, M8)
\definecolor{weightcolor}{HTML}{6A6A6A}    % Weight — gray (M9)
\definecolor{actcolor}{HTML}{4FA885}       % Activation — green (M10, M11)
% Convenience macros
\newcommand{\famFlat}{\textcolor{flatcolor}{\textbf{Flat}}}
\newcommand{\famText}{\textcolor{textcolor}{\textbf{Text}}}
\newcommand{\famStruct}{\textcolor{structcolor}{\textbf{Structural}}}
\newcommand{\famHier}{\textcolor{hiercolor}{\textbf{Hierarchical}}}
\newcommand{\famRefine}{\textcolor{refinecolor}{\textbf{Refinement}}}
\newcommand{\famWeight}{\textcolor{weightcolor}{\textbf{Weight}}}
\newcommand{\famAct}{\textcolor{actcolor}{\textbf{Activation}}}
% Family icons (fontawesome5) -- colored to match each family
\newcommand{\icoFlat}{\textcolor{flatcolor}{\faBars}}
\newcommand{\icoText}{\textcolor{textcolor}{\faFile}}
\newcommand{\icoStruct}{\textcolor{structcolor}{\faLink}}
\newcommand{\icoHier}{\textcolor{hiercolor}{\faSitemap}}
\newcommand{\icoRefine}{\textcolor{refinecolor}{\faFilter}}
\newcommand{\icoWeight}{\textcolor{weightcolor}{\faBrain}}
\newcommand{\icoAct}{\textcolor{actcolor}{\faBolt}}
% Family labels for table cells (icon + colored abbreviation)
\newcommand{\tabFlat}{\icoFlat~\textcolor{flatcolor}{Flat}}
\newcommand{\tabText}{\icoText~\textcolor{textcolor}{Text}}
\newcommand{\tabStruct}{\icoStruct~\textcolor{structcolor}{Struct}}
\newcommand{\tabHier}{\icoHier~\textcolor{hiercolor}{Hier}}
\newcommand{\tabRefine}{\icoRefine~\textcolor{refinecolor}{Refine}}
\newcommand{\tabWeight}{\icoWeight~\textcolor{weightcolor}{Weight}}
\newcommand{\tabAct}{\icoAct~\textcolor{actcolor}{Act}}

Memory system configurations are highly complex, with many tunable parameters and modular
design choices, so rather than asking which method is inherently superior, we examine which
configurations yield specific benefits and trade-offs under different task settings and
operating regimes, the empirical prerequisite for any universal memory system that
selects substrates dynamically. 

\paragraph{Substrate overview.}
We study eleven memory substrates arranged into seven families (\cref{tab:substrate} in Appendix~\ref{app:substrate-details}). On the external side, the eight methods span a structure spectrum.
\icoFlat~\textcolor{flatcolor}{\textbf{Flat Index}}: M1 Dense Vector and M2 Sparse
Vector~\citep{xiongapproximate, robertson2009probabilistic} require no LLM calls and establish
the retrieval floor. \icoText~\textcolor{textcolor}{\textbf{Text Record}}: M3 Gist
Index~\citep{leehuman} adds write-side gisting and read-side page selection.
\icoStruct~\textcolor{structcolor}{\textbf{Structural}} introduces relational links: M4
Evolving Notes~\citep{xu2025mem} writes interlinked notes with evolution-triggered
rewrites, and M5 Dual-Level Graph~\citep{guo2024lightrag} builds a dual-level
entity-relation graph. \icoHier~\textcolor{hiercolor}{\textbf{Hierarchical}}: M6 Hierarchical
Tree~\citep{sarthi2024raptor, sun2026h} recursively clusters and summarizes memory into a
multi-level hierarchy retrieved via collapsed-tree dense search.
\icoRefine~\textcolor{refinecolor}{\textbf{Refinement}} distills experience rather than accumulating
it: M7 Distilled Strategies~\citep{ouyang2025reasoningbank} judges trajectories and
extracts reusable strategies, and M8 Skill Bundles~\citep{zhang2026memskill} clusters
experiences into skill bundles. On the internal side, the
\icoWeight~\textcolor{weightcolor}{\textbf{Weight}} family, M9 Adapter Tuning~\citep{zhang2023lora},
bakes knowledge into weights via fine-tuning, while the
\icoAct~\textcolor{actcolor}{\textbf{Activation}} family includes M10 Full Context and M11 Episode-Clustered Re-prefill~\citep{kim2025epicache}, which
clusters turns into episodes and re-prefills selected episodes at read time. We report Zep~\citep{rasmussen2025zep}, Mem0~\citep{chhikara2025mem0}, and
MemGPT~\citep{packer2023memgpt} in an auxiliary-cost analysis (Appendix~\ref{app:excluded-substrates}) rather than the main controlled comparison because their defining mechanisms are management and controller policies outside the single-variable frame, and their production pipelines introduce substantially different auxiliary-LLM budgets.

\paragraph{Scope of the comparison.}
We define a substrate by four properties, its representation and its write, read, and management operations (Appendix~\ref{app:substrate-def}). These mechanisms act at different stages of one pipeline and can be composed, so each comparison contrasts configured substrates under a shared harness rather than mutually exclusive replacements. Shared settings and per-substrate hyperparameters follow a fixed calibration protocol (Appendix~\ref{app:calibration}). Two methods are harness-compatible approximations of their reference systems. M8 replaces MemSkill's PPO-trained controller with a zero-shot LLM over all skill bundles, and M11 replaces EpiCache's in-place KV-cache edits with utterance-level re-prefill. Results reported for M8 and M11 refer to these approximations.

\section{Experiment Design}
\label{sec:setup}

Our goal is to isolate the implemented memory substrate family under a shared harness, backbone, prompting, and auxiliary-LLM setting. We ask two questions: (1)~how much of an agent's behavior, in terms of task performance and efficiency, is attributable to the substrate alone, and (2)~whether that attribution is stable across
operating regimes. We test every substrate on both user-centric benchmarks, where the agent retrieves facts from long conversational histories, and agent-centric benchmarks, where the agent must execute precise action sequences. Three configurations are excluded by design: M7 only applies to trajectory-bearing agent-centric tasks (its strategy-distillation pipeline has no analog on dialogue inputs), M9 is incompatible with the A4B MoE routing in
\gemma, and M10 (full context) is omitted from both agent-centric benchmarks (ALFWorld and BigCodeBench-Hard) because the cumulative context exceeds every tested context window. %As our results show, the optimal substrate and retrieval

\paragraph{Benchmarks.}
We select four benchmarks that collectively span the user-centric and agent-centric divide. On the user-centric side, \textbf{LoCoMo}~\citep{maharana2024evaluating} provides ten multi-session dialogues with $1{,}986$ questions across five categories, and \textbf{MemoryAgentBench (MAB)}~\citep{hu2025evaluating} factors long-context memory into four capabilities, Accurate Retrieval (AR, instantiated on MemoryAgentBench's LongMemEval-S$^\ast$ subset~\citep{wulongmemeval}), Long-Range Understanding (LRU), Test-Time Learning (TTL), and Conflict Resolution (CR); we additionally use CR as our scalability probe by sweeping content length. In the result tables we surface AR under its LongMemEval-S (LME-S) label as a separate column alongside the LRU, TTL, and CR capabilities. Together these benchmarks stress recall-oriented capabilities, the regime where broad retrieval helps. On the agentic side, \textbf{ALFWorld}~\citep{shridhar2020alfworld} provides $134$ valid-unseen embodied-planning tasks evaluated in a cross-episode regime (recovery from a shared pre-built bank of successes and failures), and \textbf{BigCodeBench-Hard} (BCB-Hard)~\citep{zhuobigcodebench} provides $148$ code tasks graded with retrieval over a cross-task pool of successes and failures. These benchmarks stress the opposite regime, where retrieval noise is toxic and the substrate must distill experience into compact, high-precision representations. Full benchmark setups, capability definitions, and pool construction details are in
Appendix~\ref{app:benchmarks}, which also maps each benchmark onto the stages of the memory lifecycle it exercises (Appendix~\ref{app:lifecycle}).

%The user-centric / agent-centric divide is the primary axis of our harness: as we show in Section~\ref{sec:results}, the optimal substrate and retrieval strategy \emph{reverse} across this boundary.

\paragraph{Models, inference, and metrics.}
The backbone models are \qwens, \qwenl~\citep{yang2025qwen3}, and \gemma~\citep{gemma2026gemma4}, tested on all four benchmarks using vLLM~\citep{kwon2023efficient} on $4{\times}$H200 GPUs, the single configuration behind every latency and wall-clock number we report. For consistency, the auxiliary LLM is fixed to \textsc{gpt-4o-mini}~\citep{openai2024gpt4omini} across all methods that require one for memory writing, management, or input organization. LLM-as-a-judge evaluation also uses \textsc{gpt-4o-mini} with fixed judging templates, and three further judges agree substantially with its labels (Appendix~\ref{app:judge}). We instrument every run with the 26 metrics defined in Table~\ref{tab:metric-taxonomy} in Appendix~\ref{app:metrics}, organized into task-performance metrics and efficiency metrics, including benchmark-native primary metrics such as Pass@1 on BCB-Hard, task success on
ALFWorld, and SubEM on MAB, as well as storage, latency, token, call, total-time, and cost measurements. %All metrics are computed in a single pass, ensuring that efficiency and quality measurements remain synchronized with the corresponding task scores.

\paragraph{Ablations.}
We isolate two design axes while holding the substrate implementation, backbone, and prompt template fixed. First, a top-$k$ retrieval-breadth sweep on M1, M2, M4, and M6, plus M7 on ALFWorld: $k \in \{1,2,5,10,20\}$ on LoCoMo and $k \in \{1,2,3,4,5\}$ on ALFWorld, with a no-memory baseline that receives the same template with an empty retrieved block, so the curves attribute quality to retrieval rather than scaffolding. Second, a context-window stress on MAB CR under three conversation-length regimes (6K, 32K, 262K tokens) with the question pool, gold answers, and judge held constant. Together these ablations separate substrates whose advantage comes from retrieval breadth from those driven by recency-aware update semantics.

\section{Results}
\label{sec:result}

\newcommand{\logoQwen}{\raisebox{-2.5pt}{\includegraphics[height=12pt]{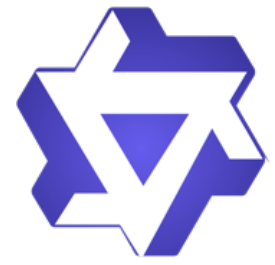}}}
\newcommand{\logoGoogle}{\raisebox{-2.5pt}{\includegraphics[height=12pt]{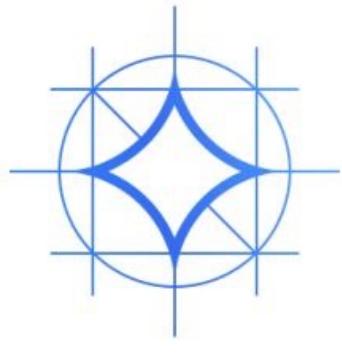}}}

\definecolor{bestcell}{HTML}{D4EFCB}     % soft green  — best on metric
\definecolor{worstcell}{HTML}{F8D6D6}    % soft red    — worst on metric
\definecolor{headerslate}{HTML}{2E3440}  % header bar

% ------------------------------------------------------------------

% ------------------------------------------------------------

We organize the results along the user-centric and agent-centric divide that defines our
harness. Table~\ref{tab:locomo-mab-main} reports user-centric results on LoCoMo, LME-S, and MAB, where the substrate must surface relevant facts from accumulated dialogue history;
Table~\ref{tab:alfworld-bcb-main} reports agent-centric results on ALFWorld and
BigCodeBench-Hard, where the substrate must inform sequential decisions without diluting
attention over action-critical context. Full per-metric breakdowns across all 26 metrics
are in Appendix~\ref{app:full_table}. The memory-quality metrics that the suite computes by one rule for every method already separate the substrates before the regime divide. On LME-S, the compression ratio $P_5$ ranges from 1.00 for the verbatim flat stores (M1, M2) through 2.80 for M6 to 12.50 for M8 (Appendix~\ref{app:lme-mab-deep-analysis}), and the MAB columns of Table~\ref{tab:locomo-mab-main} rank the substrates differently on LRU, TTL, and CR.

\begin{table*}[t]
\centering
\renewcommand{\arraystretch}{0.95}
\setlength{\tabcolsep}{6pt}
\caption{Memory substrates on \textbf{LoCoMo}, \textbf{LME-S}, and \textbf{MAB} (LRU, TTL, CR). $P_4$ $\uparrow$: LLM-judge score; $E_{15}$ $\downarrow$: per-query latency (s). Within each model band, green and red mark the per-column best and worst; \textbf{bold} marks the per-model P4 winner on each benchmark.
}
\vspace{-0.5em}
\label{tab:locomo-mab-main}
\resizebox{\textwidth}{!}{%
\begin{tabular}{@{}c l l *{2}{c} *{2}{c} *{2}{c} *{2}{c} *{2}{c}@{}}
\toprule
 & & & \multicolumn{2}{c}{\textbf{LoCoMo}}
   & \multicolumn{2}{c}{\textbf{LME-S}}
   & \multicolumn{2}{c}{\textbf{MAB LRU}}
   & \multicolumn{2}{c}{\textbf{MAB TTL}}
   & \multicolumn{2}{c}{\textbf{MAB CR}} \\
\cmidrule(lr){4-5} \cmidrule(lr){6-7} \cmidrule(lr){8-9} \cmidrule(lr){10-11} \cmidrule(lr){12-13}
\textbf{Model} & \textbf{Sub.} & \textbf{Family}
 & \boldmath$P_4$\unboldmath\,$\uparrow$ & \boldmath$E_{15}$\unboldmath\,$\downarrow$
 & \boldmath$P_4$\unboldmath\,$\uparrow$ & \boldmath$E_{15}$\unboldmath\,$\downarrow$
 & \boldmath$P_4$\unboldmath\,$\uparrow$ & \boldmath$E_{15}$\unboldmath\,$\downarrow$
 & \boldmath$P_4$\unboldmath\,$\uparrow$ & \boldmath$E_{15}$\unboldmath\,$\downarrow$
 & \boldmath$P_4$\unboldmath\,$\uparrow$ & \boldmath$E_{15}$\unboldmath\,$\downarrow$ \\
\midrule
%% ======== Qwen3-8B ========
\multirow{12}{*}{\rotatebox[origin=c]{90}
{\scalebox{0.85}
{\logoQwen\hspace{3pt}\qwens}}}
 & \multicolumn{12}{@{}l}{\hspace{-2pt}\textit{\footnotesize External}} \\
 & M1  & \tabFlat & 0.561 & 0.77 & 0.478 & 7.30 & 0.578 & 9.94 & 0.560 & 0.92 & 0.270 & 1.65 \\
 & M2  & \tabFlat & 0.470 & \cellcolor{bestcell}0.33 & 0.480 & \cellcolor{bestcell}1.75 & 0.620 & \cellcolor{bestcell}3.23 & 0.470 & \cellcolor{bestcell}0.33 & 0.270 & \cellcolor{bestcell}1.39 \\
 & M3  & \tabText & 0.562 & 4.18 & 0.495 & 55.58 & 0.563 & 146.65 & \cellcolor{bestcell}\textbf{0.640} & \cellcolor{worstcell}27.15 & \cellcolor{worstcell}0.240 & 2.30 \\
 & M4  & \tabStruct & 0.545 & 6.36 & 0.430 & \cellcolor{worstcell}225.67 & 0.493 & 274.06 & 0.480 & 12.65 & 0.270 & 1.50 \\
 & M5  & \tabStruct & \cellcolor{bestcell}\textbf{0.648} & \cellcolor{worstcell}27.84 & \cellcolor{bestcell}\textbf{0.537} & 187.21 & 0.578 & \cellcolor{worstcell}313.18 & 0.540 & 23.70 & \cellcolor{bestcell}\textbf{0.440} & \cellcolor{worstcell}6.01 \\
 & M6  & \tabHier & 0.556 & 1.39 & 0.507 & 19.96 & 0.493 & 59.98 & 0.290 & 3.24 & 0.290 & 3.30 \\
 & M8  & \tabRefine & 0.509 & 3.03 & 0.490 & 102.85 & 0.592 & 197.09 & 0.560 & 1.02 & 0.340 & 2.07 \\
 & \multicolumn{12}{@{}l}{\hspace{-2pt}\textit{\footnotesize Internal}} \\
 & M9  & \tabWeight & 0.379 & 1.76 & 0.250 & 51.50 & \cellcolor{worstcell}0.300 & 85.00 & \cellcolor{worstcell}0.220 & 8.50 & 0.260 & 1.95 \\
 & M10 & \tabAct & 0.589 & 4.55 & 0.120 & 6.61 & \cellcolor{bestcell}\textbf{0.676} & 6.58 & 0.480 & 2.28 & 0.270 & 1.64 \\
 & M11 & \tabAct & \cellcolor{worstcell}0.307 & 2.03 & \cellcolor{worstcell}0.090 & 14.61 & 0.634 & 21.30 & 0.560 & 0.56 & \cellcolor{worstcell}0.240 & 1.76 \\
\midrule
%% ======== Qwen3-32B-AWQ ========
\multirow{12}{*}{\rotatebox[origin=c]{90}
{\scalebox{0.85}
{\logoQwen\hspace{3pt}\qwenl}}}
 & \multicolumn{12}{@{}l}{\hspace{-2pt}\textit{\footnotesize External}} \\
 & M1  & \tabFlat & 0.582 & 2.59 & 0.548 & 12.00 & 0.676 & 32.39 & 0.810 & 1.81 & 0.300 & 0.85 \\
 & M2  & \tabFlat & 0.487 & \cellcolor{bestcell}2.37 & 0.497 & \cellcolor{bestcell}7.31 & 0.662 & \cellcolor{bestcell}23.14 & 0.780 & \cellcolor{bestcell}1.20 & 0.280 & \cellcolor{bestcell}0.50 \\
 & M3  & \tabText & 0.586 & 6.47 & 0.560 & 67.42 & \cellcolor{bestcell}\textbf{0.775} & \cellcolor{worstcell}477.97 & \cellcolor{bestcell}\textbf{0.850} & \cellcolor{worstcell}94.29 & 0.270 & 1.19 \\
 & M4  & \tabStruct & 0.452 & 8.70 & 0.490 & \cellcolor{worstcell}234.00 & 0.535 & 278.93 & 0.710 & 17.18 & 0.290 & 2.35 \\
 & M5  & \tabStruct & \cellcolor{bestcell}\textbf{0.683} & \cellcolor{worstcell}29.22 & \cellcolor{bestcell}\textbf{0.627} & 194.08 & 0.578 & 333.95 & 0.630 & 26.02 & 0.310 & 3.00 \\
 & M6  & \tabHier & 0.573 & 5.20 & 0.593 & 20.27 & 0.577 & 73.16 & 0.330 & 3.77 & 0.270 & 1.13 \\
 & M8  & \tabRefine & 0.533 & 5.53 & 0.533 & 106.78 & 0.592 & 228.66 & 0.810 & 1.93 & \cellcolor{bestcell}\textbf{0.320} & 1.08 \\
 & \multicolumn{12}{@{}l}{\hspace{-2pt}\textit{\footnotesize Internal}} \\
 & M9  & \tabWeight & 0.399 & 4.73 & 0.290 & 171.70 & \cellcolor{worstcell}0.422 & 253.63 & \cellcolor{worstcell}0.250 & 22.15 & \cellcolor{worstcell}0.170 & 3.27 \\
 & M10 & \tabAct & 0.669 & 16.25 & 0.190 & 22.98 & 0.620 & 27.43 & 0.760 & 2.43 & 0.270 & 0.51 \\
 & M11 & \tabAct & \cellcolor{worstcell}0.274 & 4.17 & \cellcolor{worstcell}0.113 & 52.39 & 0.620 & 25.00 & 0.810 & 1.45 & \cellcolor{bestcell}\textbf{0.320} & \cellcolor{worstcell}3.41 \\
\midrule
%% ======== Gemma-4-26B ========
\multirow{11}{*}{\rotatebox[origin=c]{90}
{\scalebox{0.85}
{\logoGoogle\hspace{3pt}\gemma}}}
 & \multicolumn{12}{@{}l}{\hspace{-2pt}\textit{\footnotesize External}} \\
 & M1  & \tabFlat & 0.574 & 0.93 & 0.515 & 8.49 & 0.676 & 16.50 & \cellcolor{bestcell}\textbf{0.770} & 14.50 & 0.460 & 4.09 \\
 & M2  & \tabFlat & 0.555 & \cellcolor{bestcell}0.49 & 0.513 & \cellcolor{bestcell}3.37 & 0.662 & \cellcolor{bestcell}9.18 & 0.730 & \cellcolor{bestcell}0.20 & 0.460 & 3.88 \\
 & M3  & \tabText & 0.599 & 4.81 & 0.530 & 31.78 & \cellcolor{bestcell}\textbf{0.720} & 238.78 & 0.750 & \cellcolor{worstcell}36.21 & 0.430 & 3.67 \\
 & M4  & \tabStruct & 0.476 & 6.61 & 0.523 & \cellcolor{worstcell}230.39 & 0.535 & 267.00 & 0.720 & 16.12 & 0.460 & 3.97 \\
 & M5  & \tabStruct & \cellcolor{bestcell}\textbf{0.719} & \cellcolor{worstcell}25.95 & \cellcolor{bestcell}\textbf{0.667} & 190.93 & 0.451 & \cellcolor{worstcell}332.94 & 0.720 & 25.79 & \cellcolor{bestcell}\textbf{0.550} & \cellcolor{worstcell}8.23 \\
 & M6  & \tabHier & 0.592 & 1.42 & 0.593 & 17.19 & 0.507 & 62.76 & \cellcolor{worstcell}0.250 & 22.00 & 0.400 & 6.50 \\
 & M8  & \tabRefine & 0.554 & 3.22 & 0.537 & 103.80 & 0.535 & 228.16 & \cellcolor{bestcell}\textbf{0.770} & 1.68 & 0.500 & 2.82 \\
 & \multicolumn{12}{@{}l}{\hspace{-2pt}\textit{\footnotesize Internal}} \\
 & M10 & \tabAct & 0.688 & 5.17 & \cellcolor{worstcell}0.110 & 6.91 & 0.577 & 10.38 & 0.460 & 1.36 & 0.450 & 4.24 \\
 & M11 & \tabAct & \cellcolor{worstcell}0.330 & 2.16 & 0.115 & 18.50 & \cellcolor{worstcell}0.318 & 76.00 & 0.480 & 1.50 & \cellcolor{worstcell}0.180 & \cellcolor{bestcell}1.48 \\
\bottomrule
\end{tabular}%
}
\end{table*}

\subsection{User-centric Benchmark}
\label{sec:user-centric}

\textbf{No single substrate wins every capability, ruling out single-substrate universal memory by
design.} 
Table~\ref{tab:locomo-mab-main} shows that the best substrate changes with the capability being tested, and this pattern is broadly visible across backbones despite model-specific interactions. On LoCoMo and LME-S, M5 performs best across model bands, suggesting that combining entity-level graph traversal with chunk-level vector search is useful for fact retrieval over long dialogue histories. On MAB LRU, less lossy access mechanisms often perform better: M10 leads \qwens, while M3 leads \qwenl. On MAB TTL, M3 leads both Qwen bands and M1 and M8 share the \gemmashort{} lead, whereas the M9 adapter-tuning implementation performs poorly because newly acquired facts are not directly inspectable or selectively queried. On MAB CR, M5 leads two of three models, consistent with the value of explicit update semantics for recency-sensitive conflicts. Cost separates otherwise competitive methods: M2 is consistently cheap, while M5 can be 10--100$\times$ slower and is only worthwhile when its structural mechanism matches the task bottleneck.

\begin{table*}[t]
\centering
\renewcommand{\arraystretch}{0.95}
\setlength{\tabcolsep}{4pt}
\caption{Memory substrates evaluated on \textbf{ALFWorld-unseen} and \textbf{BigCodeBench-Hard}. Within each model band, green marks the best substrate value on a metric, red marks the worst. $P_{\mathrm{avg}}$ is the mean per-task partial-success score. \textbf{Bold} marks the per-model performance winner (TSR for ALFWorld, Pass@1 for BCB). M10 is omitted from both agent-centric benchmarks because the cumulative context exceeds the tested context windows; M9 is omitted from the \gemmashort{} band because it is incompatible with the A4B MoE routing.}
\label{tab:alfworld-bcb-main}
\resizebox{\textwidth}{!}{%
\begin{tabular}{@{}c l l *{5}{c} *{3}{c}@{}}
\toprule
 & & & \multicolumn{5}{c}{\textbf{ALFWorld-unseen}} & \multicolumn{3}{c}{\textbf{BigCodeBench-Hard}} \\
\cmidrule(lr){4-8} \cmidrule(lr){9-11}
\textbf{Model} & \textbf{Sub.} & \textbf{Family}
 & \textbf{TSR}\,$\uparrow$ & \boldmath$\mathrm{Steps}\!\mid\!\mathcal{S}$\unboldmath\,$\downarrow$ & \boldmath$P_{\mathrm{avg}}$\unboldmath\,$\uparrow$ & \boldmath$E_{2}^{\mathrm{p90}}$\unboldmath\,$\downarrow$ & \boldmath$E_{15}$\unboldmath\,$\downarrow$
 & \textbf{Pass@1}\,$\uparrow$ & \boldmath$E_{2}^{\mathrm{avg}}$\unboldmath\,$\downarrow$ & \boldmath$E_{15}$\unboldmath\,$\downarrow$ \\
\midrule
%% ======== Qwen3-8B ========
\multirow{11}{*}{\rotatebox[origin=c]{90}
{\scalebox{0.85}
{\logoQwen\hspace{3pt}\qwens}}}
 & - & NoMem  & 5.7 & 12.2 & 13.5 & 13.2 & 488 & 8.1 & 3.8 & 4 \\
 & M1  & \tabFlat & 5.2 & \cellcolor{bestcell}5.9 & 6.6 & 16.1 & 672 & 9.5 & 5.0 & 5 \\
 & M2  & \tabFlat & 6.7 & 6.0 & 9.4 & 15.7 & 647 & 10.1 & 4.9 & \cellcolor{bestcell}5 \\
 & M3  & \tabText & 5.2 & 8.3 & 8.4 & 16.3 & 653 & 8.1 & 5.3 & 5 \\
 & M4  & \tabStruct & 7.5 & 7.9 & 8.5 & 14.8 & 605 & 10.1 & 4.6 & 5 \\
 & M5  & \tabStruct & 9.0 & 17.4 & 12.5 & 15.7 & 620 & \cellcolor{bestcell}\textbf{15.5} & \cellcolor{worstcell}28.3 & \cellcolor{worstcell}28 \\
 & M6  & \tabHier & 4.5 & 15.0 & 8.3 & 15.6 & 656 & 12.2 & 4.8 & 7 \\
 & M7  & \tabRefine & 7.5 & 13.7 & 9.6 & 16.0 & 629 & 12.2 & \cellcolor{bestcell}4.6 & 17 \\
 & M8  & \tabRefine & 8.2 & 18.5 & 9.5 & 15.3 & 620 & 14.9 & 5.1 & 19 \\
 & M9  & \tabWeight & \cellcolor{worstcell}3.0 & \cellcolor{worstcell}19.5 & \cellcolor{worstcell}3.2 & \cellcolor{worstcell}22.0 & \cellcolor{worstcell}921 & \cellcolor{worstcell}7.4 & 12.7 & 13 \\
 & M11 & \tabAct & \cellcolor{bestcell}\textbf{11.9} & 14.2 & \cellcolor{bestcell}14.5 & \cellcolor{bestcell}13.5 & \cellcolor{bestcell}553 & 13.5 & 16.8 & 17 \\
\midrule
%% ======== Qwen3-32B ========
\multirow{11}{*}{\rotatebox[origin=c]{90}
{\scalebox{0.85}
{\logoQwen\hspace{3pt}\qwenl}}}
 & - & NoMem  & 22.4 & 11.3 & 23.7 & 20.3 & 583 & 17.6 & 3.9 & 4 \\
 & M1  & \tabFlat & 27.6 & \cellcolor{bestcell}9.6 & 16.2 & \cellcolor{worstcell}31.1 & 1.08k & 18.2 & 5.0 & \cellcolor{bestcell}5 \\
 & M2  & \tabFlat & \cellcolor{worstcell}21.6 & 10.8 & \cellcolor{worstcell}15.4 & 30.8 & \cellcolor{worstcell}1.09k & \cellcolor{bestcell}\textbf{19.6} & 5.0 & \cellcolor{bestcell}5 \\
 & M3  & \tabText & 26.9 & 9.9 & 21.8 & 30.3 & 950 & 18.2 & 5.4 & 8 \\
 & M4  & \tabStruct & 26.9 & \cellcolor{worstcell}12.4 & 24.3 & 28.6 & 816 & 17.6 & 4.9 & 20 \\
 & M5  & \tabStruct & 23.1 & 11.4 & 23.0 & 27.4 & 851 & 16.2 & \cellcolor{worstcell}31.2 & \cellcolor{worstcell}31 \\
 & M6  & \tabHier & 23.9 & 9.9 & 22.8 & 26.8 & 817 & 16.2 & 4.4 & 7 \\
 & M7  & \tabRefine & \cellcolor{bestcell}\textbf{32.1} & 11.1 & \cellcolor{bestcell}29.9 & 27.2 & 716 & 17.6 & \cellcolor{bestcell}4.3 & 21 \\
 & M8  & \tabRefine & 22.4 & \cellcolor{worstcell}12.4 & 21.3 & 28.0 & 886 & \cellcolor{worstcell}13.5 & 4.6 & 22 \\
 & M9  & \tabWeight & 29.9 & 12.1 & 27.4 & \cellcolor{bestcell}21.4 & \cellcolor{bestcell}612 & 14.2 & 14.6 & 15 \\
 & M11 & \tabAct & 26.9 & 12.3 & 24.7 & 22.0 & 783 & \cellcolor{worstcell}13.5 & 21.3 & 21 \\
\midrule
%% ======== Gemma-4-26B ========
\multirow{10}{*}{\rotatebox[origin=c]{90}{\scalebox{0.85}{\logoGoogle\hspace{3pt}\gemma}}}
 & - & NoMem  & 7.5 & 11.3 & 17.3 & 16.4 & 671 & 14.2 & 19.8 & 20 \\
 & M1  & \tabFlat & \cellcolor{bestcell}\textbf{10.4} & \cellcolor{worstcell}15.0 & 16.3 & 21.6 & 786 & 18.2 & 21.9 & 22 \\
 & M2  & \tabFlat & 9.7 & 11.5 & 17.2 & 21.4 & 753 & 19.6 & \cellcolor{bestcell}20.6 & \cellcolor{bestcell}21 \\
 & M3  & \tabText & \cellcolor{bestcell}\textbf{10.4} & 14.8 & 14.9 & 20.5 & 781 & \cellcolor{worstcell}14.2 & 23.4 & 23 \\
 & M4  & \tabStruct & 8.2 & 14.5 & 13.9 & \cellcolor{worstcell}22.5 & \cellcolor{worstcell}942 & 18.9 & 22.4 & 23 \\
 & M5  & \tabStruct & \cellcolor{worstcell}4.5 & \cellcolor{bestcell}8.5 & \cellcolor{worstcell}9.4 & 19.9 & 749 & \cellcolor{bestcell}\textbf{20.9} & \cellcolor{worstcell}45.6 & \cellcolor{worstcell}46 \\
 & M6  & \tabHier & \cellcolor{bestcell}\textbf{10.4} & 11.4 & 18.1 & 21.1 & 741 & 18.9 & 24.3 & 28 \\
 & M7  & \tabRefine & 8.2 & 10.1 & 17.8 & 21.1 & 788 & 16.9 & 24.1 & 37 \\
 & M8  & \tabRefine & 8.2 & 12.5 & 18.5 & 20.1 & 763 & 18.2 & 21.2 & 26 \\
 & M11 & \tabAct & 9.0 & 11.7 & \cellcolor{bestcell}20.1 & \cellcolor{bestcell}17.1 & \cellcolor{bestcell}710 & 16.9 & 31.6 & 32 \\
\bottomrule
\end{tabular}%
}
\end{table*}

\subsection{Agent-centric Benchmark}
\label{sec:agent-centric}

\textbf{The same retrieval operation that sharpens code generation starves embodied
planning, splitting the agent-centric regime in two.} On ALFWorld with the two Qwen backbones, the substrates that win
are those that \emph{denoise} accumulated experience before it reaches the policy: M11, in its utterance-level re-prefill approximation of EpiCache,
doubles NoMem on \qwens by re-prefilling only the matched episode rather than the full trajectory buffer, and M7 achieves the table's peak TSR of
$32.1\%$ on \qwenl by distilling trajectories
into compact, reusable strategies, significantly above flat retrieval with M2 ($21.6\%$, exact McNemar $p{=}0.016$, Appendix~\ref{app:significance}). Both substrates share a common operation: they suppress
task-irrelevant tokens before retrieval, either by clustering raw turns into episodes
(M11) or by abstracting trajectories into reasoning templates (M7). Skill-based agent memories take the same route and store reusable skills distilled from past trajectories~\citep{zhang2026memskill,xu2026hyperskill}. Substrates that lack this denoising step show inconsistent gains across backbones. On \qwenl, flat retrievers range from slightly below NoMem (M2, 21.6 vs.\ 22.4) to well above it (M1, 27.6), and the text and structural stores (M3 26.9, M4 26.9, M5 23.1) likewise match or exceed NoMem, yet none close the gap to the refinement winner M7 (32.1). These gaps to NoMem lie within the bootstrap intervals of Appendix~\ref{app:significance}, so we read them as descriptive. On \qwens, however, flat retrievers straddle NoMem (M1 5.2 below, M2 6.7 above 5.7), and on \gemmashort{} the flat, text, and hierarchical stores share the lead (10.4), suggesting that the value of raw-text retrieval on embodied planning is backbone-dependent rather than universally harmful. On embodied planning, retrieved trajectories compete for
attention with the current observation and admissible-action list, so on the Qwen backbones the substrate's
value lies in noise removal rather than recall. Memories that compress raw
experience into actionable abstractions help most, while raw retrieval gains less because it dilutes the cue the policy
depends on. This is consistent with the action bottleneck reported for agentic RL, where the
training signal concentrates on the few action tokens of a trajectory~\citep{he2026resolving}.

On BigCodeBench-Hard the picture inverts because retrieved code snippets do not compete
with the task prompt, they extend it: a working solution to a related problem is reusable
scaffolding rather than a distractor.\enlargethispage{-2\baselineskip} Retrieval therefore helps broadly on \qwens{} and \gemmashort, with M5 leading
both ($15.5\%$ and $20.9\%$ Pass@1) by surfacing structurally related
code through its graph and vector hybrid, and M2 lifts Pass@1 above NoMem on every
backbone. Yet the cost trade-off is steep: on \qwenl{} M2 ($19.6\%$) overtakes M5
($16.2\%$) at roughly $1/6$ the latency, and across backbones M5 pays 2 to 6 times the
inference cost of flat substrates, making it a strictly dominated choice when latency
matters. These Pass@1 gaps are small, so we read the BigCodeBench-Hard rankings as descriptive and rest the cost argument on the latency gaps. The two benchmarks therefore stress opposite memory operations: ALFWorld rewards
substrates that abstract noise away before retrieval, while BigCodeBench rewards
substrates that retrieve broadly without paying for abstraction, an asymmetry consistent
with the attention shift of Section~\ref{sec:retrieval-scaling}.

\begin{figure}[t]
  \centering
  \includegraphics[width=\textwidth]{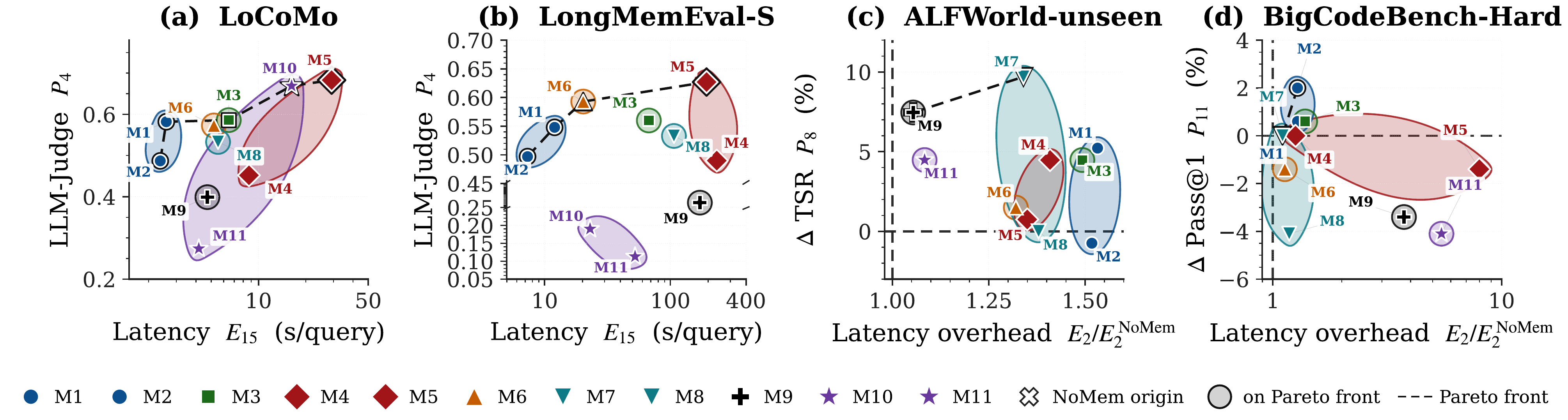}
    \caption{Performance vs.\ latency across four benchmarks on \qwenl. (a) LoCoMo and (b) LongMemEval-S: absolute $P_4$. (c) ALFWorld-unseen and (d) BigCodeBench-Hard: $\Delta$-performance vs.\ inference-time overhead $E_2/E_2^{\mathrm{NoMem}}$ (p90 in (c), mean in (d)) anchored at NoMem.}
  \label{fig:perf}
  \vspace{-1em}
\end{figure}

\begin{figure}[h]
  \centering
  \includegraphics[width=\textwidth]{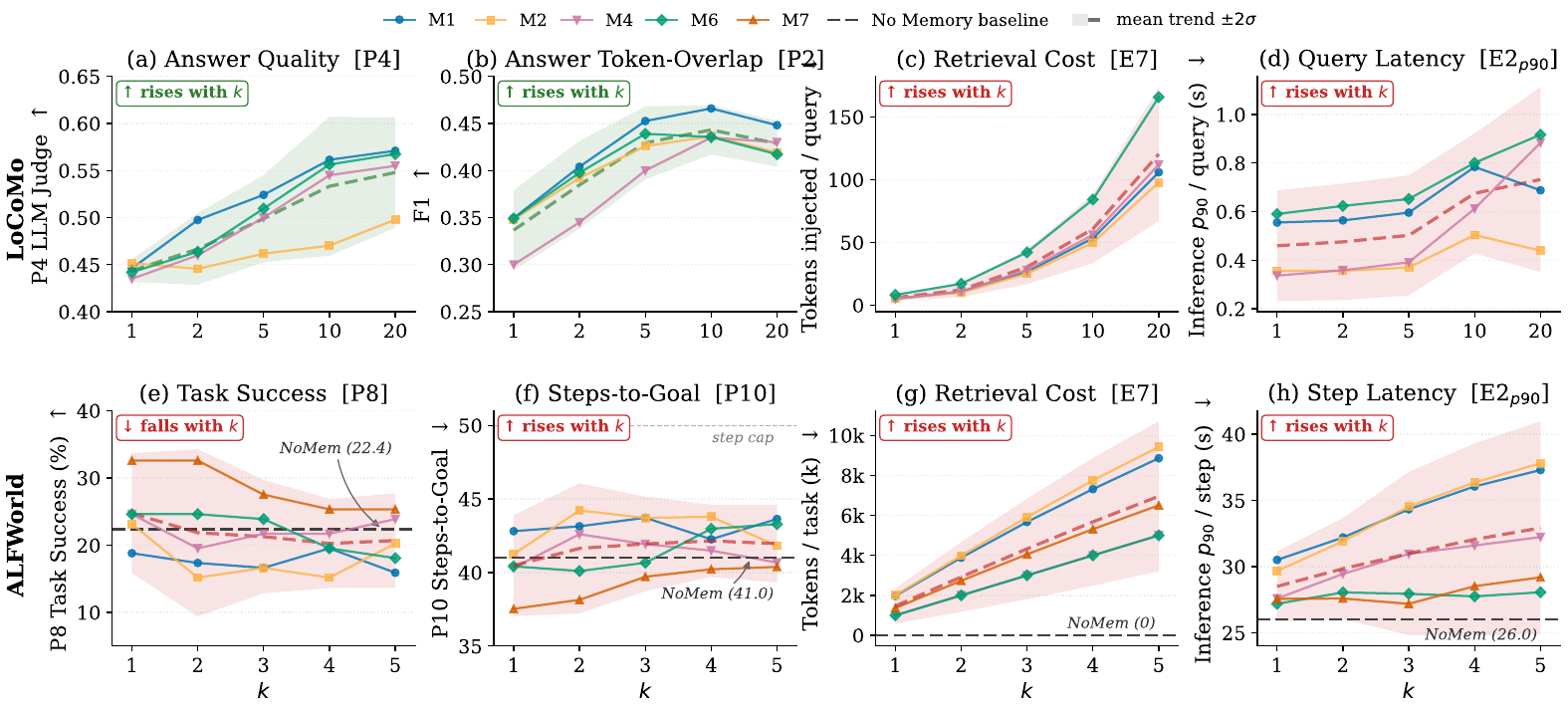}
  \caption{Retrieval breadth $k$ scales oppositely on long-context QA versus sequential decision-making. Top: LoCoMo (\qwens, $n{=}1986$, $k{\in}\{1,2,5,10,20\}$). Bottom: ALFWorld-unseen (\qwenl, $n{=}134$, $k{\in}\{1,2,3,4,5\}$). \textcolor[HTML]{2E7D32}{green} when the trend is in the preferred direction, \textcolor[HTML]{C62828}{red} otherwise. Panel (f) shows mean steps over all episodes, unlike $P_{10}$, which averages successful episodes only. Black dashed lines mark NoMem, shown as No Memory baseline in the legend.}
  %\vspace{-2em}
  \label{fig:abalation}
\end{figure}

\section{Discussion}
\label{sec:discussion}

\subsection{Performance--Latency Tradeoffs Across Substrate Families}
\label{sec:perf-latency}

Figure~\ref{fig:perf} plots every memory substrate against its per-query latency under a shared \qwenl{} backbone. Panels (a, b) report absolute $P_4$ on
long-form dialog QA; panels (c, d) re-anchor at NoMem so the y-axis becomes the gain over
no-memory and the x-axis the inference-time overhead $E_{2}/E_{2}^{\mathrm{NoMem}}$ (p90 on ALFWorld, mean on BCB-Hard).

\textbf{Disjoint Pareto frontiers across regimes.}
The QA frontier (a, b) rises from the flat indices through M6, and through M3 and M10 on LoCoMo, to M5, which wins at $10$ to $30\times$ the latency of the
baselines. M6's plotted latency is dominated by a one-off tree
build absorbed by the first query (Appendix~\ref{app:full_table}), and
its steady-state per-query cost is comparable to the Flat baselines. The agentic frontier (c, d) ends elsewhere, at flat retrieval on BCB-Hard (M2, $+2.0$\% at $1.28\times$
overhead) and at the adapter M9 ($+7.5$\% at $1.05\times$) and refinement-based distillation M7 ($+9.7$\% at $1.34\times$) on ALFWorld.
That the quality ends of the two frontiers share \emph{no} substrate confirms the divide of
Section~\ref{sec:agent-centric}: the regimes reward fundamentally different operations,
recall over a stored corpus versus concise procedural cues, so a substrate engineered for
one provides no advantage at the other. Read as an allocation rule, a QA deployment moves along its frontier from the flat indices through M6 to M5, which buys the highest $P_4$ at 10 to 30 times the baseline latency, while an agentic deployment relies on M2 for code reuse and on M7 for embodied planning.

\textbf{Cost is heavy-tailed and frequently unrewarded.}
Additional latency rarely buys quality monotonically: on LME-S, M4 at $234$~s underperforms
M2 at $7.3$~s, and M9 at $172$~s buys little quality ($P_4{=}0.29$). The asymmetry sharpens on
agentic tasks, where M5, M9, and M11 incur $3.7$ to $8.0$ times overhead on BCB-Hard while
landing strictly below NoMem ($\Delta$Pass@1 between $-1.4$\% and $-4.1$\%). Substrate
cost therefore reflects mechanism complexity rather than expected gain: heavy machinery
earns its overhead only when its mechanism aligns with the task bottleneck, and turns into
pure tax otherwise. Attributing cost to stages changes the picture further. M4 and M5 spend most of their run time in the write phase ($E_{14}$ of roughly 68--70K\,s and 56--58K\,s on LME-S, against mostly under 5K\,s for M1, M2, and M10; Appendix~\ref{app:lme-mab-deep-analysis}), so read-side accounting alone understates their cost. Because M9 pays for adapter training in GPU time while most other substrates pay in auxiliary-LLM tokens, we report wall-clock ($E_{14}$) and token counts ($E_{12}$) on separate axes rather than converting them into one currency. Family labels predict performance only weakly (Flat is mediocre on QA
yet leads BCB-Hard; Refine spans M7 best on ALFWorld and M8 near the bottom on two
panels), ruling out family-level routing and motivating the per-substrate analysis we
develop next.

\subsection{Retrieval Scaling Reverses Across Regimes}
\label{sec:retrieval-scaling}

Figure~\ref{fig:abalation} sweeps retrieval depth $k$ on LoCoMo with \qwens and on ALFWorld with \qwenl. On LoCoMo, $P_4$ at $k{=}20$ exceeds $P_4$ at $k{=}1$ for every substrate: each
retrieval raises the chance of surfacing a gold evidence turn, and surplus context is
passively ignorable because factual QA only requires the model to \emph{find and extract}.
On ALFWorld the sign flips: task success \emph{falls} with $k$ (M7 falls by about seven points between
$k{=}1$ and $k{=}5$, while the flat retrievers M1 and M2 fall below the no-memory baseline of $22.4\%$), and
mean steps per episode rise slightly (panel~f). Because the dashed mean trend and its shaded band flip sign
between the two sweeps, which also differ in backbone, retrieval breadth rather than any individual substrate carries the reversal. The mean trend rises at every step on LoCoMo and stays below its $k{=}1$ value at every larger depth on ALFWorld, where M7 leads at every depth, so the reversal is not tied to one retrieval depth (Appendix~\ref{app:calibration}).

\begin{wrapfigure}[15]{r}{0.5\textwidth}
  %\vspace{-12pt}
  \centering
  \includegraphics[width=0.46\textwidth]{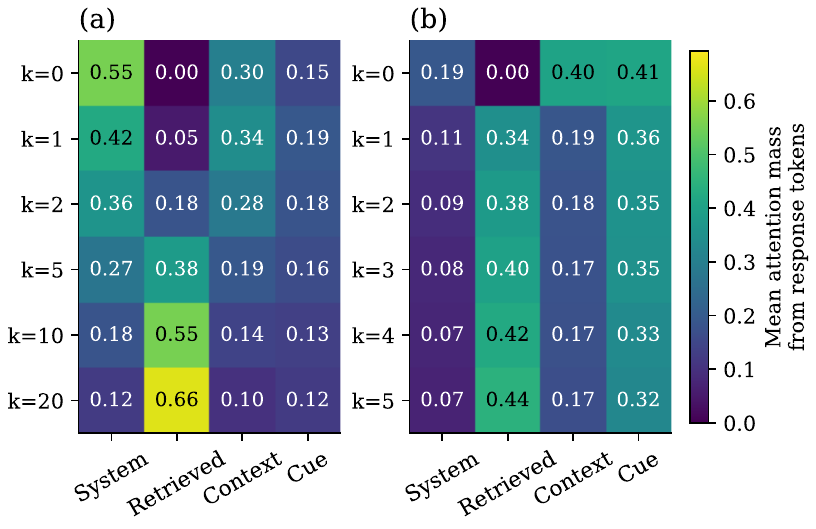}
  \caption{\textbf{Response-token attention under M1 retrieval} for (a) LoCoMo and (b) ALFWorld.}
  \label{fig:heatmap}
  %\vspace{-10pt}
\end{wrapfigure}

% \paragraph{Design implication: trade read breadth for write depth.}
% The asymmetry prescribes opposite strategies.
% For user-centric memory, maximize $k$ with cheap-write substrates (M1, M2): the accuracy-per-token slope stays positive even at $k{=}20$.
% For agent-centric memory, minimize $k$ to 1 and invest in write-time distillation: M7 at $k{=}1$ achieves 35\% TSR with 1.4k retrieved tokens, strictly dominating M1 at $k{=}5$ (15\% TSR, 5k tokens).
% When retrieval noise is cheap, retrieve broadly; when retrieval noise is toxic, distill aggressively and retrieve narrowly.

\textbf{Retrieval shifts attention in both regimes, but its consequence depends on where the answer
lives.} As shown in Figure~\ref{fig:heatmap}, we probe the attention of the response tokens of
\qwens (eager attention, upper-half layers, head-summed mass) under M1
retrieval and bin attention into four regions: System, Retrieved, task Context
(transcript, trajectory, observation, admissible actions), and Cue (question and
next-action prompt). On \emph{both} tasks, growing $k$ funnels mass out of Context into
Retrieved (LoCoMo: $0.34\!\to\!0.10$ vs.\ $0.05\!\to\!0.66$; ALFWorld:
$0.19\!\to\!0.17$ vs.\ $0.34\!\to\!0.44$): the mechanism is identical, what differs is
which region carries the answer. On LoCoMo the load-bearing facts live in the retrieved
utterances, so siphoning attention toward them is precisely the desired behavior and
$P_4$ rises. On ALFWorld the load-bearing region is the observation and admissible-action
list, so the same shift starves the decision and TSR falls; Cue mass also loses
$\approx 4$\% between $k{=}1$ and $k{=}5$. The probe covers one backbone and M1, so it is consistent with the TSR decline of the \qwenl{} sweep rather than a test of it. The implication for memory design is that retrieval
breadth is not a universal hyperparameter but a regime-conditioned one: any routing
decision should be informed by whether the prompt's answer lives in retrieved memory or in
the task context window.

% The attention probe shows that as k grows, attention shifts away from the task context and toward the retrieved block. This same shift has opposite consequences depending on the task. For LoCoMo, retrieved utterances often contain the answer, so moving attention toward retrieval is beneficial. For ALFWorld, the crucial information is in the current observation and admissible-action list, so moving attention away from task context starves the policy and hurts decision-making.

\subsection{Scalability Under Growing Input Context}
\label{sec:scalability}

\begin{wrapfigure}[12]{l}{0.5\textwidth}
  %\vspace{-12pt}
  \centering
  \includegraphics[width=0.46\textwidth]{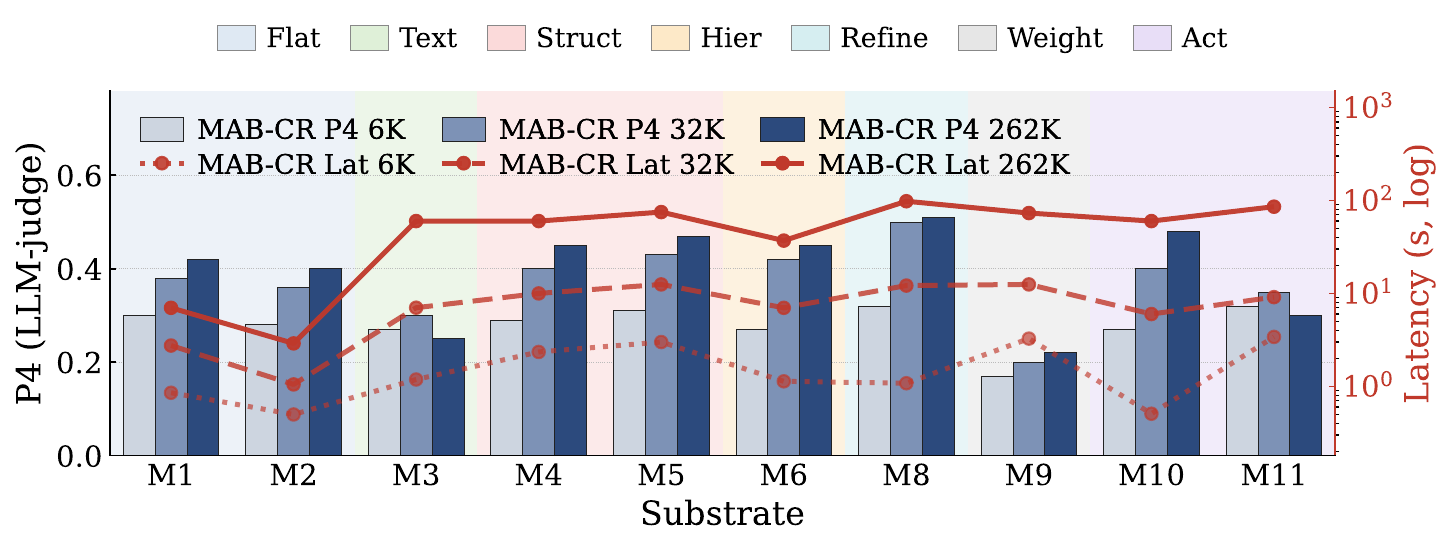}
  \caption{MAB CR scalability on \qwenl: $P_4$ (bars, left) and per-query latency (lines, right, log) at 6K, 32K, 262K tokens.}
  \label{fig:cr-scale}
  %\vspace{-10pt}
\end{wrapfigure}

\textbf{Scalability is a third routing axis: substrates that win at moderate scale can
become costly at long horizons, and vice versa.} Figure~\ref{fig:cr-scale} sweeps
content length on MAB Conflict Resolution from 6K to 262K tokens. Most external substrates
improve with scale: M8 climbs from $0.32$ to $0.51$ and M5 from $0.31$ to $0.47$, because
longer histories give their write-side abstraction and entity-relation updates more
material to disambiguate which fact is current. Quality and cost grow together. The refinement family, represented here by the harness approximation M8, gains the most quality
but also the most latency, from about 1\,s at 6K to about 98\,s at 262K, the highest in the sweep,
because every refined bundle is prefilled at read time.

The structural M4 and M5 and the full-context M10 reach 60 to 75\,s at 262K, and only the flat indices stay under 10\,s. Among the internal substrates, M10 also gains in quality ($0.27{\to}0.48$), although at 262K its buffer exceeds the context window and truncation keeps only the most recent entries, which on this recency task drops superseded facts; M9 stays at 0.17--0.22, lacking any mechanism for selective temporal update.
\textbf{History depth is a selection signal beyond the current query.}
Beyond regime (Section~\ref{sec:perf-latency}) and retrieval depth
(Section~\ref{sec:retrieval-scaling}), the substrates whose quality grows most with history (M8, M5, M10) pay one to two orders of magnitude more latency at 262K, while the flat indices stay cheap but gain less. A
substrate optimal for short sessions can thus become a liability over a long-running
agent's lifetime.

% This is the key takeaway from this paper. Consider to further highlight this paragraph using a box (or seperate subsection).  
% \vspace{-0.5em}
% \paragraph{Toward universal memory: multi-substrate or multi-agent composition.}
% The three diagnostic axes developed across Sections~\ref{sec:perf-latency},
% \ref{sec:retrieval-scaling}, and \ref{sec:scalability} converge on the same conclusion: no
% single substrate can simultaneously serve recall over long histories, action selection
% under attention pressure, and graceful scaling to long horizons, because the operations
% these regimes reward are not just quantitatively but qualitatively different. The path
% forward is therefore not a better single substrate but a \emph{multi-substrate or
% multi-agent memory system} in which heterogeneous stores are composed and each handles the
% slice of the workload it is suited for: refinement substrates (M7, M8) curate
% \emph{abstractions} such as reusable skills and reasoning templates, structural and text
% substrates (M3, M5) hold \emph{facts} and entity-relation knowledge with explicit update
% semantics, and activation or flat substrates (M10, M11, M1, M2) preserve raw
% \emph{experience} for verbatim recall when retrieval noise is cheap. The role of a
% universal memory harness is then to route each query to the substrate whose stored
% representation aligns with the load-bearing region of that query's prompt, a routing
% problem whose empirical signal the present work makes available for the first time.

\newtcolorbox{papertakeaway}{
  enhanced,
  colback=blue!5!white,
  colframe=blue!60!black,
  boxrule=1pt,
  arc=4pt,
  left=12pt, right=10pt, top=6pt, bottom=6pt,
  fontupper=\normalsize,
  before skip=6pt,
  after skip=6pt,
  attach boxed title to top left={xshift=12pt, yshift=-8pt},
  boxed title style={
    colback=blue!60!black,
    colframe=blue!60!black,
    boxrule=0pt,
    arc=3pt,
    left=8pt, right=8pt, top=3pt, bottom=3pt,
  },
  title={\textsc{\textbf{No Substrate Wins Alone: Composition Is a Path to Universal Memory}}},
  coltitle=white,
  fonttitle=\small,
  drop shadow={black!30!white},
}

\begin{papertakeaway}
The three diagnostic axes of Sections~\ref{sec:perf-latency}--\ref{sec:scalability}
converge: no single substrate can simultaneously serve recall over long histories, action
selection under attention pressure, and graceful scaling to long horizons, because these
regimes reward qualitatively different operations. A more promising path than a better single
substrate is a \emph{multi-substrate or multi-agent memory system} composed by role, in which
\textcolor{refinecolor}{\textbf{refinement}} substrates curate \emph{abstractions},
\textcolor{structcolor}{\textbf{structural}} and \textcolor{textcolor}{\textbf{text}}
substrates hold \emph{facts}, and \textcolor{actcolor}{\textbf{activation}} and
\textcolor{flatcolor}{\textbf{flat}} substrates preserve raw \emph{experience}. Substrate selection is therefore a design axis of the
harness. Choosing a substrate per observable regime beats the best fixed substrate on our benchmarks, and an ideal selector over two or three heterogeneous substrates shows how much a finer router could still gain (Appendices~\ref{app:routing} and~\ref{app:portfolio}).
\end{papertakeaway}

\section{Related Work}
\label{sec:related}
\paragraph{Memory-augmented LLM agents.}
Agent memory systems span a broad substrate design space, from external stores (vector
indices, text buffers, knowledge graphs, hierarchies, episodic narratives, distilled skill
memories~\citep{park2023generative,packer2023memgpt,chhikara2025mem0,rasmussen2025zep,jiang2026magma,sarthi2024raptor,sun2026h,zhou2026amory,zhang2026memskill})
to internal stores (parametric updates~\citep{zhang2023lora} and KV-cache
activations~\citep{li2024snapkv,kim2025epicache}), with hybrids combining
several~\citep{li2025memos,latimer2025hindsight,wang2025mirix}. Most systems commit to a
fixed configuration at design time; we instead isolate the substrate as the experimental
variable and identify when each is preferable.

\paragraph{Memory evaluation benchmarks and practices.}
Existing benchmarks cover user-centric~\citep{maharana2024evaluating,wulongmemeval,hu2025evaluating,zhang2026memorycd,ma2026memaudit}
and agent-centric settings~\citep{shridhar2020alfworld,zhuobigcodebench,jimenez2023swe}, but
the two regimes are typically evaluated in isolation, and evaluations emphasize end-task
accuracy while omitting deployment-critical dimensions such as write cost, latency,
storage, and management overhead. Our harness addresses both gaps; details are in
Appendix~\ref{app:related_work}.

\section{Conclusion}
\label{sec:conclusion}
We presented a unified metric suite and a controlled harness evaluation of agent memory across 11 substrates, three
backbones, four benchmarks, and 26 metrics. The optimal substrate \emph{reverses} between
long-form QA and agentic decision-making and shifts again with history depth. These reversals make substrate selection a design axis. Choosing a substrate per observable regime beats the best fixed substrate on our benchmarks, a router that reads only the question embedding gains on LoCoMo when its training data come from the same conversations, and most of the gap to an oracle selector remains open.

\begin{ack}
We thank the anonymous reviewers for their constructive feedback. This work is supported by the BDSC Lab at the University of Illinois Chicago through U.S. National Science Foundation (NSF) grant POSE-2346158. It is also supported in part by NSF CAREER Award 2337877; the Schmidt Sciences Award on AI \& Advanced Computing through the Science of Trustworthy AI program; and the University of Washington Tech Policy Lab. Any opinions, findings, conclusions, or recommendations expressed in this material are those of the authors and do not necessarily reflect those of the NSF or Schmidt Sciences.
\end{ack}

\bibliographystyle{unsrtnat}
\bibliography{references}

%%%%%%%%%%%%%%%%%%%%%%%%%%%%%%%%%%%%%%%%%%%%%%%%%%%%%%%%%%%%

\newpage

\appendix
\raggedbottom

\section{Landscape of LLM Memory-Augmented Systems}
\label{app:landscape}
This appendix expands the motivation in Section~\ref{sec:intro} with the full survey of
\textbf{52 memory-augmented LLM systems} released between early 2023 and April 2026. We
document not the mechanisms of individual systems, which are catalogued by the recent
surveys of agentic memory~\cite{huang2026rethinking,hu2025memory,guo2026multimodal}, but how the current
work evaluates them. The result is shown in Table~\ref{tab:app-systems}.

\subsection{Inclusion criteria and data collection}
\label{app:landscape-criteria}

From this collection, we included papers that (a) propose a
novel memory method with an associated substrate or architecture, (b) report evaluation on
at least one memory benchmark, and (c) have a publicly available preprint. We excluded
survey papers, benchmark-only papers, and systems assessed only qualitatively. For each
candidate, we independently verified the arXiv identifier, title, year of first public
preprint, and evaluation setting against primary sources, discarding entries whose benchmarks or backbone
models could not be reliably confirmed.

For each system we record four evaluation dimensions.
(i)~\textbf{Benchmarks}, grouped into six canonical categories:
\textsc{LC}=LoCoMo~\citep{maharana2024evaluating},
\textsc{LME}=LongMemEval~\citep{wulongmemeval},
\textsc{MH}=multi-hop QA (HotPotQA, MuSiQue, 2WikiMQA),
\textsc{DS}=DialSim, \textsc{DMR}=DMR, \textsc{Other}=other (MSC, Carecall, Conversation
Chronicles, MT-Bench+, Time-Dialog, RULER, NarrativeQA, PopQA, NQ, LongBench,
$\infty$-Bench, ESConv, PersuasionForGood, MemoryAgentBench, PersonaMem, custom
evaluations, \ldots).
(ii)~\textbf{Performance metrics} (\textsc{A}=accuracy, \textsc{P}=per-type breakdown).
(iii)~\textbf{Efficiency metrics} (\textsc{T}=token usage, \textsc{L}=latency,
\textsc{R}=runtime, \textsc{M}=memory footprint, \textsc{K}=API-call count); only numeric
reports count. The five-dimension scheme used here is the most that any prior system
reports; the harness in Section~\ref{sec:setup} extends this to 15 efficiency metrics
($E_1$--$E_{15}$) covering write, read, and management phases separately, together with
aggregate token, call, and latency totals.
(iv)~\textbf{Base LLM family}: \textbf{G} for GPT-series (GPT-3.5/4/4o/4.1/5, including
\emph{-mini} variants) or \textbf{O} for non-GPT open-source models (Llama, Qwen, DeepSeek,
\ldots).

\subsection{Aggregate statistics}
\label{app:landscape-stats}

\paragraph{Benchmark concentration.}
Of 76 benchmark--system pairs in the 52 curated papers, LoCoMo accounts for 29 (38\%) and
LongMemEval for 18 (24\%), together covering 62\% of adopted benchmarks. Adding DMR (2)
and DialSim (1) pushes the dialogue-centric share to 66\%. The multi-hop QA cluster
contributes 4 pairs, drawn mainly from the HippoRAG/PropRAG lineage. The residual ``Other''
category (22) is heterogeneous but, on closer inspection, remains overwhelmingly
dialogue-based: MSC, Carecall, Conversation Chronicles, MT-Bench+, and PersonaMem are all
multi-session dialogue benchmarks; MemoryAgentBench~\citep{hu2025evaluating}, adopted
as a primary evaluation by Mem-$\alpha$~\citep{wang2025memalpha}, is likewise a
multi-turn dialogue memory benchmark that re-packages QA, classification, and
summarization over incremental context. Only ReadAgent, RAPTOR,
EM-LLM~\citep{fountashuman}, ComoRAG~\citep{wang2026comorag}, and
MemAgent~\citep{yu2025memagent} target non-dialogue long-document settings (QuALITY,
NarrativeQA, LongBench, $\infty$-Bench, RULER). None of the 52 systems that evaluate on a
memory benchmark also reports an evaluation on agent-centric task execution. Memory systems
evaluated only on agent benchmarks, such as ReasoningBank~\citep{ouyang2025reasoningbank},
fall outside criterion (b) and are therefore not in Table~\ref{tab:app-systems}. Our
harness evaluates every substrate on ALFWorld and BigCodeBench-Hard next to the memory
benchmarks.

\paragraph{Metric coverage.}
Every system (100\%) reports end-task accuracy and 35 (67\%) also report a per-type
breakdown. By contrast, only 11 (21\%) report any efficiency metric: latency (12\%),
token consumption (10\%), runtime (4\%), memory footprint (4\%), and API-call count (2\%),
with zero reports of management cost or compression ratio.
LightMem~\citep{fang2025lightmem} is the sole system to report four efficiency dimensions
simultaneously (token, latency, runtime, and API-call count);
Mem0~\citep{chhikara2025mem0} and Memoria~\citep{sarin2025memoria} each report two; the
remainder report one or none.

\paragraph{Base-model concentration.}
42 of 52 systems (81\%) use a GPT-family model as their primary backbone, typically
GPT-4o-mini for extraction combined with GPT-4o or GPT-4.1-mini for generation.
The non-GPT systems span at least four lineages: the HippoRAG / PropRAG / R$^3$Mem family
on multi-hop QA~\citep{gutierrez2024hipporag,wang2025proprag,wang2025r3mem}, the RL line
including Memory-R1 and Memory-T1 on Llama-3.1-8B and
Qwen-2.5~\citep{yan2025memory,du2025memory},
Mem-$\alpha$ on Qwen3-4B~\citep{wang2025memalpha}, and MemAgent on Qwen-2.5 with
RL~\citep{yu2025memagent}.

\paragraph{Joint coverage.}
Figure~\ref{fig:landscape}(c) projects each system onto the \emph{benchmark breadth}
$\times$ \emph{efficiency depth} plane. 26 of 52 systems (50\%) occupy the origin cell
$(1,0)$, a single benchmark with no efficiency metric. Only two
systems~\citep{yan2025memory,wang2026memmachine} evaluate on three or more benchmarks, and
only LightMem reports three or more efficiency dimensions. No system occupies the
upper-right region $({\geq}3, {\geq}3)$ that would indicate broad coverage together with
comprehensive efficiency reporting.

\subsection{Trends over time}
\label{app:landscape-trends}

The 2026 cohort does not close the evaluation gap. Only 2 of 9 (22\%) report any
efficiency metric, statistically indistinguishable from the pre-2026 cohort. Per-type
breakdowns, however, have become near-universal: 9 of 9 2026 systems (100\%) report them,
up from 60\% in pre-2026 work. The field has converged on fine-grained accuracy analysis
as a standard but treats efficiency as an afterthought, an asymmetry we identify as the
single largest obstacle to principled substrate routing of the kind that universal
agent memory would need.

\begin{table*}[t]
\centering
\caption{\textbf{Landscape of 52 memory-augmented LLM systems (2023--2026).}
\textbf{Year}: first public preprint.
Benchmark codes: LC\,=\,LoCoMo, LME\,=\,LongMemEval, MH\,=\,Multi-hop QA, DS\,=\,DialSim, DMR\,=\,DMR.
Metric codes: A\,=\,accuracy, P\,=\,per-type, T\,=\,token, L\,=\,latency, R\,=\,runtime, M\,=\,memory size, K\,=\,API calls.
\textbf{\#B}: distinct benchmarks used.
\textbf{\#E}: efficiency dimensions reported (bold\,${\geq}2$).
\textbf{Base}: G\,=\,GPT-series, O\,=\,open-source.
\textbf{Summary}: 81\% GPT-family; 62\% of benchmark pairs on LC\,+\,LME; only 21\% report any efficiency metric; no system reaches ${\geq}3$ benchmarks $\times$ ${\geq}3$ efficiency dimensions.}
\label{tab:app-systems}
\vspace{2pt}
\resizebox{\textwidth}{!}{%
\renewcommand{\arraystretch}{1.15}
\begin{tabular}{@{}l c l l c c c@{\hspace{8pt}}l c l l c c c@{}}
\toprule
\textbf{System} & \textbf{Year} & \textbf{Bench.} & \textbf{Metrics} & \textbf{\#B} & \textbf{\#E} & \textbf{Base}
& \textbf{System} & \textbf{Year} & \textbf{Bench.} & \textbf{Metrics} & \textbf{\#B} & \textbf{\#E} & \textbf{Base} \\
\midrule
Generative Agents~\citep{park2023generative} & 2023 & Other     & A              & 1 & 0 & G
& Mem0~\citep{chhikara2025mem0}             & 2025 & LC        & A, P, T, L     & 1 & \textbf{2} & G \\
MemGPT~\citep{packer2023memgpt}             & 2023 & DMR       & A              & 1 & 0 & G
& Mem-$\alpha$~\citep{wang2025memalpha}     & 2025 & Other     & A, P, M        & 1 & 1 & O \\
MemoChat~\citep{lu2023memochat}             & 2023 & Other     & A              & 1 & 0 & G
& MemAgent~\citep{yu2025memagent}           & 2025 & Other     & A              & 1 & 0 & O \\
MemoryBank~\citep{zhong2024memorybank}      & 2023 & Other     & A              & 1 & 0 & G
& MemOS~\citep{li2025memos}                 & 2025 & LC, LME   & A, P           & 2 & 0 & G \\
RSum~\citep{wang2025recursively}                   & 2023 & Other     & A              & 1 & 0 & G
& Memoria~\citep{sarin2025memoria}               & 2025 & LME       & A, T, L        & 1 & \textbf{2} & G \\
SCM~\citep{wang2025scm}                     & 2023 & Other     & A              & 1 & 0 & G
& MemoryOS~\citep{kang2025memory}         & 2025 & LC        & A, P           & 1 & 0 & G \\
\cmidrule(lr){1-7}\cmidrule(lr){8-14}
AI PERSONA~\citep{wang2024ai}        & 2024 & Other     & A              & 1 & 0 & G
& Memory-R1~\citep{yan2025memory}            & 2025 & LC, LME, Other & A, P      & 3 & 0 & O \\
EM-LLM~\citep{fountashuman}             & 2024 & Other     & A, P           & 1 & 0 & O
& Memory-T1~\citep{du2025memory}            & 2025 & LC, Other & A, P           & 2 & 0 & O \\
HippoRAG~\citep{gutierrez2024hipporag}      & 2024 & MH        & A, P           & 1 & 0 & O
& MIRIX~\citep{wang2025mirix}               & 2025 & LC, Other & A, P, M        & 2 & 1 & G \\
LD-Agent~\citep{li2025hello}              & 2024 & Other     & A, P           & 1 & 0 & G
& MMS~\citep{zhang2025multiple}                       & 2025 & LC        & A, P           & 1 & 0 & G \\
MemTree~\citep{rezazadeh2024isolated}        & 2024 & Other     & A              & 1 & 0 & G
& Nemori~\citep{nan2025nemori}                 & 2025 & LC, LME   & A, P           & 2 & 0 & G \\
RAPTOR~\citep{sarthi2024raptor}             & 2024 & Other     & A              & 1 & 0 & G
& O-Mem~\citep{wang2025mem}                    & 2025 & LC, LME   & A, P           & 2 & 0 & G \\
ReadAgent~\citep{leehuman}          & 2024 & Other     & A              & 1 & 0 & G
& PRINCIPLES~\citep{kim2025principles}      & 2025 & Other     & A              & 1 & 0 & G \\
THEANINE~\citep{ong2025towards}            & 2024 & Other     & A              & 1 & 0 & G
& PropRAG~\citep{wang2025proprag}           & 2025 & MH, Other & A, P           & 2 & 0 & O \\
\cmidrule(lr){1-7}\cmidrule(lr){8-14}
A-Mem~\citep{xu2025mem}                    & 2025 & LC, DS    & A, P, T        & 2 & 1 & G
& R$^3$Mem~\citep{wang2025r3mem}            & 2025 & Other     & A              & 1 & 0 & O \\
ComoRAG~\citep{wang2026comorag}               & 2025 & Other     & A              & 1 & 0 & G
& RGMem~\citep{tian2025rgmem}               & 2025 & LC, Other & A, P           & 2 & 0 & G \\
GAM~\citep{yan2025general} & 2025 & LC  & A              & 1 & 0 & G
& RMM~\citep{tan2025prospect}                    & 2025 & LC        & A, P           & 1 & 0 & G \\
H-Mem~\citep{ye2026h}                   & 2025 & LC        & A, P, R        & 1 & 1 & O
& SeCom~\citep{pansecom}                & 2025 & LC        & A, P           & 1 & 0 & G \\
Hindsight~\citep{latimer2025hindsight}      & 2025 & LC, LME   & A, P           & 2 & 0 & G
& SGMem~\citep{wu2025sgmem}                   & 2025 & LC, LME   & A, P           & 2 & 0 & G \\
HippoRAG\,2~\citep{gutierrez2025rag}  & 2025 & MH        & A, P           & 1 & 0 & O
& EMem~\citep{zhou2025simple}& 2025 & LC, LME   & A, P           & 2 & 0 & G \\
LightMem~\citep{fang2025lightmem}           & 2025 & LC, LME   & A, P, T, L, R, K & 2 & \textbf{4} & G
& Zep~\citep{rasmussen2025zep} & 2025 & DMR, LME  & A, P, L        & 2 & 1 & G \\
LiCoMemory~\citep{huang2025licomemory}           & 2025 & LC, LME   & A, P, L        & 2 & 1 & G
& & & & & & \\
\cmidrule(lr){1-7}\cmidrule(lr){8-14}
Agentic Memory~\citep{yu2026agentic} & 2026 & LC        & A, P           & 1 & 0 & G
& MemMachine~\citep{wang2026memmachine}         & 2026 & LC, LME, MH & A, P, T      & 3 & 1 & G \\
EverMemOS~\citep{hu2026evermemos}             & 2026 & LC, LME   & A, P           & 2 & 0 & G
& Memori~\citep{borro2026memori}                 & 2026 & LC        & A, P           & 1 & 0 & G \\
Cognis~\citep{daftari2026cognis}          & 2026 & LC, LME   & A, P           & 2 & 0 & G
& MMM~\citep{hu2026memory} & 2026 & LC, LME & A, P          & 2 & 0 & G \\
MAGMA~\citep{jiang2026magma}                     & 2026 & LC, LME   & A, P           & 2 & 0 & G
& SwiftMem~\citep{tian2026swiftmem}             & 2026 & LC, LME   & A, P, L        & 2 & 1 & G \\
 & & & & & &
& TSM~\citep{su2026beyond}                       & 2026 & LC, LME   & A, P           & 2 & 0 & G \\
\bottomrule
\end{tabular}%
}
\end{table*}

\section{Memory Substrate Implementation Details}
\label{app:substrate-details}
\begin{table*}[t]
\centering
\caption{\textbf{Memory substrate configurations.} \textbf{TF}: training-free (\yes) or requires weight modification (\no). \textbf{LLM}: auxiliary LLM required (\yes) beyond final-answer generation. $^{\ddagger}$ attention-guided utterance selection with re-prefill (Qwen3 hybrid-attention adaptation). $^{\S}$ zero-shot LLM controller in place of PPO.}
\label{tab:substrate}
\scriptsize
\setlength{\tabcolsep}{4pt}
\renewcommand{\arraystretch}{1.1}
\begin{tabularx}{\textwidth}{@{}l l c c Y Y Y@{}}
\toprule
\textbf{Family} & \textbf{Method} & \textbf{TF} & \textbf{LLM} & \textbf{Write} & \textbf{Read} & \textbf{Mgmt} \\
\midrule
\classband{bandExt}{\textit{External Memory}} \\
\famcell{Flat Index}
   & \textbf{M1}~~Dense Vector~\citep{xiongapproximate}               & \yes & \no  & Encode; add to index                              & Dense top-$k$                                           & --- \\
   & \textbf{M2}~~Sparse Vector~\citep{robertson2009probabilistic}   & \yes & \no  & Tokenize; append to BM25                          & Sparse top-$k$                                          & --- \\
\cmidrule(l{0.4em}r{0.4em}){1-7}
\famcell{Text Record}
   & \textbf{M3}~~Gist Index~\citep{leehuman}                         & \yes & \yes & LLM gist per page; index gists                    & LLM selects pages to expand                             & Text de-dup \\
\cmidrule(l{0.4em}r{0.4em}){1-7}
\famcell{Structural}
   & \textbf{M4}~~Evolving Notes~\citep{xu2025mem}                   & \yes & \yes & LLM note per write; build links                   & ANN $+$ link expansion                                  & Evolution-triggered neighbour rewrites \\
   & \textbf{M5}~~Dual-Level Graph~\citep{guo2024lightrag}           & \yes & \yes & LLM extracts entities $+$ relations; build KG     & KG $+$ vector hybrid (mix mode)                         & Lazy batch indexing; cached graph \\
\cmidrule(l{0.4em}r{0.4em}){1-7}
\famcell{Hierarchical}
   & \textbf{M6}~~Hierarchical Tree~\citep{sarthi2024raptor,sun2026h} & \yes & \yes & Embed; cluster into a four-level tree with LLM summaries & Collapsed-tree top-$k$; all levels                          & Top-level de-dup; decay \\
\cmidrule(l{0.4em}r{0.4em}){1-7}
\famcell{Refinement}
   & \textbf{M7}~~Distilled Strategies~\citep{ouyang2025reasoningbank}       & \yes & \yes & LLM judges; distills strategies; MaTTS            & Dense over strategies; top-1                            & Trajectory judge; strategy de-dup \\
   & \textbf{M8}~~Skill Bundles~\citep{zhang2026memskill}\rlap{$^{\S}$} & \yes & \yes & Cluster experiences into bundles               & Zero-shot LLM; all bundles prefilled                    & Merge and prune on overflow \\
\midrule
\classband{bandInt}{\textit{Internal Memory}} \\
\famcell{Weight}
   & \textbf{M9}~~Adapter Tuning~\citep{zhang2023lora}                & \no  & \yes & LLM QA pairs; LoRA answer-only loss               & Direct generation on base $+$ adapter                   & --- \\
\cmidrule(l{0.4em}r{0.4em}){1-7}
\famcell{Activation}
   & \textbf{M10}~~Full Context                                       & \yes & \no  & Append to buffer                                  & Concatenate; generate                                   & Truncate at window \\
   & \textbf{M11}~~Episode-Clustered Re-prefill~\citep{kim2025epicache}\rlap{$^{\ddagger}$} & \yes & \no & Cluster into episodes; keep 25\%             & Match to centroid; re-prefill episode                   & Episodic clustering; budgeting \\
\bottomrule
\end{tabularx}
\end{table*}

This appendix provides full implementation details for each of the eleven memory
substrates summarized in Section~\ref{sec:substrates} and listed in
\cref{tab:substrate}. It opens with the definition of a memory substrate and the calibration protocol shared by all substrates.

% Appendix B.1 and B.2: definition of a memory substrate, and the calibration protocol.
% Input at the top of Appendix B, before \subsection{External Substrates}.

\subsection{Definition of a Memory Substrate}
\label{app:substrate-def}

\paragraph{Definition.}
We use the term memory substrate for the storage-and-access layer of an agent harness.
It sits between the frozen backbone model and the agent controller, and
Section~\ref{sec:intro} introduces it as the infrastructure layer of the harness. A
substrate is specified by four properties.
\begin{enumerate}[leftmargin=*, itemsep=1pt]
  \item \textbf{Representation.} The form in which past experience is persisted, such as
  raw text, dense or sparse vectors, structured notes, an entity-relation graph,
  hierarchical summaries, distilled strategies or skill bundles, adapter weights, or
  context and activation state.
  \item \textbf{Write.} How an incoming interaction is encoded and committed to that
  representation.
  \item \textbf{Read.} How, given a query, the substrate surfaces stored content back into
  the model's context, or, for parametric substrates, into its weights or activations.
  \item \textbf{Management.} How stored content is consolidated, updated, deduplicated, or
  forgotten over time.
\end{enumerate}
This usage follows the write, read, and management decomposition of recent surveys and
taxonomies of agent memory~\citep{zhang2024survey, wu2025human, du2025rethinking,
du2026memoryautonomous, jiang2026anatomy, huang2026rethinking}. The partition into
external substrates, which keep information in data structures outside the model, and
internal substrates, which keep it in weights or activations, is taken from
\citet{huang2026rethinking} and is the one used in Table~\ref{tab:substrate}.

\paragraph{Scope.}
The substrate is the only component we vary. Above it, the prompt template, the
scaffolding, the final answer call, and the auxiliary LLM are held fixed. Below it, the
backbone weights are held fixed, with the qualifications that follow from the definition
itself. M9 is a weight change, so its write operation updates an adapter rather than an
external store. M10 and M11 are context or activation changes, so their read operation
fills the prompt or the KV cache rather than returning retrieved entries. This isolation
makes substrate-level effects more attributable than a comparison of complete systems
would allow. Where the evaluated method is a harness-compatible approximation of its
reference, we say so explicitly. This applies to M8 and M11, whose deviations are
described in their own paragraphs later in this appendix and noted in Appendix~\ref{app:limitations}.

\paragraph{Complementarity.}
Retrieval is a read-stage operation, and distillation is a write-stage choice about how
experience is stored. The two co-occur inside one pipeline. M7 distills trajectories into
strategies at write time and retrieves the single most relevant strategy at read time, in
the same way that M1 and M2 retrieve raw entries. What the study varies is therefore
mainly the storage and representation strategy (raw text in M1 and M2, gists in M3, notes
in M4, a graph in M5, a tree in M6, distilled strategies in M7, skill bundles in M8,
weights in M9, and context or activation state in M10 and M11), with the read interface
held fixed where applicable. Several of these mechanisms can be composed inside one system
rather than being mutually exclusive alternatives. A comparison between two rows of
Table~\ref{tab:substrate} therefore identifies which configured substrate is preferable
under an operating regime, not a universally superior replacement. Some pairs are close
substitutes, for example M1 and M2, which share a representation family and a read
interface. Comparisons across stages, for example a parametric substrate against a
retrieval index, are best read as contrasts between design choices under the shared
harness.

\paragraph{Component provenance.}
Table~\ref{tab:fidelity} records, for each substrate, where each component comes from. R
marks a component taken from the reference implementation, or from the standard baseline
for M1, M2, and M10, which have no single reference system. H marks a component
standardized by the harness. A marks an approximation of the reference mechanism, and a
dash marks a component that is not part of the method. The harness standardizes the
backbone, the prompt template, the answer call, the auxiliary LLM (\textsc{gpt-4o-mini}),
the embedding encoder (\texttt{text-embedding-3-small}), the judge, decoding, and the
agent controller loop that calls write and read and produces the answer, for every method,
so the Controller column is H in every row. Because the encoder is shared, the dense
indices of M1, M4, M5, M6, and M7 are built with the harness encoder rather than with the
encoders of their reference implementations. Two read components are approximations. M8
replaces MemSkill's PPO-trained controller with a zero-shot LLM that reads all refined
skill bundles prefilled into the prompt, while its skill-evolution pipeline, and therefore
its write and management components, follow the reference. M11 replaces EpiCache's
in-place KV-cache edits with utterance-level re-prefill of the selected episode, while its
attention-based episode scoring and its 25\% token budget follow the reference. Its
model-state entry is marked A for the same reason, since the harness fills the context
rather than the KV cache. M6 keeps
RAPTOR's tree construction but defers it to the first read call, a scheduling choice that
leaves the mechanism unchanged and moves the one-off build cost into the first query
(Table~\ref{tab:lme-s-latency}). Management is a dash for M1, M2, and M9, for which no
management operation on stored content is defined, although Appendix~\ref{app:metrics}
books the adapter training steps of M9 under the management phase for cost attribution. The marks describe the provenance of mechanisms. The
values of the settings behind them are listed in Table~\ref{tab:calibration}.

\begin{table*}[!htbp]
\centering
\caption{Provenance of each substrate component. R: reference implementation. H: standardized by the harness. A: approximated. ---: not part of the method. $^{\dagger}$: built at the first read call.}
\label{tab:fidelity}
\footnotesize
\setlength{\tabcolsep}{4pt}
\renewcommand{\arraystretch}{1.1}
\begin{tabularx}{\textwidth}{@{}l Y c c c l c@{}}
\toprule
\textbf{Method} & \textbf{Stored representation} & \textbf{Write} & \textbf{Read} & \textbf{Mgmt} & \textbf{Model state} & \textbf{Controller} \\
\midrule
\classband{bandExt}{\textit{External Memory}} \\
\textbf{M1}~Dense Vector~\citep{xiongapproximate} & 1536-d dense embeddings of entries & R & R & --- & --- & H \\
\textbf{M2}~Sparse Vector~\citep{robertson2009probabilistic} & BM25 inverted index over entries & R & R & --- & --- & H \\
\textbf{M3}~Gist Index~\citep{leehuman} & one-sentence gist per page (indexed), original pages fetched on expansion & R & R & R & --- & H \\
\textbf{M4}~Evolving Notes~\citep{xu2025mem} & notes with description, keywords, tags, and semantic links & R & R & R & --- & H \\
\textbf{M5}~Dual-Level Graph~\citep{guo2024lightrag} & entity-relation graph plus chunk vector index & R & R & R & --- & H \\
\textbf{M6}~Hierarchical Tree~\citep{sarthi2024raptor} & multi-level summary tree over raw entries & R$^{\dagger}$ & R & R & --- & H \\
\textbf{M7}~Distilled Strategies~\citep{ouyang2025reasoningbank} & distilled reasoning strategies & R & R & R & --- & H \\
\textbf{M8}~Skill Bundles~\citep{zhang2026memskill} & skill bundles & R & A & R & --- & H \\
\midrule
\classband{bandInt}{\textit{Internal Memory}} \\
\textbf{M9}~Adapter Tuning~\citep{zhang2023lora} & LoRA adapter ($r{=}8$) & R & R & --- & R, weights & H \\
\textbf{M10}~Full Context & raw history buffer & R & R & R & R, context & H \\
\textbf{M11}~Episode-Clustered Re-prefill~\citep{kim2025epicache} & episode clusters under a 25\% token budget & R & A & R & A, context (re-prefill) & H \\
\bottomrule
\end{tabularx}
\end{table*}

\paragraph{Production systems outside this frame.}
Appendix~\ref{app:excluded-substrates} reports Mem0~\citep{chhikara2025mem0},
Zep~\citep{rasmussen2025zep}, and MemGPT~\citep{packer2023memgpt} separately. Their
defining mechanisms are management or controller policies that sit outside this
single-variable frame. Mem0 arbitrates between add, update, delete, and no-op operations
over extracted facts. Zep maintains a temporal knowledge graph with fact invalidation.
MemGPT runs a paging controller over its context tiers. M3 and M5 share parts of the
storage representation with these systems but not these policies, so results for M3 and
M5 are not results for Mem0, Zep, or MemGPT.

\subsection{Calibration Protocol}
\label{app:calibration}

\paragraph{Rule.}
Two rules govern how substrates are configured. The shared experimental frame is
standardized, so that a difference between two substrates cannot come from the frame.
Every substrate-specific hyperparameter takes the value of its reference implementation,
so that no benchmark-specific tuning is done for any substrate.

\paragraph{Standardized settings.}
The frame fixes the following settings.
\begin{itemize}[leftmargin=*, itemsep=1pt]
  \item \textbf{Backbones.} \qwens, \qwenl, and \gemma, served with vLLM
  (Section~\ref{sec:setup}).
  \item \textbf{Decoding.} Greedy decoding with temperature $0.0$ and top-p $1.0$, at most
  $512$ new tokens, and seed $42$.
  \item \textbf{Auxiliary LLM.} \textsc{gpt-4o-mini} for every write-stage and management
  operation that calls an LLM.
  \item \textbf{Embedding encoder.} \texttt{text-embedding-3-small} (1536 dimensions) for
  every substrate that embeds text.
  \item \textbf{Chunking.} Inputs that arrive as a single text blob rather than as turns
  are split into 500-word chunks before they are written.
  \item \textbf{Judge.} \textsc{gpt-4o-mini} at temperature $0$ with the fixed templates of
  Appendix~\ref{app:metrics}.
\end{itemize}

\paragraph{Per-substrate settings.}
Table~\ref{tab:calibration} lists, for each substrate, the settings this paper states and
the source of each value. Harness marks a value fixed by the shared frame. Reference marks
the value of the reference implementation. Deviation marks one of the two documented
departures from a reference, the M8 read and the M11 read.

\begin{table*}[!htbp]
\centering
\caption{Substrate-specific settings stated in this paper and the source of each value (Harness: shared frame. Reference: reference implementation. Deviation: documented departure). On LoCoMo, M1, M2, M4, and M6 read $k{=}10$ entries.}
\label{tab:calibration}
\footnotesize
\setlength{\tabcolsep}{4pt}
\renewcommand{\arraystretch}{1.0}
\begin{tabularx}{\textwidth}{@{}>{\raggedright\arraybackslash}p{2.75cm} l Y l@{}}
\toprule
\textbf{Method} & \textbf{Setting} & \textbf{Value} & \textbf{Source} \\
\midrule
\rowcolor{bandExt}\multicolumn{4}{@{}l@{}}{\,\textbf{\textit{External Memory}}} \\
\textbf{M1}~Dense Vector & Encoder & \texttt{text-embedding-3-small}, cosine similarity & Harness \\
\addlinespace[2pt]
\textbf{M2}~Sparse Vector & Scoring & BM25 over tokenized entries & Reference \\
\addlinespace[2pt]
\textbf{M3}~Gist Index & Write & one gist per page (about 500 words per page) & Reference \\
 & Read & LLM selects the pages to expand & Reference \\
\addlinespace[2pt]
\textbf{M4}~Evolving Notes & Write & LLM note (description, keywords, tags) and cosine links & Reference \\
 & Read & top-$k$ notes by cosine, then link expansion & Reference \\
 & Management & evolution-triggered neighbor rewrites, consolidated every 100 evolutions & Reference \\
\addlinespace[2pt]
\textbf{M5}~Dual-Level Graph & Read & KG and vector hybrid (mix mode), reciprocal rank fusion & Reference \\
 & Management & lazy batch indexing with a cached graph & Reference \\
\addlinespace[2pt]
\textbf{M6}~Hierarchical Tree & Construction & recursive clustering and LLM summaries into a four-level tree, built at the first read call & Reference \\
 & Read & collapsed-tree top-$k$ over all levels & Reference \\
\addlinespace[2pt]
\textbf{M7}~Distilled Strategies & Write & trajectory judge and MaTTS distillation & Reference \\
 & Read & top-1 strategy by dense similarity & Reference \\
\addlinespace[2pt]
\textbf{M8}~Skill Bundles & Write and mgmt & skill evolution (clustering, summarization, merging, pruning on overflow; bank capped at 12 skills) & Reference \\
 & Read & zero-shot LLM over all prefilled bundles & Deviation \\
\midrule
\rowcolor{bandInt}\multicolumn{4}{@{}l@{}}{\,\textbf{\textit{Internal Memory}}} \\
\textbf{M9}~Adapter Tuning & Adapter & LoRA-FA, rank $r{=}8$ & Reference \\
 & Training signal & answer-only loss on auxiliary-LLM QA pairs & Reference \\
\addlinespace[2pt]
\textbf{M10}~Full Context & Management & truncation at the context window & Reference \\
\addlinespace[2pt]
\textbf{M11}~Episode-Clustered Re-prefill & Budget & 4 episodes, 25\% of utterances kept per episode & Reference \\
 & Read & utterance-level re-prefill of the matched episode & Deviation \\
\bottomrule
\end{tabularx}
\end{table*}

\paragraph{Consequence.}
Taking each hyperparameter from its reference implementation has one consequence for how
the tables should be read. Even at the same $k$ the retrieved budget differs across
substrates, because the retrieved unit is an entry for M1 and M2, a note for M4, a tree
node for M6, and a strategy for M7, and the retrieved-token metric $E_7$ records this
difference directly. Each comparison in Tables~\ref{tab:locomo-mab-main}
and~\ref{tab:alfworld-bcb-main} is therefore a comparison between configured methods as
their authors specify them, not between mechanisms at a matched budget. The $k$ sweep of
Section~\ref{sec:retrieval-scaling} is the matched-depth reading for the substrates that
expose $k$.

\paragraph{Sensitivity to retrieval depth.}
Figure~\ref{fig:abalation} is the one setting in which substrates are compared at matched
depth, and it also shows which conclusions are robust to the depth chosen. The sweep
covers LoCoMo with \qwens for M1, M2, M4, and M6 at $k\in\{1,2,5,10,20\}$ ($n{=}1986$) and
ALFWorld with \qwenl for M1, M2, M4, M6, and M7 at $k\in\{1,2,3,4,5\}$ ($n{=}134$). On
LoCoMo, $P_4$ at $k{=}20$ is above $P_4$ at $k{=}1$ for every swept substrate, and the
cross-substrate mean trend rises at every step of the sweep. On ALFWorld, the
cross-substrate mean trend stays below its $k{=}1$ value at every larger depth. TSR at $k{=}5$ is below TSR at $k{=}1$ for
M1, M2, M6, and M7, and M4 is the only swept substrate whose $k{=}5$ value is not clearly
below its $k{=}1$ value. M7 stays above every other swept substrate and above the no-memory
baseline of $22.4\%$ at every depth, falling by about seven points between $k{=}1$ and
$k{=}5$ (Section~\ref{sec:retrieval-scaling}), with its margin over the next substrate
narrowing at the larger depths. Within the sweep, the lead of M7 therefore does not depend on
the depth, and the two regimes disagree across the tested depths
rather than crossing at one setting. The per-query retrieved tokens ($E_7$, panels c and
g) differ across substrates at every $k$, which is the budget mismatch of the
previous paragraph measured directly. The paper contains no sweep over chunk granularity,
so the sensitivity of the user-centric results to the 500-word chunk size is not
characterized here.

\subsection{External Substrates}

\paragraph{M1: Dense Vector (Flat Index).}
Each incoming utterance is encoded into a 1536-dimensional embedding via OpenAI \texttt{text-embedding-3-small} and appended to a flat index. At read time, the query is embedded with the same encoder and the top-$k$ nearest neighbours are returned by cosine similarity. No LLM is invoked during write or read. No management operation is defined; the index grows monotonically.

\paragraph{M2: Sparse Vector (Flat Index).}
Each utterance is tokenized and appended to a BM25 inverted index. At read time, the query is tokenized and scored against all entries via BM25. No LLM is invoked. 
%Write latency for BM25 is amortized: lazy batch indexing ensures the full index rebuild cost is not assigned to the final turn. 
No management operation is defined.

\paragraph{M3: Gist Index (Text Record).}
Implements ReadAgent~\citep{leehuman}. At write time, an auxiliary LLM generates a one-sentence gist summary for each page of the input and indexes these gists. At read time, the same LLM reads all gists and selects which pages to expand to their full original text; the expanded pages are concatenated as context for the final answer. Management consists of text-level deduplication over stored gists. The two-stage read pipeline (gist filter then page expand) means read cost scales with the number of stored pages, which becomes expensive on long histories.
%(e.g., 477.97s per query on MAB LRU with Qwen3-32B-AWQ).

\paragraph{M4: Evolving Notes (Structural).}
Implements A-Mem~\citep{xu2025mem}. At write time, 
an auxiliary LLM generates a structured note (with contextual description, 
keywords, and tags) for each incoming message, then computes cosine 
similarity between the new note's dense embedding and existing notes to 
establish semantic links. At read time, the query embedding is matched 
against stored notes via cosine similarity, returning the top-k most 
relevant notes; the result set is further expanded by following the 
note-level links established at write time. Management is triggered by 
an evolution detector: when a new note significantly updates an existing 
one, the LLM rewrites the affected neighbours to maintain consistency.
%This evolution-triggered rewriting is the mechanism that addresses staleness hardness.

\paragraph{M5: Dual-Level Graph (Structural).}
Implements LightRAG~\citep{guo2024lightrag}. At write time, an auxiliary LLM extracts entities and relations from the input and builds a knowledge graph (KG) alongside a chunk-level vector index. At read time, queries are routed through a hybrid mode that combines KG traversal (entity and relation matching) with dense vector search over the chunk store, merging results via reciprocal rank fusion. Management uses lazy batch indexing: the KG and vector index are rebuilt periodically rather than incrementally, with a cached graph serving queries between rebuilds.

\paragraph{M6: Hierarchical Tree (Hierarchical).}
Implements RAPTOR~\citep{sarthi2024raptor}. Incoming entries are embedded at write time; on the first read call,
entries are recursively clustered and each cluster is summarised by
an auxiliary LLM to form 
the next level of the tree, with clustering and summarization repeated until
the tree has four levels. The resulting four-level tree contains
the raw entries at the leaves and progressively more abstract summaries at
higher internal nodes. At read time, we use RAPTOR's collapsed-tree variant: 
nodes from every level are flattened into a single dense index and the query 
embedding selects the top-$k$ most relevant nodes regardless of depth, so 
fine-grained leaves and coarse summaries compete for the same retrieval 
slots.
In our implementation, clustering and summarisation are deferred to
the first read call (lazy construction); \texttt{write()} only embeds
and appends entries, and the one-off build cost is therefore absorbed
by the first query (see Tab.~\ref{tab:lme-s-latency}).
Management performs top-level deduplication and temporal decay, 
removing or demoting summary nodes superseded by more recent entries.

\paragraph{M7: Distilled Strategies (Refinement).}
Implements ReasoningBank~\citep{ouyang2025reasoningbank}. At write time, an auxiliary LLM judges each completed trajectory (success or failure), then distills the judgment into a structured, reusable reasoning strategy via memory-aware test-time scaling (MaTTS). Strategies are indexed via dense embedding. At read time, the query is embedded and the most relevant strategy is retrieved. Management performs trajectory-level judging and strategy deduplication: redundant or superseded strategies are merged or pruned.

\paragraph{M8: Skill Bundles (Refinement).}
Implements MemSkill~\citep{zhang2026memskill} with one deviation. We
keep the online skill-evolution pipeline (clustering, summarisation,
merging, pruning), so M8 incurs the write- and management-side LLM
cost reported in the efficiency tables. We replace only the
PPO-trained controller that originally selects a single skill bundle
at read time with a zero-shot LLM: all refined skill bundles are
prefilled into the prompt, and the model attends to the full skill
set when generating the answer for the current query. We make this
swap because no other substrate involves RL training, and keeping M8
RL-driven would make its training cost dominate the efficiency axis.
%The skill-bundle 
% representation itself (member experiences plus a summary description) 
% is unchanged.

% \paragraph{Deviation from reference.} The original MemSkill uses a PPO-trained controller
% to select a single skill bundle at read time. We replace this learned selection with a
% zero-shot LLM that consumes all refined skill bundles prefilled into its context, so the
% model attends to the full curated skill set rather than a single controller-selected
% bundle. The skill-evolution pipeline (clustering, refinement, merging, pruning) is
% unchanged. We make this swap because no other substrate in \cref{tab:substrate} involves
% RL training, and keeping M8 RL-driven would make it the only method whose training cost
% dominates the efficiency axis, confounding the cross-substrate comparison.

\subsection{Internal Substrates}

\paragraph{M9: Adapter Tuning (Weight).}
Implements LoRA-FA~\citep{zhang2023lora}. At write time, an auxiliary LLM synthesizes question-answer pairs from the incoming content (e.g., Q: ``Where does the user live?'' A: ``San Francisco''). A LoRA adapter ($r{=}8$) is then fine-tuned on these pairs using answer-only loss on the base model. At read time, the query is passed directly to the base model with the adapter applied; no explicit retrieval step occurs. 

% No management operation is defined: knowledge encoded in weights cannot be selectively updated or deleted, which is the root cause of M9's failure on verification-heavy tasks.

\paragraph{M10: Full Context (Activation).}
The full-history concatenation baseline. At write time, each incoming message is appended
verbatim to a running buffer. At read time, the entire buffer is concatenated and passed
as context to the model for generation. Management consists solely of truncation: when
the buffer exceeds the model's context window, the oldest entries are dropped. This
substrate establishes the ``raw history retention'' reference point against which all
other substrates are compared. We omit M10 from ALFWorld evaluation because cumulative
trajectory text exceeds the context window of every backbone we test.

\paragraph{M11: Episode-Clustered Re-prefill (Activation).}
Implements EpiCache~\citep{kim2025epicache} with one deviation (marked $^{\ddagger}$ in \cref{tab:substrate}). At write time, past conversational turns are clustered into episodic groups, and a token budget of 25\% is applied (only the top quarter of tokens per episode are retained). At read time, the query is encoded and matched to episode centroids; the best-matching episode is re-prefilled into the KV cache for generation.

\textbf{Deviation from reference.} The original EpiCache edits the KV cache in place using custom attention kernels. Qwen3's hybrid attention architecture (mixing full attention and sliding-window attention across layers) is not supported by the authors' released kernels. Our implementation preserves the paper's attention-based episode scoring but applies it at the level of whole utterances: the selected episode's utterances are re-prefilled into the context rather than surgically inserted into the KV cache. This changes the read-time mechanism (re-prefill vs.\ in-place edit) but preserves the core design principle of episodic selection with budget constraints.

\subsection{Memory Substrate and Method Exclusion}
\label{app:excluded-substrates}

We considered three additional substrates, MemGPT~\citep{packer2023memgpt}, Mem0~\citep{chhikara2025mem0}, and Zep~\citep{rasmussen2025zep}, which are widely cited as production-grade memory systems but which we exclude from the main comparison because their defining mechanisms are policies outside the single-variable frame and because of their auxiliary-LLM budget. \cref{tab:excluded-cost} reports the auxiliary-LLM cost each substrate incurs on a single end-to-end LoCoMo run (ten conversations, 1,986 queries, GPT-4o-mini aux model), measured with the same harness used for every substrate in \cref{tab:substrate}. Embedding calls and embedding latency are excluded because they are served by a local FAISS index shared across runs.

\begin{table*}[t]
\centering
\renewcommand{\arraystretch}{1.0}
\setlength{\tabcolsep}{4pt}
\caption{Auxiliary-LLM cost on a full LoCoMo run (ten conversations, 1,986 queries; GPT-4o-mini aux model, embedding excluded). Costs are one to two orders of magnitude above the lightest retrieval baselines (e.g.\ M2 BM25 issues zero aux-LLM calls and finishes in roughly ten minutes).}
\label{tab:excluded-cost}
\footnotesize
\begin{tabular*}{\textwidth}{@{\extracolsep{\fill}}l l r r@{}}
\toprule
\textbf{Substrate} & \textbf{Method} & \textbf{Auxiliary LLM calls} & \textbf{Execution\ time} \\
\midrule
Text Record  & MemGPT~\citep{packer2023memgpt}  & 2,739  & 16.3\,h \\
Text Record  & Mem0~\citep{chhikara2025mem0}     & 8,984  & 11.8\,h \\
Hierarchical & Zep~\citep{rasmussen2025zep}      & 7,624  & 23.1\,h \\
\bottomrule
\end{tabular*}
\end{table*}

These three systems are LLM-call-heavy by design: every write triggers multiple auxiliary calls (entity extraction, fact deduplication, conflict resolution, temporal invalidation, agent tool dispatch), and every read fires further reranker, summary, or agent-loop calls. On a single LoCoMo run this manifests as roughly 2,700 to 9,000 auxiliary LLM calls per substrate and wall-clock budgets of 11 to 23 hours, one to two orders of magnitude above the lighter retrieval baselines.

Because the goal of our evaluation is a controlled cross-substrate ablation rather than a production-cost benchmark, we report these systems here as a production-cost reference and exclude them from the main comparison. Their defining mechanisms are policies rather than storage layers. Mem0 arbitrates every extracted fact with add, update, delete, or no-op operations, Zep maintains a temporal knowledge graph that invalidates superseded facts, and MemGPT lets the model page content between memory tiers through function calls. These policies sit outside the single-variable frame of Appendix~\ref{app:substrate-def}. Lighter substrates in \cref{tab:substrate} share parts of their storage representation (per-input text distillation in M3, entity and relation indexing in M5, a record store in M3) but not their management or control policies, so results for M3 and M5 should not be read as results for Mem0, Zep, or MemGPT. Including the three systems would also dominate the wall-clock and token-count columns by an order of magnitude and obscure the architectural differences our evaluation is designed to surface.

\section{Benchmark Setups and Bank Alignment}
\label{app:benchmarks}

This appendix expands the benchmark descriptions in Section~\ref{sec:setup} with full
capability definitions, evaluation splits, and pool construction details, and audits the
semantic alignment between the offline retrieval banks and their evaluation pools for the
two agent-centric benchmarks.

\subsection{Benchmark Setups and Capability Definitions}
\label{app:benchmark-setups}

\paragraph{LoCoMo.} The benchmark provides ten multi-session dialogues with $1{,}986$
questions across five categories, serving as an aggregate quality indicator that mixes
hardness dimensions within a single benchmark.

\paragraph{MemoryAgentBench (MAB).} MAB factors long-context memory into four capabilities:

\begin{itemize}[leftmargin=1.2em, itemsep=2pt]
  \item \textit{Accurate Retrieval (AR)}, measured on the LongMemEval-S$^\ast$
  split~\citep{wulongmemeval,hu2025evaluating}, requires the substrate to locate relevant
  information within a massive history. MAB builds this split by concatenating the
  LongMemEval chat history into five long dialogues of about $355$K tokens each and asks
  $300$ questions over them; the asterisk is MAB's label for this reformulated split.
  \item \textit{Long-Range Understanding (LRU)} tests whether the substrate preserves signal
  as content grows.
  \item \textit{Test-Time Learning (TTL)} requires the agent to classify newly acquired
  patterns, implicitly demanding that stored knowledge be inspectable, a capability that
  parametric substrates fundamentally lack.
  \item \textit{Conflict Resolution (CR)} presents contradictory facts at different time
  points and tests whether the substrate surfaces the most recent version. We additionally
  use CR as our scalability probe by sweeping content length.
\end{itemize}

\paragraph{ALFWorld.} The benchmark provides $134$ valid-unseen embodied-planning tasks
under a binary environment reward. We report a cross-episode regime in which every method
receives the same offline bank consisting of $100$ AgentGym successes~\citep{xi2025agentgym}
combined with $100$ \qwenl{} real failures (stressing recovery, as the agent must
bootstrap from a pre-built memory).

\paragraph{BigCodeBench-Hard.} The benchmark provides $148$ code tasks graded under real
\textsc{pytest} execution. Each method is populated once from an offline pool of cross-task
solutions and harvested failures, with Hard-set overlap removed before any run. Retrieved
entries are rendered as labelled \textsc{Successful} and \textsc{Failed} blocks prepended to
the official instruction prefix. BCB-Hard stresses both accumulation (encoding a diverse
solution pool without drowning retrieval signal) and staleness (distinguishing current
successes from outdated failures).

\subsection{Bank--Test Semantic Alignment for ALFWorld and BCB-Hard}
\label{app:bank-test-similarity}

A retrieval-memory benchmark is meaningful only if its offline bank is \emph{semantically
related} to its evaluation pool. If the two come from disjoint distributions, every
substrate is reduced to retrieving unrelated text and the rankings degenerate to base-model
noise. We verify here that the offline banks we use for ALFWorld and BCB-Hard both carry
strong, statistically distinguishable semantic alignment with their evaluation pools.

We embed every goal/instruction with \texttt{text-embedding-3-small} ($1536$-D,
$\ell_2$-normalized) and report the nearest-neighbor cosine similarity from each test item
to the corresponding bank. As a null baseline we replace the bank with i.i.d.\ Gaussian
unit-norm vectors and recompute nearest cosines; this is the similarity that random
retrieval would surface.

\paragraph{ALFWorld.}
The bank holds $200$ trajectories ($100$ AgentGym successes~\citep{xi2025agentgym} $+$ $100$
\qwenl{} real failures, balanced across the six task types). The evaluation pool is
the canonical $n{=}134$ \textsc{valid\_unseen} split. Nearest-bank cosine for the test goals:

\begin{itemize}[leftmargin=*,itemsep=2pt]
  \item mean $= \mathbf{0.879}$, median $= 0.881$, min $= 0.568$;
  \item top-$5$ mean cosine $= 0.761$ (the test goal is close to a \emph{neighbourhood} of
  bank items, not a single outlier);
  \item distribution: $30.6\%$ in $[0.95, 1.00]$, $35.8\%$ in $[0.85, 0.95)$, $22.4\%$ in
  $[0.75, 0.85)$, $6.0\%$ in $[0.65, 0.75)$, $5.2\%$ in $[0.00, 0.65)$.
  \item \textbf{Random-bank control: mean $= 0.068$}, an order of magnitude below the
  real bank.
\end{itemize}

Every test goal therefore has a structurally analogous example in the bank, typically the
same task type with a different target object/receptacle. The bank excludes the
\textsc{valid\_unseen} environment instances themselves, so the alignment is
template-level rather than instance-level. Substrates can borrow \emph{strategy}
(e.g.\ ``go to drawer, open it, take key, go to desk''), not the answer.

\paragraph{BigCodeBench-Hard.}
The cross-task BCB pool (cross-task solution snippets harvested from non-Hard tasks plus
failures collected during preliminary runs) provides retrieval items for $n{=}148$ Hard
instructions. The Hard-set overlap is removed before any run
(Appendix~\ref{app:benchmark-setups}).
Nearest-bank cosine on the Hard split:

\begin{itemize}[leftmargin=*,itemsep=2pt]
  \item mean $= \mathbf{0.594}$, median $= 0.586$, min $= 0.303$;
  \item top-$5$ mean cosine $= 0.526$;
  \item $25\%$ of Hard tasks have a bank example with cos $\geq 0.65$ (concrete reusable
  scaffolding such as ``download files via FTP and parse'' $\to$ ``connect to S3 and
  parse''), $40\%$ in $[0.55, 0.65)$ (shared library or pattern, e.g.\
  \texttt{requests}/\texttt{json}/\texttt{pandas} idioms), and $35\%$ below $0.55$
  (genuinely out-of-pool).
  \item \textbf{Random-bank control: mean $= 0.061$}, again an order of magnitude below
  the real bank.
\end{itemize}

BCB-Hard's lower mean cosine vs.\ ALFWorld is expected, because code instructions span a much
larger surface area (different libraries, data sources, signatures) than the templated
goals of ALFWorld, so a diverse cross-task pool can match each Hard task at most along an
abstract dimension (library family, control-flow pattern). Crucially the alignment is still
ten-times the random null and concentrated in the $[0.55, 0.85]$ regime that semantic
encoders typically reserve for ``same topic, different surface form''.

\paragraph{Summary.}
Both banks pass the basic sanity check that a memory benchmark requires. The retrieval
target distribution overlaps the evaluation distribution well above chance, so substrate
differences in Tab.~\ref{tab:alfworld-bcb-main} reflect their use of \emph{available} signal
rather than their luck against an unrelated bank. \textbf{ALFWorld} provides high-density
alignment ($0.88$ mean cosine, $30\%$ near-duplicate templates), exposing pure
strategy-transfer ability; \textbf{BCB-Hard} provides moderate, broadly distributed
alignment ($0.59$ mean cosine), stressing diverse-pool encoding and cross-task
generalisation. The two benchmarks therefore probe complementary memory regimes within the
same evaluation framework.

% Appendix C, final subsection: coverage of the memory lifecycle.
% Input at the end of Appendix/03_Alignment.tex, after the bank-alignment subsection.

% Marks for Table~\ref{tab:lifecycle}. fontawesome5 is loaded in the main preamble.
\newcommand{\lcycFull}{\faCircle}
\newcommand{\lcycHalf}{\faAdjust}
\newcommand{\lcycNone}{\faCircle[regular]}

\subsection{Coverage of the Memory Lifecycle}
\label{app:lifecycle}

The memory of a long-running agent passes through more operations than retrieval. Content
is written while the agent works, stored facts are updated or superseded, contradictions
are resolved, long inputs are consolidated, and some content is eventually forgotten or
deleted~\citep{du2025rethinking,huang2026rethinking}. The four benchmarks of
Section~\ref{sec:setup} exercise only part of this lifecycle, and the parts they exercise
differ from one benchmark to the next. This subsection states, for each benchmark, which
demands it places on the substrate and which it does not.

\paragraph{Lifecycle demands.}
Table~\ref{tab:lifecycle} rates each benchmark on eight demands. \emph{Incremental
writing} (Incr.\ write) asks whether memory is written while the task is being performed,
with reads interleaved between writes. \emph{Update} asks whether a stored fact is later
replaced by a newer value that the agent must use. \emph{Conflict} asks whether the store
holds contradictory statements that the agent must resolve. \emph{Consolidate} asks whether
the answer requires integrating content spread over a long input rather than locating a few
items. \emph{Temporal} asks whether the task requires reasoning over times, dates, or
order. \emph{Cross-session} asks whether the current query or decision depends on content
from an earlier session or episode. \emph{Past failures} asks whether memory holds failed
attempts that the agent must learn from or avoid. \emph{Forget} asks whether the benchmark
rewards removing or suppressing stored content. A filled mark denotes a demand the
benchmark was built to test, a half mark denotes a demand that is present but incidental or
only partly tested, and an empty mark denotes a demand the benchmark does not place. Where
a benchmark's original protocol and our harness differ, the mark follows the harness as
run.

\begin{table}[!htbp]
\centering
\caption{Memory-lifecycle demands exercised by each benchmark as run in the harness.
\lcycFull{} primary demand, \lcycHalf{} partial or indirect, \lcycNone{} not exercised.}
\label{tab:lifecycle}
\footnotesize
\setlength{\tabcolsep}{3pt}
\renewcommand{\arraystretch}{1.15}
\begin{tabularx}{\textwidth}{@{}Y *{8}{c}@{}}
\toprule
\textbf{Benchmark}
 & \makecell{\textbf{Incr.}\\\textbf{write}}
 & \textbf{Update}
 & \textbf{Conflict}
 & \textbf{Consolidate}
 & \textbf{Temporal}
 & \makecell{\textbf{Cross-}\\\textbf{session}}
 & \makecell{\textbf{Past}\\\textbf{failures}}
 & \textbf{Forget} \\
\midrule
\multicolumn{9}{@{}l}{\textit{User-centric}} \\
LoCoMo                  & \lcycHalf & \lcycNone & \lcycNone & \lcycHalf & \lcycFull & \lcycFull & \lcycNone & \lcycNone \\
LME-S (MAB AR)          & \lcycHalf & \lcycFull & \lcycHalf & \lcycHalf & \lcycFull & \lcycFull & \lcycNone & \lcycHalf \\
MAB LRU                 & \lcycHalf & \lcycNone & \lcycNone & \lcycFull & \lcycNone & \lcycHalf & \lcycNone & \lcycNone \\
MAB TTL                 & \lcycHalf & \lcycNone & \lcycNone & \lcycHalf & \lcycNone & \lcycHalf & \lcycNone & \lcycNone \\
MAB CR                  & \lcycHalf & \lcycFull & \lcycFull & \lcycNone & \lcycHalf & \lcycHalf & \lcycNone & \lcycHalf \\
\midrule
\multicolumn{9}{@{}l}{\textit{Agent-centric}} \\
ALFWorld                & \lcycHalf & \lcycNone & \lcycNone & \lcycHalf & \lcycNone & \lcycFull & \lcycFull & \lcycNone \\
BCB-Hard                & \lcycNone & \lcycHalf & \lcycNone & \lcycHalf & \lcycNone & \lcycHalf & \lcycFull & \lcycNone \\
\bottomrule
\end{tabularx}
\end{table}

\paragraph{User-centric benchmarks.}
In the harness every conversation or stream is written in full before its questions are
asked (write phase, then read phase, Appendix~\ref{app:metrics}), and MemoryAgentBench
specifies the same protocol, in which chunks are absorbed one by one and the questions
follow the last chunk~\citep{hu2025evaluating}. Writing is therefore incremental in arrival
order, but no read is interleaved with the writes, which is why all five user-centric rows
carry a half mark for incremental writing.

LoCoMo~\citep{maharana2024evaluating} asks $1{,}986$ questions over ten multi-session
dialogues in five categories, namely single-hop, multi-hop, temporal reasoning,
open-domain knowledge, and adversarial. The temporal category tests temporal reasoning
directly, multi-hop questions synthesize evidence from several sessions, and single-hop
questions target one session within a dialogue of up to 35 sessions, so temporal reasoning
and cross-session dependence are primary. Consolidation is partial because multi-hop
synthesis spans a few turns and LoCoMo's event-summarization task is not part of the
suite. No category targets a changed fact or a contradiction, the adversarial category
tests abstention on unanswerable questions, and nothing rewards forgetting.

LME-S inherits LongMemEval's question types~\citep{wulongmemeval}, namely information
extraction, multi-session reasoning, knowledge updates, temporal reasoning, and abstention,
and the judge rubric dispatches on these labels (Appendix~\ref{app:metrics}).
Temporal-reasoning questions make temporal reasoning primary, and multi-session reasoning
questions, which draw on several sessions of one history, make cross-session dependence
primary. Knowledge-update questions ask for the current value of a fact that changed across
sessions, so update is primary, and because the older and newer statements conflict,
contradiction handling is exercised indirectly. Multi-session reasoning questions aggregate
or compare facts across sessions, which exercises consolidation partially within histories
of roughly $355$K tokens.

The remaining three MAB capabilities are built from long inputs split into chunks of a
single stream, so their cross-session dependence is across chunks rather than across
separately delimited sessions, and we mark it partial. LRU is defined as integrating
information distributed across the entire sequence~\citep{hu2025evaluating}, which is
consolidation in the strict sense. TTL requires classifying new inputs from labeled
examples spread over the stream, so consolidation is partial, and because the examples are
correct labels rather than the agent's own errors, learning from failures is not exercised.
CR presents a fact and a later rewritten version that contradicts it and asks for the
newest version, so update and contradiction handling are primary. Recency is signaled by a
serial-number cue rather than by dates (Appendix~\ref{app:full_table}), which we count as
partial temporal reasoning. MemoryAgentBench frames this capability as revising or
overwriting stored information in the face of contradictory
evidence~\citep{hu2025evaluating}, and a substrate can satisfy it without removing
anything, so forgetting is exercised only indirectly. The scalability probe sweeps CR
content from 6K to 262K tokens (Section~\ref{sec:scalability}), which raises the cost of
locating the newest fact but does not turn the task into consolidation.

\paragraph{Agent-centric benchmarks.}
In the cross-episode ALFWorld regime~\citep{shridhar2020alfworld}, every
substrate starts from the same offline bank of $100$ AgentGym successes~\citep{xi2025agentgym}
and $100$ \qwenl{} failures, so the current episode depends on earlier episodes and half of
the bank records failures, which makes cross-session dependence and learning from failures
primary.
Trajectories completed during the run accumulate in memory (Appendix~\ref{app:full_table}),
but the bank is in place before the first episode and recovery from it is what the regime
tests, so we mark incremental writing partial. Consolidation is partial because the bank
must be encoded before use, and refinement substrates compress it into strategies or skill
bundles while flat substrates store it raw (Appendix~\ref{app:substrate-details}).

BCB-Hard~\citep{zhuobigcodebench} populates memory once from an offline pool of cross-task
solutions and harvested failures before any run, so incremental writing is not exercised.
Consolidation is partial because the pool must be encoded before use, as in the
cross-episode ALFWorld regime. Retrieved entries are rendered as labeled \textsc{Successful} and \textsc{Failed} blocks
(Appendix~\ref{app:benchmarks}), which makes learning from failures primary. We read the
staleness demand stated there, distinguishing current successes from outdated failures, as
a partial form of update, since no stored entry is ever rewritten. The $148$ tasks are
independent single-turn problems and the pool comes from other tasks, so cross-session
dependence is partial. In both agent-centric benchmarks the bank that the regime tests is
built from other runs. On cross-episode ALFWorld the agent's own completed trajectories join
the store as the run proceeds, but no task is built to depend on them.

\paragraph{What the suite does not exercise.}
Four parts of the lifecycle lie outside the suite. The first is deliberate forgetting or
deletion. No benchmark asks the agent to forget or delete on instruction, and the update
and conflict items only require that superseded facts not surface at read time, which is
why the Forget column holds no filled mark. The second is revision of memory over an
evolving long-running project. The update items in LME-S and MAB CR replace single facts
inside a fixed history, whereas a project rewrites earlier entries as its goals and
artifacts change. The third is self-correction of stored memory across many sessions, in
which the agent detects and repairs its own earlier errors. None of the benchmarks scores
such repairs. The fourth is agentic work built so that later sessions depend on memory the
agent itself wrote in earlier sessions. No benchmark in the main suite is built this way, and the supplementary MemoryArena evaluation of Appendix~\ref{app:memoryarena} covers such tasks on one backbone and two domains. The
cross-session dependence of the agent-centric rows comes from banks built by other runs, the
trajectories that accumulate during the cross-episode ALFWorld run enter memory incidentally
rather than by benchmark design, and the dependence of the user-centric rows comes from
dialogue histories written before any question is asked.

\section{Metrics Definition and Assessment}
\label{app:metrics}
This appendix expands Table~\ref{tab:metric-taxonomy} with full mathematical
definitions, evaluation protocols, and applicability notes for every
metric we report. We organize the discussion by the two families
introduced in the main text: \emph{Performance} (\S\ref{app:metrics:perf}) and
\emph{Efficiency} (\S\ref{app:metrics:eff}). Where a metric is benchmark-specific, we
state the reporting convention used by that benchmark; where a metric is
substrate-level (\textsc{All} in the table), we describe how we compute
it uniformly across the four benchmarks. Within the performance family, the compression ratio P5 and the MAB capability scores (LRU, TTL, and CR) form the memory-quality group, which we compute by one rule for every method (P5 is undefined for M9 to M11, which keep no textual store).

\begin{table*}[!tbp]
\centering
\caption{\textbf{Metric taxonomy for evaluating memory substrates.} Two families:
\textit{Performance} (P1--P6, P8--P12) captures answer accuracy and retrieval quality;
\textit{Efficiency} (E1--E15) captures latency, storage, LLM call costs, and aggregate
totals. The \textbf{Benchmark} column records which benchmarks officially report each
metric. Abbreviations: \textsc{LC} = LoCoMo, \textsc{MA} = MemoryAgentBench,
\textsc{ALF} = ALFWorld, \textsc{BC} = BigCodeBench-Hard; \textsc{All} denotes
substrate-level metrics applicable across all four. Arrows indicate the preferred
direction.}
\label{tab:metric-taxonomy}
\footnotesize
\setlength{\tabcolsep}{5pt}
\renewcommand{\arraystretch}{1.18}
\begin{tabularx}{\textwidth}{@{}l l >{\raggedright\arraybackslash}X l@{}}
\toprule
\textbf{ID} & \textbf{Metric} & \textbf{Definition} & \textbf{Benchmark} \\
\midrule
\metricband{bandExt}{\textit{Performance}: answer accuracy and retrieval quality} \\
P1\,$\uparrow$  & \textbf{Exact Match}        & $\mathbf{1}[\mathtt{norm}(\hat{y})=\mathtt{norm}(y)]$; lowercase, strip articles, punctuation, whitespace & \textsc{LC} \\
P2\,$\uparrow$  & \textbf{Token F1}           & Harmonic mean of token level precision and recall between $\mathtt{norm}(\hat{y})$ and $\mathtt{norm}(y)$ & \textsc{LC} \\
P3\,$\uparrow$  & \textbf{BLEU\,1}            & Clipped unigram precision with brevity penalty: $\mathrm{BP}\cdot\sum_w\min(c_w^{\hat{y}},c_w^{y})/|\hat{y}|$ & \textsc{LC} \\
P4\,$\uparrow$  & \textbf{LLM Judge}          & LLM labels the answer CORRECT/WRONG given $(q,y,\hat{y})$; fraction correct in $[0,1]$ & \textsc{LC},\,\textsc{MA} \\
P5  & \textbf{Compression Ratio}  & $|\text{input tokens}|/|\text{stored tokens}|$; larger values indicate more lossy compression (unsigned) & \textsc{All} \\
P6\,$\uparrow$  & \textbf{Recall@$k$}         & $|\mathcal{G}\cap\mathcal{R}|/|\mathcal{G}|$; coverage of gold evidence \texttt{dia\_id}s & \textsc{LC} \\
P8\,$\uparrow$  & \textbf{Task Success}       & Fraction of episodes where all goal conditions are satisfied (binary per episode); ALFWorld primary metric & \textsc{ALF} \\
P9\,$\uparrow$  & \textbf{GC Success}         & Fraction of individual sub goal conditions satisfied (ALFRED inherited) & \textsc{ALF} \\
P10\,$\downarrow$ & \textbf{Steps to Goal}    & Environment actions per successful episode; auxiliary efficiency signal, not officially reported by ALFWorld & \textsc{ALF}\\
P11\,$\uparrow$ & \textbf{Pass@1}             & Unit test pass rate under greedy decoding; BigCodeBench primary metric & \textsc{BC} \\
P12\,$\uparrow$ & \textbf{SubEM}              & $\mathbf{1}[\mathtt{norm}(y)\subseteq\mathtt{norm}(\hat{y})]$; substring exact match. MA primary metric for Accurate Retrieval and Conflict Resolution & \textsc{MA} \\
\midrule
\metricband{bandInt}{\textit{Efficiency}: latency, storage, LLM call costs, and aggregate totals} \\
E1\,$\downarrow$  & \textbf{Memory Size}      & Memory unit's data structures in bytes; for parametric methods (M9), includes edited weights or adapter size & \textsc{All} \\
E2\,$\downarrow$  & \textbf{Inference Time}   & Total per query wall time: \texttt{read()}\,$+$\,\texttt{generate()}; mean, p50, p90 (ms) & \textsc{All} \\
E3\,$\downarrow$  & \textbf{Write Latency}    & Per message \texttt{write()} wall time; mean, p50, p90 (ms) & \textsc{All} \\
E4\,$\downarrow$  & \textbf{Retrieval Latency}& Per query \texttt{read()} wall time; mean, p50, p90 (ms) & \textsc{All} \\
E5\,$\downarrow$  & \textbf{Write Tokens}     & LLM tokens consumed during the write phase (input\,$+$\,output) & \textsc{All} \\
E6\,$\downarrow$  & \textbf{Write Calls}      & Number of LLM calls during the write phase & \textsc{All} \\
E7               & \textbf{Retrieved Tokens}  & Tokens surfaced by memory to support answer generation & \textsc{All} \\
E8\,$\downarrow$  & \textbf{Read Tokens}      & LLM tokens consumed during read phase calls (input\,$+$\,output) & \textsc{All} \\
E9\,$\downarrow$  & \textbf{Read Calls}       & Number of LLM calls during the read phase & \textsc{All} \\
E10\,$\downarrow$ & \textbf{Mgmt.\ Tokens}    & LLM tokens for memory management (dedup, update, delete, conflict resolution) & \textsc{All} \\
E11\,$\downarrow$ & \textbf{Mgmt.\ Calls}     & Number of LLM calls for memory management & \textsc{All} \\
E12\,$\downarrow$ & \textbf{Total Tokens}     & Aggregate auxiliary-LLM tokens across phases: $E_5 + E_8 + E_{10}$ (excludes retrieved tokens $E_7$) & \textsc{All} \\
E13\,$\downarrow$ & \textbf{Total Calls}      & Aggregate LLM calls across phases: $E_6 + E_9 + E_{11}$ & \textsc{All} \\
E14\,$\downarrow$ & \textbf{Total Wall-clock} & End-to-end run time across the benchmark (s); includes write, read, and management phases & \textsc{All} \\
E15\,$\downarrow$ & \textbf{Per-query Latency}& $E_{14}/n_q$, the principal cost reported in the main tables (s/query) & \textsc{All} \\
\bottomrule
\end{tabularx}
\\[3pt]
\end{table*}

\subsection{Notation}
\label{app:metrics:notation}

Throughout we let $q$ denote a query, $y$ the gold answer, and $\hat{y}$
the predicted answer. For retrieval based substrates, we write
$\mathcal{R}=\{r_1,\ldots,r_k\}$ for the set of items returned by
\texttt{read}$(q)$ at retrieval depth $k$, and $\mathcal{G}$ for the set
of gold evidence identifiers (in LoCoMo, dialogue \texttt{dia\_id}s). The text
normalisation operator
$\mathtt{norm}(\cdot)$ lowercases the input, strips a fixed list of
articles ($\{$\texttt{a, an, the}$\}$), removes punctuation, and
collapses whitespace; this is the SQuAD style normaliser shared by
LoCoMo and MemoryAgentBench.

A \emph{conversation} (LoCoMo) or \emph{stream} (MemoryAgentBench)
generates a sequence of \texttt{write} calls followed by a batch of
\texttt{read} calls. We refer to the union of \texttt{write} calls as
the \emph{write phase} and to the per query \texttt{read}$\to$\texttt{generate}
loop as the \emph{read phase}. Substrates that perform asynchronous
maintenance (for example, A\,Mem evolution or MemSkill clustering)
accumulate this work in a third \emph{management phase}, whose LLM
tokens and calls enter E10 and E11 and whose wall-clock enters E14. The
adapter training steps of M9 are booked in the same phase. They issue
no LLM call, so they contribute to E14 and E15 only.

\subsection{Performance metrics}
\label{app:metrics:perf}

\paragraph{P1: Exact Match.}
$\textrm{EM}(\hat{y},y) = \mathbf{1}[\mathtt{norm}(\hat{y}) = \mathtt{norm}(y)]$.
We average over all queries in the benchmark and report a single scalar
in $[0,1]$. EM is the classical short answer metric; it is brittle to
paraphrase and therefore complementary to P2, P3, and P4.

\paragraph{P2: Token F1.}
Let $T(\cdot)$ tokenise on whitespace after $\mathtt{norm}$. With
$P=|T(\hat{y})\cap T(y)|/|T(\hat{y})|$ and
$R=|T(\hat{y})\cap T(y)|/|T(y)|$, F1$=2PR/(P+R)$. We use multiset
intersection so that repeated tokens contribute proportionally. Empty
$T(\hat{y})$ contributes F1$=0$.

\paragraph{P3: BLEU 1.}
$\textrm{BLEU}\,1 = \mathrm{BP}\cdot\sum_{w\in T(\hat{y})}
\min(c_w^{\hat{y}}, c_w^{y}) / |T(\hat{y})|$, where $c_w^{x}$ counts
occurrences of $w$ in $x$ and the brevity penalty is
$\mathrm{BP}=\min(1, \exp(1-|T(y)|/|T(\hat{y})|))$. P3 differs from P2
in two ways: it ignores recall, and it applies BP to discourage
truncated answers.

\paragraph{P4: LLM Judge.}
A judge model $J$ receives the query $q$, the gold answer $y$, and
the model's answer $\hat{y}$ and labels the answer as CORRECT or
WRONG. P4 reports the fraction of CORRECT labels across the
evaluation set. We use \textsc{gpt-4o-mini} at temperature $0$ with
deterministic templates that vary only by benchmark family: LoCoMo
and MemoryAgentBench use the binary rubric introduced by Mem0
\citep{chhikara2025mem0}; LongMemEval uses the original authors' per-question-type
rubric, which dispatches to one of four variants (\textsc{standard},
\textsc{temporal-reasoning}, \textsc{knowledge-update},
\textsc{abstention}) based on the question's type label
\citep{wulongmemeval}. The full templates are reproduced verbatim
in the released code repository. The judge sees only $(q, y, \hat{y})$, not the retrieved evidence or memory trace, so P4 measures answer
correctness, not retrieval grounding.

\paragraph{P5: Compression Ratio.}
$\textrm{P5} = N_{\text{input}} / N_{\text{stored}}$, where
$N_{\text{input}}$ is the total tokens passed to \texttt{write} across
the whole benchmark and $N_{\text{stored}}$ is the total tokens
materialised in the substrate's persistent store after writes
complete. Larger P5 means the substrate discards more raw text in
favour of summaries or compressed structures. P5$=1$ means lossless
caching; P5$<1$ means the substrate \emph{expands} the raw stream
(for example, when notes, graph records, or summary nodes are stored
alongside the raw entries, as for M4, M5, and M6 on LoCoMo). P5 is
unsigned: which direction is
``better'' depends on the downstream use, hence no arrow in the table.

% \paragraph{P5: Faithfulness.}
% We decompose $\hat{y}$ into atomic claims using a
% \texttt{gpt\,4o\,mini} extraction prompt (one declarative sentence per
% claim). Each claim is paired with the concatenation of $\mathcal{R}$
% and scored by a DeBERTa\,v3 large NLI model; we count a claim as
% \emph{entailed} if the entailment probability exceeds $0.5$ and is the
% arg max of \{entail, neutral, contradict\}. P5 is the per query mean
% fraction of entailed claims, averaged over the benchmark. Substrates
% with no $\mathcal{R}$ (full context, parametric) are excluded from P5
% aggregates rather than scored as $0$, since the metric is undefined.

\paragraph{P6: Recall@$k$.}
$\textrm{Recall@}k = |\mathcal{G}\cap\mathcal{R}|/|\mathcal{G}|$. For
LoCoMo we resolve $\mathcal{R}$ to gold $\texttt{dia\_id}$s by finding,
for each retrieved chunk, the smallest dialogue message span that
covers it. P6 is reported for LoCoMo only. Queries with
$|\mathcal{G}|=0$ (for example, adversarial or unanswerable items) are
excluded.

% \paragraph{P7: Precision@$k$.}
% $\textrm{Precision@}k = |\mathcal{G}\cap\mathcal{R}|/|\mathcal{R}|$.
% P7 is sensitive to over retrieval: a substrate that always returns
% $k=20$ items can match P6 at the cost of P7. We therefore report both
% and treat the operating point as a substrate design choice rather than
% a hyperparameter to maximise.

\paragraph{P8: ALFWorld Task Success.}
Per ALFWorld convention, an episode is successful iff every goal
condition specified by the PDDL goal predicate is satisfied at the
final step or earlier. P8 is the unweighted mean over the 134-episode
\texttt{valid\_unseen} test split. We follow the official evaluator
(\texttt{alfworld.agents.environment.AlfredTWEnv}) and use a
\texttt{max\_steps=50} cap.

\paragraph{P9: ALFWorld GC Success.}
For each episode, GC Success is the fraction of individual sub goal
predicates achieved (numerator), divided by the total predicates
(denominator). The episode level GC contribution can therefore be
non zero even when P8 is zero. We report the macro average over
episodes, matching ALFRED's reporting convention.

\paragraph{P10: Steps to Goal.}
For each successful episode (P8$=1$), we record the number of
environment steps; P10 is the mean over successes, with the count of
successes reported alongside for context. Failed episodes are
\emph{excluded}, not penalised, so P10 is a conditional efficiency
measure and should not be compared across substrates with very
different P8. The Steps column of Table~\ref{tab:alfworld-P-appendix}
is a different quantity, the mean number of environment steps per
episode over all 134 episodes, successes and failures alike.

\paragraph{P11: BigCodeBench Pass@1.}
We sample one greedy completion ($T=0$) per problem and execute it
against the per problem unit test suite inside the official
\texttt{bigcodebench.eval} sandbox. P11 is the fraction of problems
that pass all tests. We use the \texttt{instruct} variant of
BigCodeBench Hard and the \texttt{calibrated=False} evaluator so that
imports are the substrate's responsibility.

\paragraph{$P_{\mathrm{avg}}$: Partial success.}
Tables~\ref{tab:alfworld-bcb-main} and~\ref{tab:alfworld-P-appendix}
report $P_{\mathrm{avg}}$, the mean per-task partial-success score on
ALFWorld, next to the binary task success.

\paragraph{P12: SubEM.}
$\textrm{SubEM}(\hat{y},y) = \mathbf{1}[\mathtt{norm}(y) \subseteq
\mathtt{norm}(\hat{y})]$, where $\subseteq$ is contiguous substring
containment after normalisation. SubEM is MemoryAgentBench's primary
metric for the Accurate Retrieval (AR) and Conflict Resolution (CR)
capabilities; it is more lenient than EM for tasks where the gold is a
short fact embedded in a longer free form answer.

\subsection{Efficiency metrics}
\label{app:metrics:eff}

All wall-clock and latency measurements are taken on a single
configuration, vLLM serving on $4{\times}$H200 GPUs. Tokens are counted
with the substrate's own tokenizer (for example, the Qwen3 BPE for
Qwen3 substrates, or \texttt{cl100k\_base} for OpenAI API substrates).
Latencies are measured per event. The tables print the mean, and the
p90 where a column is marked p90; Table~\ref{tab:alfworld-bcb-main}
prints the p90 of E2 on ALFWorld and the mean on BigCodeBench-Hard.

\textbf{Online versus offline cost.} The metrics keep online inference
apart from memory construction and training. E2 covers only the online
\texttt{read}$\to$\texttt{generate} path of a query, and E3 and E4
time the write and read calls separately. E14 and E15 include every
phase, so they also carry memory construction, index building, the
auxiliary-LLM calls of the write and management phases, the one-off
tree build that M6 defers to its first query, and the adapter training
steps of M9, which we book under the management phase for cost attribution
(Appendix~\ref{app:metrics:notation}). Auxiliary-LLM calls go to the
\textsc{gpt-4o-mini} API, so they add wall-clock time without using
the local GPUs. Embedding tokens are excluded from E8.

\paragraph{E1: Memory Size.}
The disk or RAM resident bytes occupied by the substrate's data
structures after the write phase has completed. For RAG style
substrates this is the size of the vector index plus any auxiliary
metadata; for graph substrates we add the on disk DB
(\texttt{neo4j} or \texttt{kuzu}) directory size. For parametric
substrates we report the trainable parameter footprint:
the LoRA adapter size for M9. E1 is comparable across families because all
substrates eventually have to store their state on disk to support
out of process evaluation.

\paragraph{E2: Inference Time.}
Per query \texttt{read}$(q)\to$\texttt{generate} wall time, end to end,
including any context assembly. We exclude the model load time
(amortised) and any benchmark side overhead.

\paragraph{E3: Write Latency.}
Per \texttt{write} call. For substrates that batch writes
(for example, ReadAgent's gisting), we attribute the batch wall time
to the constituent calls in proportion to their input length so that
the distribution is well defined.

\paragraph{E4: Retrieval Latency.}
Per \texttt{read} call, excluding the downstream
\texttt{generate} call. For substrates with no retrieval
(full context, parametric), E4 is undefined and excluded from
aggregates.

\paragraph{E5: Write Tokens.}
Sum of input plus output tokens across all LLM calls issued during the
write phase. Embedding only substrates (M1, M2) have E5$=0$.

\paragraph{E6: Write Calls.}
Number of LLM completions issued during the write phase. We count
each chat completion request as one call regardless of streaming.

\paragraph{E7: Retrieved Tokens.}
The token count of the context block surfaced by \texttt{read} to the
generator. This is informational rather than directional (no arrow):
larger E7 buys more recall but inflates E8 downstream.

\paragraph{E8: Read Tokens.}
LLM tokens consumed during the read phase, summed across the generator
call and any reader side LLM calls (for example, ReadAgent's lookup or
RAPTOR's re ranking). Excludes embedding tokens.

\paragraph{E9: Read Calls.}
LLM call count for the read phase, by the same convention as E6.

\paragraph{E10: Management Tokens.}
LLM tokens consumed by maintenance work that is neither
write nor read attributable: A\,Mem evolution, MemSkill
re clustering, and similar. Substrates
without maintenance loops have E10$=0$.

\paragraph{E11: Management Calls.}
LLM call count for the management phase, by the same convention as E6.

% \subsection{Quality metrics}
% \label{app:metrics:qual}

% \paragraph{Q1: Compression Ratio.}
% $\textrm{Q1} = N_{\text{input}} / N_{\text{stored}}$, where
% $N_{\text{input}}$ is the total tokens passed to \texttt{write} across
% the whole benchmark and $N_{\text{stored}}$ is the total tokens
% materialised in the substrate's persistent store after writes
% complete. Larger Q1 means the substrate discards more raw text in
% favour of summaries or compressed structures. Q1$=1$ means lossless
% caching; Q1$<1$ means the substrate \emph{expands} the raw stream
% (for example, via gist annotations). Q1 is unsigned: which direction is
% ``better'' depends on the downstream use, hence no arrow in the table.

% \paragraph{Q2: Coverage Diversity.}
% We embed every stored entry with the same embedding model used by
% retrieval (\texttt{text\,embedding\,3\,small}), run K means with $k=5$
% clusters (fixed seed), and report the Shannon entropy of the cluster
% assignment distribution in nats:
% $\textrm{Q2}= -\sum_{c=1}^{5} p_c\log p_c$. Maximum is $\log 5\approx
% 1.609$. Q2 captures whether the substrate concentrates its memory on a
% narrow subset of topics (low entropy) or spreads it broadly (high
% entropy); it is computed once per substrate $\times$ benchmark.

\subsection{Aggregation and reporting conventions}
\label{app:metrics:aggregation}

\textbf{Per query versus per conversation.} P1 to P4, P6, and P12 are
per query metrics; we report the unweighted mean over all queries in
the benchmark. P8 to P11 are per episode or per problem; we report the
unweighted mean over episodes or problems. E2 to E4 latencies are
per event; the tables report the empirical mean, and the p90 of E2
where a column is marked p90.

\textbf{Sub domain breakdowns.} For LoCoMo the tables report the
aggregate over the five question types (single hop, multi hop,
temporal, open domain, adversarial). For MemoryAgentBench we report
each capability (AR, TTL, LRU, CR) separately because the capabilities
target different sub skills.

\textbf{Cross substrate comparability.} A small number of
metric$\times$substrate cells are undefined: P6 and E4 are
not defined for full context or parametric substrates (no retrieval
set); E5, E6, E10, and E11 are zero by construction for embedding only
substrates. We render undefined cells as ``--'' in the result tables
rather than imputing zeros, since a zero would bias direction aware
aggregates.

\textbf{Statistical uncertainty.} The main tables report point estimates
from single greedy runs. The sample sizes are
$n{=}1986$ for LoCoMo, $n{=}300$ for LME-S, $n{=}134$ for ALFWorld, and
$n{=}148$ for BCB-Hard, so one task is worth 0.75 TSR points on ALFWorld
and 0.68 Pass@1 points on BCB-Hard. Appendix~\ref{app:significance}
reports bootstrap confidence intervals and a paired exact McNemar test
for the ALFWorld comparison, the rerun stability of the scores, and the
reading resolution that these sample sizes give.

\section{Additional Analyses}
\label{app:analyses}
This appendix reports the statistical uncertainty of the main tables, the judge-agreement
study, the routing and portfolio analyses, and a supplementary evaluation on MemoryArena.
%% Appendix/07a_Validation.tex
%% Input at the start of the Additional Analyses appendix, before Appendix/07b_Routing_Arena.tex.
%% Two subsections: statistical significance of the agent-centric comparison, and
%% LLM-judge agreement for the P4 score.

\subsection{Statistical Significance}
\label{app:significance}

The main-text tables and the full tables of Appendix~\ref{app:full_table} report point
estimates from single greedy runs (Appendix~\ref{app:metrics}). This subsection quantifies
the uncertainty behind the agent-centric comparison that the main text builds on, where the
per-task outcome is binary and the sample sizes are among the smallest in the study, and it
states how stable the point estimates are under rerunning.

\paragraph{Method.}
For each compared substrate we compute a 95\% confidence interval for its task success rate by
bootstrapping over benchmark instances. We resample the 134 games with replacement,
recompute the task success rate on each resample, and report the resulting 95\%
bootstrap interval. To compare two substrates we run an exact McNemar test on their
paired per-game outcomes. The test conditions on the games where the two substrates
disagree, which makes it the appropriate paired test for two binary scorers evaluated
on the same instances. We test the ALFWorld-unseen cell of
Table~\ref{tab:alfworld-bcb-main} with \qwenl{} ($n{=}134$ games), which carries the
table's peak TSR.

\begin{table}[!htbp]
\centering
\caption{Bootstrap 95\% confidence intervals and the exact McNemar test for the
ALFWorld-unseen cell of Table~\ref{tab:alfworld-bcb-main}. Scores are percentages.}
\label{tab:significance}
\footnotesize
\setlength{\tabcolsep}{5pt}
\renewcommand{\arraystretch}{1.1}
\begin{tabular}{@{}l l c c@{}}
\toprule
\textbf{Cell} & \textbf{Substrate} & \textbf{TSR [95\% CI]} & \textbf{Exact McNemar $p$} \\
\midrule
ALFWorld-unseen       & M7 & 32.1 [24.6, 40.3] & \multirow{2}{*}{0.016 (M7 vs.\ M2)} \\
\qwenl, $n{=}134$     & M2 & 21.6 [14.9, 29.1] & \\
\bottomrule
\end{tabular}
\end{table}

\paragraph{Results.}
The comparison that the main text builds on is significant. On ALFWorld-unseen with
\qwenl, M7 reaches a TSR of 32.1 [24.6, 40.3] against 21.6 [14.9, 29.1] for M2, and the
exact McNemar test on the 134 paired outcomes gives $p{=}0.016$.

\paragraph{Rerun stability.}
With generation at temperature 0, reruns reproduce M2 and M10 bit for bit, and the
residual variation of the remaining substrates between reruns is at most 0.02 on the
$[0,1]$ scale. The best and worst substrates in every column of
Table~\ref{tab:locomo-mab-main} are separated by 0.1 or more, so that separation is not
decoding noise. Orderings among substrates whose scores lie within 0.02 of each other, such
as M1 and M2 on LME-S with \qwens{} (0.478 against 0.480), are within the rerun variation
and should not be read as rankings.

\paragraph{Sample sizes.}
Long-horizon agentic tasks are expensive to evaluate. Every ALFWorld game is a
multi-step episode under a 50-step cap and every BCB-Hard task executes a
unit-test suite, so evaluation sets with hundreds rather than thousands of instances are
the norm, and we use the full standard splits rather than subsamples. The 134-game
\texttt{valid\_unseen} split of ALFWorld~\citep{shridhar2020alfworld} is the evaluation
set used by ReAct~\citep{yao2023react}, Reflexion~\citep{shinn2023reflexion}, and
ElasticMem~\citep{feng2026elasticmem}. The 148-task BCB-Hard
subset~\citep{zhuobigcodebench} is the split used by
\citet{sharifloo2025struggle}, CodeBoost~\citep{wang2025codeboost}, and
DAJ~\citep{qin2026daj}. At these sizes one task corresponds to 0.75 TSR points on
ALFWorld and 0.68 Pass@1 points on BCB-Hard, which is the resolution at which a single
agent-centric cell of Table~\ref{tab:alfworld-bcb-main} can be read. Gaps of a few points
on either benchmark amount to a handful of tasks. The intervals in Table~\ref{tab:significance} extend roughly seven to eight points on
either side of the estimate. Of the ALFWorld pairs we report only M7 against M2. The
4.5-point gap between M7 and M1 (27.6) lies inside these interval widths, so we make no
claim that M7 beats M1. The sample sizes of the LoCoMo and LME-S
cells are listed in Appendix~\ref{app:metrics}. The retrieval-depth sweep of
Section~\ref{sec:retrieval-scaling} and the content-length sweep of
Section~\ref{sec:scalability} are likewise read from point estimates.

\subsection{LLM-Judge Agreement}
\label{app:judge}

$P_4$, the LLM-judge score of Table~\ref{tab:locomo-mab-main}, is produced by
\textsc{gpt-4o-mini} under the templates described in Appendix~\ref{app:metrics}. This
subsection measures how much of $P_4$ depends on that particular judge.

\paragraph{Protocol.}
We draw 595 predictions from LoCoMo, LME-S, and MAB CR, balanced between the two
labels CORRECT and WRONG, and re-score every prediction with fresh calls to four
judges, namely \textsc{gpt-4o-mini} itself, \textsc{gpt-4.1-mini}, \textsc{gpt-4.1-nano}, and
\textsc{gpt-5-nano}. Each judge sees only the question, the gold answer, and the
prediction, never the retrieved evidence or the identity of the substrate that produced
the prediction. Agreement is measured against the production \textsc{gpt-4o-mini}
labels that enter Table~\ref{tab:locomo-mab-main}. Because the first judge is the
production judge re-run with fresh calls rather than re-read from cached labels, its row
measures test-retest stability, and the other three rows measure agreement across model
generations.

Table~\ref{tab:judge} reports the per-judge agreement and Cohen's $\kappa$.

\begin{table}[!htbp]
\centering
\caption{Agreement of four judges with the production \textsc{gpt-4o-mini} labels on
595 label-balanced predictions from LoCoMo, LME-S, and MAB CR.}
\label{tab:judge}
\small
\setlength{\tabcolsep}{8pt}
\renewcommand{\arraystretch}{1.1}
\begin{tabular}{@{}l l c c@{}}
\toprule
\textbf{Judge} & \textbf{Family} & \textbf{Agreement} & \textbf{Cohen's $\kappa$} \\
\midrule
\textsc{gpt-4o-mini} (fresh calls) & GPT-4o  & 95.8\% & 0.89 \\
\textsc{gpt-4.1-mini}              & GPT-4.1 & 92.6\% & 0.85 \\
\textsc{gpt-4.1-nano}              & GPT-4.1 & 91.0\% & 0.82 \\
\textsc{gpt-5-nano}                & GPT-5   & 83.6\% & 0.67 \\
\bottomrule
\end{tabular}
\end{table}

\paragraph{Results.}
The fresh \textsc{gpt-4o-mini} run agrees with its own production labels on 95.8\% of
the items (Cohen's $\kappa{=}0.89$), so the $P_4$ score is stable under re-querying. The
two GPT-4.1 judges agree with the production labels on 92.6\% and 91.0\% of the items
($\kappa{=}0.85$ and $0.82$), and \textsc{gpt-5-nano} on 83.6\% ($\kappa{=}0.67$).
Across all four judges, Fleiss' $\kappa$ is 0.73 and the judges are unanimous on 75\%
of the items. By benchmark, the four-judge $\kappa$ is 0.89 on LME-S, 0.76 on MAB CR,
and 0.68 on LoCoMo. Because the sample is label-balanced rather than drawn per
substrate, we did not recompute substrate rankings under the alternative judges.

\paragraph{Reading the values.}
On the scale of \citet{landis1977measurement}, $\kappa$ between 0.61
and 0.80 is substantial and $\kappa$ above 0.80 is almost perfect, so the three
GPT-4-series judges fall in the almost-perfect band while \textsc{gpt-5-nano} and
the four-judge Fleiss' $\kappa$ fall in the substantial band. The values sit at or
above the agreement levels reported for LLM judges elsewhere. On MT-Bench, GPT-4 agrees
with human judges on 85\% of the non-tie pairwise votes, where human judges agree with
one another on 81\%~\citep{zheng2023judging}. Judging the Judges~\citep{thakur2025judging}
measures alignment with Scott's $\pi$ rather than Cohen's $\kappa$ and finds that only
the strongest judges (GPT-4 Turbo, Llama-3 70B, and Llama-3.1 70B) reach $\pi$ of 0.87 to
0.88 against human annotators, who reach 0.96 among themselves. Those studies measure
judge-human agreement on other tasks, whereas ours measures judge-judge agreement on
our own predictions, so the comparison locates our values rather than certifying them.

\paragraph{Evaluation coupling.}
\textsc{gpt-4o-mini} is also the auxiliary LLM for memory writing and management in
every substrate that needs one (Section~\ref{sec:setup}), so the production judge and
the memory writer are the same model. The three other judges come from later model
generations and agree substantially with it, which limits how much of $P_4$ can be
specific to that one judge, and the appendix tables report judge-free scores ($P_1$, $P_2$,
$P_3$, and $P_{12}$) next to $P_4$, while the agent-centric benchmarks are scored by environment
reward and unit-test execution without any judge. All four judges, however, come from a
single provider, so a human or cross-provider check of $P_4$ remains outside this study.

%% Appendix/07b_Routing_Arena.tex
%% Substrate routing and multi-substrate portfolios.
%% Input in the Additional Analyses appendix, right after Appendix/07a_Validation.tex.

\subsection{Substrate Routing}
\label{app:routing}

The regime-dependent reversals of Section~\ref{sec:result} and
Section~\ref{sec:discussion} make substrate selection a design axis in its own
right. This subsection asks whether the selection can be automated per query. By a
router we mean a dispatcher placed in front of a pool of candidate substrates.
When a query arrives, the router decides which substrate serves it, and that
substrate alone produces the answer. A router is therefore a selection policy and
not a new memory mechanism, and it is deployable only under two conditions. It may
read only what exists when the query arrives, namely the query itself and the
fitted policy, never the gold annotation and never which substrate turns out to be
right. It must also be judged against the alternative a practitioner would
otherwise choose, which is deploying the single best fixed substrate on the same
held-out queries. The quantity we report is accordingly the accuracy of the routed
system, not the accuracy of the routing decision.

\paragraph{Setup.}
All routers are evaluated on LoCoMo with \qwens, the setting of
Table~\ref{tab:locomo-mab-main}, over its $1{,}986$ questions, with the substrates
of that table as the candidate pool. We split the questions
into five folds, fit every router on the training folds, score it on the held-out
fold, and read the fixed baseline off the same fold, so router and baseline are
always compared on identical questions. Per-question correctness is the $P_4$ judge
label of the paper runs (Appendix~\ref{app:metrics}). We evaluate four routers.
\begin{itemize}[leftmargin=1.2em, itemsep=2pt]
  \item \textit{Multi-class.} One classifier over the whole substrate set,
  trained to predict which substrate answers a given question correctly. We run
  it twice, once over TF-IDF features and once over dense query embeddings.
  \item \textit{Nearest neighbors.} No fitted model. The query is embedded, its
  25 most similar training questions are retrieved, and the query is sent to the
  substrate that answered those neighbors best.
  \item \textit{Per-substrate outcome.} One binary classifier per substrate, each
  trained to predict whether that substrate will answer the question correctly.
  At inference the query goes to the substrate with the highest predicted
  probability. Supervision is dense, since every substrate contributes a label on
  every training question, and no tie between two correct substrates has to be
  broken.
  \item \textit{Predicted question type.} A classifier predicts the LoCoMo
  question category from the query and routes to the substrate that wins that
  category on the training folds.
\end{itemize}
Two reference rows replace the predicted category with the gold category, and
the whole policy with a per-instance oracle that picks a correct substrate
whenever one exists. Neither is observable at inference.
Table~\ref{tab:router} reports the held-out accuracy of every router next to the
best fixed substrate.

\begin{table}[!htbp]
\centering
\caption{Query-level routing on LoCoMo with \qwens ($1{,}986$ questions, five
folds). Held-out $P_4$ of the routed system and its difference from the best fixed substrate.}
\label{tab:router}
\footnotesize
\setlength{\tabcolsep}{5pt}
\renewcommand{\arraystretch}{1.05}
\begin{tabular}{@{}l l l c c@{}}
\toprule
\textbf{Router} & \textbf{Prediction target} & \textbf{Features} & \textbf{Held-out $P_4$} & \textbf{vs.\ best fixed} \\
\midrule
Best fixed substrate (M5)        & none                        & none        & 0.648 & --- \\
Multi-class                      & winning substrate           & TF-IDF      & 0.591 & $-0.057$ \\
Multi-class                      & winning substrate           & embeddings  & 0.588 & $-0.060$ \\
Nearest neighbors (25)          & none                        & embeddings  & 0.650 & $+0.002$ \\
Per-substrate outcome            & correctness of each substrate & embeddings & \textbf{0.671} & $\mathbf{+0.023}$ \\
Predicted question type          & question category           & embeddings  & 0.613 & $-0.035$ \\
\midrule
Gold question type (reference)   & given, not predicted        & none        & 0.689 & $+0.041$ \\
Per-instance oracle (reference)  & given, not predicted        & none        & 0.93 & $+0.28$ \\
\bottomrule
\end{tabular}
\end{table}

\paragraph{Results.}
The per-substrate router is the only router that clears the fixed baseline by a
margin, 0.671 against 0.648, and it does so in all five folds. On paired
per-question outcomes it is correct where the fixed choice is wrong 223 times
against 179 the other way, which gives an exact McNemar $p{=}0.032$ and a paired
bootstrap 95\% interval on the difference of $[+0.003, +0.042]$. The
nearest-neighbor router matches the baseline (0.650) and the remaining routers
fall below it. The prediction target matters more than the features. Holding the
target fixed and swapping TF-IDF for embeddings moves the multi-class score from
0.591 to 0.588, while holding the features fixed and changing the target from
naming the winner to asking each substrate whether it will answer moves the score
from 0.588 to 0.671. The multi-class target keeps one substrate per question and
discards the outcomes of all others, and it forces an arbitrary choice whenever
several substrates answer correctly. The per-substrate target uses every outcome
and never breaks a tie. The multi-class result is therefore a negative result
about that target, not about query-level routing. The winning router needs only
the question embedding, one embedding call per query with no LLM call and no
annotation.

\paragraph{Regime-level routing.}
A deployed service knows whether it answers long-context questions or drives an
agent loop, so routing on the operating regime costs nothing to observe. Because the
per-regime winners differ in Tables~\ref{tab:locomo-mab-main}
and~\ref{tab:alfworld-bcb-main}, choosing one substrate per regime beats any single fixed
substrate on our benchmarks. The choice is read off the same results, so this is an
in-sample comparison rather than a held-out test.
Within LoCoMo, routing on the annotated question category reaches 0.689 against
0.648 (gold question type row of Table~\ref{tab:router}), but the annotation is
not available at inference, so this row bounds what a perfect type predictor
could deliver rather than describing a deployable router.

\paragraph{Selection.}
Everything is fitted inside the training folds. No hyperparameter is tuned on any
fold. The classifiers are logistic regression at library defaults, and the 25 neighbors
of the nearest-neighbor router were fixed before any result was seen.
Re-selecting the best fixed substrate inside each training fold picks the same
substrate (M5) in every fold and leaves the baseline unchanged to three decimals.

\paragraph{Split protocol.}
The folds above are a random partition of the $1{,}986$ questions under a fixed
seed and are not grouped by conversation, so questions from the same conversation
can appear in both the training and the held-out folds. LoCoMo has ten
conversations (Section~\ref{sec:setup}). Table~\ref{tab:router-protocols} repeats
the comparison with conversation-grouped folds and with leave-one-conversation-out
splits, where the held-out questions always come from conversations the router
has never seen.

\begin{table}[!htbp]
\centering
\caption{Per-substrate router against the best fixed substrate (M5) under three split
protocols on LoCoMo with \qwens, with the exact McNemar $p$ and the folds the router wins.}
\label{tab:router-protocols}
\footnotesize
\setlength{\tabcolsep}{6pt}
\renewcommand{\arraystretch}{1.05}
\begin{tabular}{@{}l c c c l@{}}
\toprule
\textbf{Protocol} & \textbf{Router} & \textbf{Best fixed} & \textbf{Difference} & \textbf{Test} \\
\midrule
Question-level, random folds (reported) & 0.671 & 0.648 & $+0.023$ & $p{=}0.032$, 5 of 5 folds \\
Conversation-grouped, 5 folds           & 0.661 & 0.648 & $+0.013$ & $p{=}0.22$, 3 of 5 folds \\
Leave-one-conversation-out, 10 folds    & 0.649 & 0.648 & $+0.001$ & $p{=}0.96$, 5 of 10 folds \\
\bottomrule
\end{tabular}
\end{table}

The gain shrinks to $+0.013$ under conversation-grouped folds and disappears under
leave-one-conversation-out, and neither difference is significant. The gain is
therefore specific to the question-level protocol, and we narrow the claim
accordingly. Routing helps a deployment that has accumulated outcomes over its own
traffic and routes new questions from the same conversations. We do not claim
transfer to conversations the router has never seen.

\paragraph{Summary.}
Four points follow. Regime-dependent reversals make substrate selection a real
design axis. Choosing a substrate per observable regime gains in-sample. A router that uses
only the question embedding beats the best fixed substrate under the
question-level protocol at low inference cost. Transfer to unseen conversations
and closing the gap to the per-instance oracle (0.93) remain open.

\subsection{Multi-Substrate Portfolios}
\label{app:portfolio}

Routing picks one substrate per query from a pool. A complementary question is
how large the pool has to be. We run a small portfolio of substrates side by side
and ask what an ideal per-query selector over that portfolio would achieve. A
portfolio covers an instance when at least one member solves it, so each coverage
value below is exactly the accuracy an ideal per-instance selector over that
portfolio would reach. Coverage is computed from the per-instance outcomes of the
paper runs, namely the per-query $P_4$ judge labels on LoCoMo, LME-S, and MAB CR,
per-task success on ALFWorld, and per-task Pass@1 on BCB-Hard, with each
substrate's run matched to its cell in Table~\ref{tab:locomo-mab-main} or
Table~\ref{tab:alfworld-bcb-main}. Portfolios are picked by greedy set cover,
which coincides with the exhaustive optimum at sizes two and three on every
benchmark. Table~\ref{tab:portfolio} lists the best single substrate, the best
portfolios of size two and three, and the coverage of the full substrate pool.

\begin{table}[!htbp]
\centering
\caption{Coverage of substrate portfolios, equal to the accuracy of an ideal per-instance
selector over the portfolio. Units follow Tables~\ref{tab:locomo-mab-main} and~\ref{tab:alfworld-bcb-main}.}
\label{tab:portfolio}
\footnotesize
\setlength{\tabcolsep}{5pt}
\renewcommand{\arraystretch}{1.05}
\begin{tabular}{@{}l l l l l c@{}}
\toprule
\textbf{Benchmark} ($n$) & \textbf{Backbone} & \textbf{Best single} & \textbf{Size 2} & \textbf{Size 3} & \textbf{All substrates} \\
\midrule
LoCoMo (1986)   & \qwens      & M5 0.648  & \{M10, M4\} 0.746 & \{M10, M4, M9\} 0.825 & 0.93 \\
LME-S (300)     & \qwens      & M5 0.537  & \{M5, M8\} 0.660  & \{M2, M5, M8\} 0.700  & 0.743 \\
MAB CR (100)    & \qwens      & M5 0.440  & \{M5, M6\} 0.560  & \{M5, M6, M1\} 0.580  & 0.580 \\
ALFWorld (134)  & \qwens      & M11 11.9  & \{M11, M4\} 19.4  & \{M11, M4, M8\} 23.9  & 34.3 \\
BCB-Hard (148)  & \gemmashort & M5 20.9   & \{M5, M8\} 28.4   & \{M4, M5, M8\} 29.7   & 29.7 \\
\bottomrule
\end{tabular}
\end{table}

Small heterogeneous portfolios suffice everywhere. Two substrates add between
0.098 and 0.123 of coverage over the best single substrate on the three dialogue
benchmarks and 7.5 points on each agentic benchmark. Three substrates close the
whole gap between the best single substrate and the full pool on BCB-Hard and
MAB CR, 79\% of it on LME-S, 63\% on LoCoMo, and 54\% on ALFWorld. The members
mix families. Every size-two portfolio pairs two different families, and four of
the five size-three portfolios span three families, so heterogeneity rather than
standalone strength drives the complementarity. M4, which is never the per-model
performance winner in Table~\ref{tab:locomo-mab-main} or
Table~\ref{tab:alfworld-bcb-main}, enters the LoCoMo, ALFWorld, and BCB-Hard
portfolios because it solves instances the leaders miss.

Coverage is an upper bound that assumes a perfect selector, and
Appendix~\ref{app:routing} measures how much of it a deployable router recovers.

%% Appendix/07c_MemoryArena.tex
%% Supplementary evaluation on MemoryArena (formal reasoning in mathematics and group travel planning).

\subsection{Supplementary Evaluation on MemoryArena}
\label{app:memoryarena}

MemoryArena~\citep{he2026memoryarena} builds agentic tasks from ordered subtasks that span
several sessions, so later subtasks depend on information gathered in earlier ones. This is one of the four parts of the memory lifecycle that our main benchmarks do not exercise (Appendix~\ref{app:lifecycle}). We run every substrate and the benchmark's own BM25 memory
baseline on two of its domains, formal reasoning in mathematics (40 tasks) and group travel planning (270 tasks). The baseline is MemoryArena's own BM25 memory, whereas M2 is our BM25 substrate behind the benchmark's memory interface (Appendix~\ref{app:substrate-details}), so the two differ in how entries are written and retrieved. The task agent is \qwens{} at temperature 0, the benchmark
environments use GPT-4.1-mini with step limits of 10 and 30, and the auxiliary LLM and the
embedding encoder are those of the main experiments, so the substrate is the only change across runs apart from a larger completion budget for M9 on travel planning ($4{,}096$ tokens against $1{,}024$). Memory is written session by session. On travel planning each substrate also keeps the first $4{,}000$ characters of the opening observation of a task, because the shared trip specification appears only
there, and the injected memory is capped at $20{,}000$ characters. We report the benchmark's
metrics. The success rate (SR) counts tasks whose final subtask is solved, the progress score
(PS) is the fraction of subtasks solved, and the soft progress score (sPS) also credits
partially satisfied constraints. Following the benchmark, we use sPS for travel planning, where
PS is near zero for every method.

\begin{table}[!htbp]
\centering
\caption{Supplementary evaluation on MemoryArena with the \qwens agent. Formal
reasoning in mathematics reports SR and PS over 40 tasks, and group travel planning reports
sPS (\%) over 270 tasks. BM25 is the benchmark's own memory baseline. Bold marks the best value per column.}
\label{tab:memoryarena}
\footnotesize
\setlength{\tabcolsep}{6pt}
\renewcommand{\arraystretch}{1.05}
\begin{tabular}{@{}l c c c@{}}
\toprule
 & \multicolumn{2}{c}{\textbf{Mathematics}} & \textbf{Travel planning} \\
\cmidrule(lr){2-3}\cmidrule(l){4-4}
\textbf{Substrate} & \textbf{SR} & \textbf{PS} & \textbf{sPS (\%)} \\
\midrule
BM25 (benchmark baseline) & 0.150 & 0.175 & 3.82 \\
\midrule
\textbf{M1}~Dense Vector & 0.075 & 0.092 & \textbf{3.99} \\
\textbf{M2}~Sparse Vector & 0.000 & 0.094 & 3.51 \\
\textbf{M3}~Gist Index & 0.050 & 0.124 & 3.56 \\
\textbf{M4}~Evolving Notes & 0.200 & 0.170 & 3.29 \\
\textbf{M5}~Dual-Level Graph & 0.075 & 0.157 & 2.58 \\
\textbf{M6}~Hierarchical Tree & 0.050 & 0.163 & 3.15 \\
\textbf{M7}~Distilled Strategies & 0.050 & 0.131 & 2.76 \\
\textbf{M8}~Skill Bundles & 0.050 & 0.156 & 3.12 \\
\textbf{M9}~Adapter Tuning & 0.100 & 0.162 & 3.69 \\
\textbf{M10}~Full Context & \textbf{0.225} & \textbf{0.176} & 3.45 \\
\textbf{M11}~Episode-Clustered Re-prefill & 0.075 & 0.151 & 2.93 \\
\bottomrule
\end{tabular}
\end{table}

All scores in Table~\ref{tab:memoryarena} are low. On mathematics the full-context M10 and the structural M4 solve the most
tasks, 9 and 8 of 40 against 6 for the BM25 baseline, while their progress scores (0.176 and
0.170) stay level with the baseline (0.175). One task moves SR by 0.025, so we read these
rankings as descriptive. On travel planning every method stays between 2.6\% and 4.0\% sPS, so
this domain does not separate the substrates for an agent of this size. Interdependent
multi-session tasks therefore remain hard for every substrate family at this scale.

\section{Full Tables}
\label{app:full_table}

\subsection{Full Tables for the LoCoMo Benchmark}
\label{app:locomo-deep-analysis}

\paragraph{Substrate family vs.\ accuracy and cost.} The three
appendix tables sort the ten substrates into a clear ordering by
family. The Flat family (M1, M2) requires no auxiliary
LLM at any phase, stores the raw utterance set ($P_5{=}1.00$), and
lands at the cheap end of the curve, with M2 at 0.33\,s/query and M1
at 0.77\,s/query carrying the dense encoder advantage of $+0.09$ $P_4$
on \qwens{}. The Text and Structural families (M3, M4, M5) all push
auxiliary LLM tokens into the pipeline, but where they spend that
budget matters more than how much. M3 concentrates LLM work in the
per-query page-select call ($E_8{=}8.2$M, one call per query) and
recovers M1-level $P_4$ with a $5.3\times$ store compression. M4 spends
mostly on note generation at write ($E_5{=}4.6$M) and on the
evolution loop at management ($E_{10}{=}3.2$M), giving one of the smallest
stores in the family (178K) at the cost of one of the lowest $P_6$
Recall@$k$ values (0.465).
M5 is the most LLM-intensive substrate by a wide margin
($E_{12}{=}51$M, $E_{15}{=}28$\,s/query on \qwens{}) yet posts the top
$P_4$ of the \qwens{} band (0.648)
by raising the LLM-judge score, not retrieval recall. The
Hierarchical and Refinement families invert this trade-off. M6
pays modest write-time tree summarization ($E_5{=}382$K) and
zero read-phase LLM yet lands the highest $P_6$ Recall@$k$ in the table (0.797),
because the collapsed tree exposes nodes at every layer for the
embedding search. M8 sits between M6 and M5 in cost but
trails M6 by 4 to 5 $P_4$ points because prefilling every skill bundle
dilutes the model's attention over the one relevant to the query.

\paragraph{Backbone dependence and internal-memory families.} The
relative ordering on $P_4$ is mostly preserved across the three
backbones, and the leading substrate's margin over the flat baselines
is robust to model size. M5's $+0.09$ $P_4$ lead over M1 on \qwens{} holds
at $+0.10$ on \qwenl{} and widens to $+0.15$ on \gemmashort{}; M3's parity with
M1 holds across all three; M2 trails M1 on every backbone. The
internal-memory families illustrate the limits of backbone-only
solutions. The Weight family (M9) collapses on LoCoMo
regardless of host ($P_1$ EM~$<$~0.025). Single-pass adapter tuning
over QA pairs is not enough for the recall-style questions LoCoMo
asks. The Activation family (M10, M11) shows the
sharpest backbone dependence in the table. M10 reaches the highest $P_4$
among the Activation family on \qwenl{} (0.669), just behind structural
M5 (0.683), by paying about $6\times$ the per-query latency of M1. Its
gain over M1 narrows to $+0.03$ on \qwens{}, where the smaller model
under-utilizes long contexts, and widens to $+0.11$ on \gemmashort{} at
$5.6\times$ the latency of M1.
M11 is the only substrate that exhibits clear inverse scaling,
with $P_4$ dropping from 0.307 on \qwens{} to 0.274 on \qwenl{},
consistent with the episode clustering selecting a less relevant
episode on the larger backbone. We did not isolate the cause.
Together, these results suggest that the recall-heavy
structure of LoCoMo rewards retrieval substrates that pair a
faithful textual store ($P_5$ near 1) with broad retrieval coverage,
and that auxiliary LLM cost translates into
accuracy only when the LLM work targets the actual retrieval
bottleneck rather than synthesis quality on top of it.

%% =====================================================================
%% Appendix Table — LoCoMo Performance (P1..P4, P6)
%% =====================================================================
\begin{table*}[!t]
\centering
\caption{LoCoMo performance metrics per substrate and backbone ($P_1$ to $P_4$, $P_6$).}
\label{tab:locomo-P-appendix}
\renewcommand{\arraystretch}{1.0}
\setlength{\tabcolsep}{4pt}
\footnotesize
\begin{tabular}{@{}c l c *{5}{c}@{}}
\toprule
\textbf{Model} & \textbf{Sub.} & \textbf{Family}
& $P_1$ EM & $P_2$ F1 & $P_3$ BLEU-1 & $P_4$ Judge & $P_6$ Recall@$k$ \\
\midrule
\multirow{10}{*}{\textsc{Qwen3-8B}}
 & M1  & Flat   & 0.273 & 0.466 & 0.285 & 0.561 & 0.696 \\
 & M2  & Flat   & 0.283 & 0.436 & 0.241 & 0.470 & 0.595 \\
 & M3  & Text   & 0.190 & 0.366 & 0.269 & 0.562 & 0.620 \\
 & M4  & Struct & 0.226 & 0.435 & 0.216 & 0.545 & 0.465 \\
 & M5  & Struct & 0.196 & 0.381 & 0.252 & 0.648 & 0.455 \\
 & M6  & Hier   & 0.251 & 0.436 & 0.276 & 0.556 & 0.797 \\
 & M8  & Refine & 0.301 & 0.456 & 0.252 & 0.509 & 0.615 \\
 & M9  & Weight & 0.019 & 0.099 & 0.071 & 0.379 & --    \\
 & M10 & Act    & 0.159 & 0.342 & 0.285 & 0.589 & --    \\
 & M11 & Act    & 0.066 & 0.161 & 0.138 & 0.307 & --    \\
\midrule
\multirow{10}{*}{\textsc{Qwen3-32B-AWQ}}
 & M1  & Flat   & 0.208 & 0.408 & 0.328 & 0.582 & 0.696 \\
 & M2  & Flat   & 0.187 & 0.352 & 0.272 & 0.487 & 0.595 \\
 & M3  & Text   & 0.173 & 0.368 & 0.289 & 0.586 & 0.620 \\
 & M4  & Struct & 0.165 & 0.320 & 0.240 & 0.452 & 0.465 \\
 & M5  & Struct & 0.134 & 0.329 & 0.286 & 0.683 & 0.455 \\
 & M6  & Hier   & 0.192 & 0.387 & 0.307 & 0.573 & 0.797 \\
 & M8  & Refine & 0.190 & 0.365 & 0.281 & 0.533 & 0.615 \\
 & M9  & Weight & 0.024 & 0.101 & 0.082 & 0.399 & --    \\
 & M10 & Act    & 0.190 & 0.412 & 0.337 & 0.669 & --    \\
 & M11 & Act    & 0.035 & 0.118 & 0.095 & 0.274 & --    \\
\midrule
\multirow{9}{*}{\textsc{Gemma-4-26B}}
 & M1  & Flat   & 0.329 & 0.513 & 0.311 & 0.574 & 0.696 \\
 & M2  & Flat   & 0.302 & 0.451 & 0.257 & 0.555 & 0.595 \\
 & M3  & Text   & 0.303 & 0.484 & 0.300 & 0.599 & 0.620 \\
 & M4  & Struct & 0.310 & 0.405 & 0.232 & 0.476 & 0.465 \\
 & M5  & Struct & 0.203 & 0.451 & 0.271 & 0.719 & 0.455 \\
 & M6  & Hier   & 0.310 & 0.483 & 0.292 & 0.592 & 0.797 \\
 & M8  & Refine & 0.335 & 0.482 & 0.241 & 0.554 & 0.615 \\
 & M10 & Act    & 0.300 & 0.515 & 0.308 & 0.688 & --    \\
 & M11 & Act    & 0.085 & 0.181 & 0.149 & 0.330 & --    \\
\bottomrule
\end{tabular}
\end{table*}

%% =====================================================================
%% Appendix Table — LoCoMo Latency & Compression
%% =====================================================================
\begin{table*}[!t]
\centering
\caption{LoCoMo latency and compression per substrate and backbone ($E_2$ to $E_4$, $E_{14}$, $E_{15}$, $P_5$).}
\label{tab:locomo-latency-Q-appendix}
\renewcommand{\arraystretch}{1.0}
\setlength{\tabcolsep}{4pt}
\footnotesize
\begin{tabular}{@{}c l c *{5}{r} r@{}}
\toprule
\textbf{Model} & \textbf{Sub.} & \textbf{Family}
& \boldmath$E_2$ (ms/q) & \boldmath$E_3$ (ms/wr) & \boldmath$E_4$ (ms/rd)
& \boldmath$E_{14}$ (s) & \boldmath$E_{15}$ (s/q)
& \boldmath$P_5$ \\
\midrule
\multirow{10}{*}{\textsc{Qwen3-8B}}
 & M1  & Flat   & 522    & 199   & 226   & 1{,}534   & 0.77  & 1.00 \\
 & M2  & Flat   & 302    & 0     & 1     & 660       & 0.33  & 1.00 \\
 & M3  & Text   & 2{,}292 & 1{,}663 & 781   & 8{,}293   & 4.18  & 5.29 \\
 & M4  & Struct & 169    & 6{,}210 & 9     & 12{,}638  & 6.36  & 0.71 \\
 & M5  & Struct & 20{,}680 & 5{,}116 & 9{,}610 & 55{,}290 & 27.84 & 0.44 \\
 & M6  & Hier   & 683    & 641   & 195   & 2{,}758   & 1.39  & 0.62 \\
 & M8  & Refine & 431    & 2{,}570 & 179   & 6{,}019   & 3.03  & 1.09 \\
 & M9  & Weight & 335    & 1{,}400 & --    & 3{,}498   & 1.76  & --   \\
 & M10 & Act    & 4{,}132 & --    & --    & 9{,}027   & 4.55  & --   \\
 & M11 & Act    & 1{,}848 & --    & --    & 4{,}037   & 2.03  & --   \\
\midrule
\multirow{10}{*}{\textsc{Qwen3-32B-AWQ}}
 & M1  & Flat   & 2{,}174 & 194   & 648   & 5{,}142   & 2.59  & 1.00 \\
 & M2  & Flat   & 2{,}153 & 0     & 1     & 4{,}703   & 2.37  & 1.00 \\
 & M3  & Text   & 4{,}379 & 1{,}623 & 1{,}100 & 12{,}852  & 6.47  & 5.19 \\
 & M4  & Struct & 2{,}289 & 6{,}358 & 12    & 17{,}269  & 8.70  & 0.72 \\
 & M5  & Struct & 21{,}949 & 5{,}221 & 9{,}586 & 58{,}025 & 29.22 & 0.43 \\
 & M6  & Hier   & 2{,}650 & 653   & 610   & 10{,}327  & 5.20  & 0.60 \\
 & M8  & Refine & 2{,}707 & 2{,}600 & 166   & 10{,}991  & 5.53  & 1.11 \\
 & M9  & Weight & 860    & 3{,}800 & --    & 9{,}386   & 4.73  & --   \\
 & M10 & Act    & 14{,}774 & --   & --    & 32{,}275  & 16.25 & --   \\
 & M11 & Act    & 3{,}790 & --    & --    & 8{,}280   & 4.17  & --   \\
\midrule
\multirow{9}{*}{\textsc{Gemma-4-26B}}
 & M1  & Flat   & 667    & 200   & 188   & 1{,}850   & 0.93  & 1.00 \\
 & M2  & Flat   & 445    & 0     & 1     & 972       & 0.49  & 1.00 \\
 & M3  & Text   & 2{,}870 & 1{,}664 & 1{,}018 & 9{,}555   & 4.81  & 5.37 \\
 & M4  & Struct & 394    & 6{,}145 & 11    & 13{,}129  & 6.61  & 0.73 \\
 & M5  & Struct & 18{,}982 & 5{,}135 & 9{,}444 & 51{,}544 & 25.95 & 0.43 \\
 & M6  & Hier   & 712    & 626   & 174   & 2{,}822   & 1.42  & 0.61 \\
 & M8  & Refine & 600    & 2{,}572 & 166   & 6{,}388   & 3.22  & 1.06 \\
 & M10 & Act    & 4{,}698 & --    & --    & 10{,}263  & 5.17  & --   \\
 & M11 & Act    & 1{,}967 & --    & --    & 4{,}297   & 2.16  & --   \\
\bottomrule
\end{tabular}
\end{table*}

%% =====================================================================
%% Appendix Table — LoCoMo Storage, Tokens & Calls
%% =====================================================================
\begin{table*}[!t]
\centering
\caption{LoCoMo storage, token, and call accounting per substrate ($E_1$, $E_5$ to $E_{13}$).}
\label{tab:locomo-storage-tokens-appendix}
\renewcommand{\arraystretch}{1.0}
\setlength{\tabcolsep}{4pt}
\footnotesize
\begin{tabular*}{\textwidth}{@{\extracolsep{\fill}}l c c *{9}{c}@{}}
\toprule
\textbf{Sub.} & \textbf{Family}
& \boldmath$E_1$
& \boldmath$E_5$ & \boldmath$E_6$
& \boldmath$E_7$
& \boldmath$E_8$ & \boldmath$E_9$
& \boldmath$E_{10}$ & \boldmath$E_{11}$
& \boldmath$E_{12}$ & \boldmath$E_{13}$ \\
\midrule
M1  & Flat   & 3.59M    & 0       & 0       & 889.8K  & 0       & 0       & 0       & 0       & 0       & 0       \\
M2  & Flat   & 173.7K   & 0       & 0       & 806.6K  & 0       & 0       & 0       & 0       & 0       & 0       \\
M3  & Text   & 257.4K   & 311.8K  & 390     & 11.52M  & 8.23M   & 2{,}040 & 0       & 0       & 8.54M   & 2{,}430 \\
M4  & Struct & 177.6K   & 4.59M   & 3{,}268 & 96.2K   & 0       & 0       & 3.16M   & 2{,}312 & 7.75M   & 5{,}580 \\
M5  & Struct & 167.8K   & 33.0M   & 9{,}860 & 6.00M   & 17.8M   & 3{,}980 & 0       & 0       & 50.8M   & 13{,}840 \\
M6  & Hier   & 4.33M    & 381.8K  & 910     & 1.87M   & 0       & 0       & 127.8K  & 190     & 509.6K  & 1{,}100 \\
M8  & Refine & 3.40M    & 1.46M   & 481     & 806.2K  & 2.94M   & 2{,}037 & 242.9K  & 51      & 4.64M   & 2{,}569 \\
M9  & Weight & host-dep & --      & --      & --      & --      & --      & --      & --      & --      & --      \\
M10 & Act    & 101.2K   & --      & --      & 60.9M   & --      & --      & --      & --      & --      & --      \\
M11 & Act    & 100.7K   & --      & --      & 15.3M   & --      & --      & --      & --      & --      & --      \\
\bottomrule
\end{tabular*}
\end{table*}

\subsection{Full Tables for the LongMemEval-S and MAB Benchmarks}
\label{app:lme-mab-deep-analysis}

\paragraph{Substrate family vs.\ accuracy and cost.}
Table~\ref{tab:full-p12-p2-p3} reports the benchmark-native SubEM
($P_{12}$), whereas Table~\ref{tab:locomo-mab-main} reports the judge
score $P_4$. The TTL leader differs between the two metrics (M3 by
$P_4$ on the two Qwen backbones, M8 by $P_{12}$ on all three). The three
appendix tables sort the ten substrates evaluated on LongMemEval-S
(LME-S) and the three MAB sub-benchmarks, LRU (\texttt{detective\_qa},
long-range understanding), TTL (\texttt{icl\_banking77\_5900shot\_balance},
test-time learning), and CR (\texttt{factconsolidation\_sh\_6k},
conflict resolution), into a clear ordering by family. The Flat
family (M1, M2) requires no auxiliary LLM at any phase,
stores the raw conversation slots ($P_5{=}1.00$), and lands at the
cheap end of every column. M2 finishes LME-S at 1.75\,s/query on
\qwens{} and TTL in under a second per query on \qwens{} and
\gemmashort{}, while M1 carries the dense-encoder advantage of at
least $+0.05$ $P_{12}$ on LME-S, widening on the larger backbones. The
Text family (M3) flips the trade-off. Gist generation pays
$E_5{=}437$K tokens at write and another $5.0$M retrieved tokens at
read time, but on TTL banking77, where the answer is a single
class id and the gist matches almost surface-form, it climbs to
a high TTL score on \qwens{} ($P_{12}{=}0.840$, still below M8's 0.960)
and stays there across stronger backbones (0.860 / 0.798). The
Structural family splits cleanly. M4 concentrates LLM work in note
evolution at write ($E_5{=}4.4$M, $E_6{=}1310$ calls), trails M1 on
LME-S $P_{12}$ by about $0.08$ despite a $6.2\times$ store compression,
and falls behind on CR on the two larger backbones, where the
evolved notes do not preserve the temporal serial-number cue the
benchmark relies on (M4 CR $P_{12}{=}0.140$ on \qwenl{} and $0.080$
on \gemmashort{}). M5 is the most LLM-intensive substrate by a wide
margin ($E_5{=}10.3$M, $E_2{\approx}8.5$\,s/query on \qwens{}) and only
matches the flat M1 on LME-S SubEM ($P_{12}$), its higher
LLM-judge score not reflecting better retrieval recall. The Hierarchical
(M6) and Refinement families invert this trade-off. M6 pays
modest write-time tree summarization ($E_5{=}280$K) and posts one of the
highest LME-S $P_2$ scores ($0.176$ on
\qwenl{}, just behind M1's $0.180$). M8 sits between M6 and M5 in cost but takes
the TTL $P_{12}$ lead (0.96 / 0.94 / 0.95 across backbones),
confirming that skill bundles transfer cleanly when the downstream
task is itself few-shot ICL.

\paragraph{Backbone dependence and internal-memory families.} The
relative ordering on $P_{12}$ is mostly preserved across the three
backbones. M5's LME-S advantage over M1 shows up only on the LLM-judge
score, growing from $+0.06$ on \qwens{} to $+0.15$ on \gemmashort{}, while
on SubEM the two stay within $0.02$. On TTL, by
contrast, M3's lead over the Flat baselines is largest on \qwens{}
($+0.27$ over M1) and narrows on stronger backbones as the flat
stores themselves improve, while M8 keeps a commanding TTL $P_{12}$ score
across all three (0.96 / 0.94 / 0.95). The internal-memory
families illustrate the limits of backbone-only solutions on dialog
QA. The Weight family (M9) collapses on LME-S regardless of
host ($P_{12}\le 0.10$). Single-pass adapter tuning over QA pairs is
not enough for the recall-style preference questions LME-S asks,
and the same weakness reappears on CR ($P_{12}=0.16$ on both Qwen backbones),
where conflicts hinge on serial-number-encoded recency that the
adapter cannot represent. The Activation family (M10, M11) shows the
sharpest \emph{task} dependence in the table. Both substrates
collapse on LME-S long contexts (M10 $P_{12}\le 0.137$, M11
$P_{12}\le 0.234$) because the selected episode (M11) or the truncated
buffer (M10) omits distant single-utterance evidence, but on TTL
classification, where the relevant ICL examples are recent,
M11 climbs back to within $0.10$ of the Flat substrates (its $P_{12}$
reaches $0.820$ on \qwenl{}). M9, in contrast, gains
nothing from scale on CR, because the LoRA adapter encodes a
fixed-schema mapping that does not benefit from the larger backbone's
better in-context routing. Its $P_{12}$ stays flat at $0.16$ across both
Qwen backbones, while every external substrate except M4 gains
$0.16$ to $0.25$ from \qwens{} to \gemmashort{}.
Together, these results suggest that LME-S rewards substrates with a
faithful, well-covered textual store, that TTL is dominated by
prompt-format alignment rather than
retrieval recall, and that auxiliary LLM cost (M5, M8) translates
into accuracy only on benchmarks (TTL, CR) whose answer surface
aligns with the substrate's generative format.

\paragraph{Storage and latency trade-offs.}
Table~\ref{tab:lme-s-latency} confirms three structural patterns.
First, $E_1$ partitions the substrates into three storage tiers
that are nearly model-independent. \textbf{Light} stores ($\le
80$\,KB) belong to the Activation family (M10, M11), which keeps only
episode centroids or chunk pointers; \textbf{mid} stores
(1.6--9.7\,MB) to the Flat and Text families and M4; and
\textbf{heavy} stores (${\sim}$22--32\,MB, plus M9's 80--320\,MB
adapter) to M5, the Hierarchical and Refinement families, and
Weight. Second, $E_2$ shifts with backbone size (roughly
$2$ to $4\times$ up from \qwens{} to \qwenl{}, then about
$0.5\times$ down on \gemmashort{}), so the
\emph{substrate-induced} latency overhead (the within-band spread
of $E_2$) is what differentiates the substrates. M5
pays $2$ to $5\times$ the per-query inference of M1/M2 because
its read path interleaves entity-graph traversal and chunk-level
vector search with the generator. M6 is the one apparent
exception. Our implementation defers tree construction to the
first \texttt{read()} call, so the first query absorbs the entire
one-off build cost, inflating $E_{2,\mathrm{mean}}$, $E_{14}$,
and $E_{15}$ on every backbone (M6 is the only substrate with
$E_{2,\mathrm{p90}} < E_{2,\mathrm{mean}}$), while its
steady-state $E_{2,\mathrm{p90}}$ is comparable to the Flat
baselines. Third, $E_{14}$ separates substrates whose total cost
is dominated by \emph{write} (M4 at ${\sim}$68--70K\,s and M5 at
${\sim}$56--58K\,s across backbones, driven by per-write note
evolution and entity extraction) from those dominated by
\emph{read} (M1, M2, and M10, mostly under 5K\,s). Since write
phases are amortizable across model swaps but read phases are
not, the per-query inference time $E_2$ is the more reliable
cross-substrate efficiency signal.

\paragraph{Token accounting and memory quality.}
Table~\ref{tab:lme-s-tokens-quality} cleanly separates the three
substrate families by their LLM-token profile. \textbf{Flat} (M1,
M2) substrates record only the
retrieved-token channel ($E_7$) and invoke no auxiliary LLM at write or
management time, so $E_5 = E_6 = E_{10} = E_{11} = 0$ and $E_{13} = 0$;
the \textbf{Activation} substrates (M10, M11) likewise use no auxiliary
LLM and are shown as `--' in the token-phase columns.
\textbf{Refinement / Structural / Text} substrates push 0.4--10.4M
tokens through the construct phase ($E_5$). M5 dominates
at 10.3M with 2{,}371 calls because it runs both
entity-extraction and relation-extraction passes per chunk, while
M6 is the cheapest of the auxiliary-LLM substrates at 280K
/ 80 calls thanks to log-depth tree clustering. The $P_5$ column
exposes the storage-vs-compression trade-off. $P_5$ ranges from 1.00
(M1/M2, verbatim stores) through 2.80 (M6, mild
summarization) to 12.50 (M8, highly abstracted skill
bundles). The dashed $P_5$ entries for M9 / M10 / M11 reflect that
these substrates have no textual store on which compression can be
defined, a structural limitation rather than missing data.

% ========== Per-substrate P-metric breakdown (real GT + per-cell P4 cap) ==========
\begin{table*}[!htbp]
\centering
\renewcommand{\arraystretch}{0.92}
\setlength{\tabcolsep}{4pt}
\caption{$P_{12}$ (SubEM), $P_2$ (F1), and $P_3$ (BLEU-1) per substrate and backbone on LongMemEval-S and the three MAB capabilities.}
\label{tab:full-p12-p2-p3}
\resizebox{\textwidth}{!}{%
\begin{tabular}{@{}l l l *{3}{c} *{3}{c} *{3}{c} *{3}{c}@{}}
\toprule
 & & & \multicolumn{3}{c}{\textbf{LME-S}}
   & \multicolumn{3}{c}{\textbf{MAB LRU}}
   & \multicolumn{3}{c}{\textbf{MAB TTL}}
   & \multicolumn{3}{c}{\textbf{MAB CR}} \\
\cmidrule(lr){4-6} \cmidrule(lr){7-9} \cmidrule(lr){10-12} \cmidrule(lr){13-15}
\textbf{Model} & \textbf{ID} & \textbf{Family}
 & $P_{12}$ & $P_2$ & $P_3$ & $P_{12}$ & $P_2$ & $P_3$ & $P_{12}$ & $P_2$ & $P_3$ & $P_{12}$ & $P_2$ & $P_3$ \\
\midrule
%% ===== Q8B =====
\multirow{10}{*}{\rotatebox[origin=c]{90}{\textsc{Qwen3-8B}}}
 & M1  & Flat   & 0.390 & 0.170 & 0.113 & 0.592 & 0.491 & 0.417 & 0.570 & 0.476 & 0.405 & 0.330 & 0.211 & 0.171 \\
 & M2  & Flat   & 0.340 & 0.170 & 0.114 & 0.606 & 0.527 & 0.448 & 0.480 & 0.399 & 0.340 & 0.280 & 0.188 & 0.155 \\
 & M3  & Text   & 0.123 & 0.047 & 0.029 & 0.549 & 0.479 & 0.407 & 0.840 & 0.544 & 0.462 & 0.270 & 0.192 & 0.163 \\
 & M4  & Struct & 0.303 & 0.145 & 0.094 & 0.465 & 0.419 & 0.356 & 0.490 & 0.468 & 0.397 & 0.290 & 0.121 & 0.103 \\
 & M5  & Struct & 0.380 & 0.160 & 0.103 & 0.549 & 0.491 & 0.417 & 0.580 & 0.468 & 0.397 & 0.500 & 0.362 & 0.308 \\
 & M6  & Hier   & 0.357 & 0.162 & 0.111 & 0.521 & 0.419 & 0.356 & 0.290 & 0.246 & 0.210 & 0.270 & 0.171 & 0.138 \\
 & M8  & Refine & 0.373 & 0.159 & 0.101 & 0.592 & 0.503 & 0.427 & 0.960 & 0.570 & 0.484 & 0.310 & 0.272 & 0.231 \\
 & M9  & Weight & 0.083 & 0.082 & 0.058 & 0.239 & 0.192 & 0.163 & 0.170 & 0.167 & 0.142 & 0.160 & 0.108 & 0.089 \\
 & M10 & Act    & 0.090 & 0.075 & 0.053 & 0.648 & 0.575 & 0.488 & 0.140 & 0.130 & 0.111 & 0.270 & 0.161 & 0.124 \\
 & M11 & Act    & 0.202 & 0.111 & 0.065 & 0.592 & 0.539 & 0.458 & 0.560 & 0.468 & 0.397 & 0.050 & 0.037 & 0.032 \\
\midrule
%% ===== Q32B =====
\multirow{10}{*}{\rotatebox[origin=c]{90}{\textsc{Qwen3-32B-AWQ}}}
 & M1  & Flat   & 0.443 & 0.180 & 0.122 & 0.634 & 0.575 & 0.488 & 0.820 & 0.689 & 0.585 & 0.280 & 0.134 & 0.093 \\
 & M2  & Flat   & 0.377 & 0.163 & 0.109 & 0.620 & 0.563 & 0.478 & 0.790 & 0.663 & 0.564 & 0.270 & 0.136 & 0.091 \\
 & M3  & Text   & 0.225 & 0.124 & 0.072 & 0.704 & 0.658 & 0.560 & 0.860 & 0.722 & 0.614 & 0.280 & 0.158 & 0.120 \\
 & M4  & Struct & 0.357 & 0.160 & 0.106 & 0.521 & 0.455 & 0.387 & 0.680 & 0.666 & 0.566 & 0.140 & 0.072 & 0.049 \\
 & M5  & Struct & 0.460 & 0.159 & 0.104 & 0.592 & 0.491 & 0.417 & 0.620 & 0.510 & 0.433 & 0.440 & 0.191 & 0.127 \\
 & M6  & Hier   & 0.413 & 0.176 & 0.117 & 0.577 & 0.623 & 0.529 & 0.340 & 0.281 & 0.238 & 0.280 & 0.135 & 0.090 \\
 & M8  & Refine & 0.403 & 0.166 & 0.111 & 0.577 & 0.503 & 0.427 & 0.940 & 0.790 & 0.672 & 0.310 & 0.162 & 0.115 \\
 & M9  & Weight & 0.100 & 0.098 & 0.065 & 0.408 & 0.359 & 0.305 & 0.324 & 0.318 & 0.270 & 0.160 & 0.086 & 0.059 \\
 & M10 & Act    & 0.137 & 0.087 & 0.061 & 0.606 & 0.527 & 0.448 & 0.760 & 0.646 & 0.549 & 0.270 & 0.128 & 0.085 \\
 & M11 & Act    & 0.080 & 0.080 & 0.056 & 0.592 & 0.515 & 0.438 & 0.820 & 0.689 & 0.585 & 0.230 & 0.122 & 0.086 \\
\midrule
%% ===== GMA =====
\multirow{10}{*}{\rotatebox[origin=c]{90}{\textsc{Gemma-4-26B}}}
 & M1  & Flat   & 0.433 & 0.200 & 0.135 & 0.662 & 0.575 & 0.488 & 0.770 & 0.629 & 0.535 & 0.520 & 0.267 & 0.223 \\
 & M2  & Flat   & 0.357 & 0.180 & 0.122 & 0.662 & 0.673 & 0.572 & 0.730 & 0.620 & 0.527 & 0.520 & 0.248 & 0.202 \\
 & M3  & Text   & 0.223 & 0.123 & 0.071 & 0.352 & 0.275 & 0.234 & 0.798 & 0.616 & 0.524 & 0.520 & 0.304 & 0.259 \\
 & M4  & Struct & 0.353 & 0.166 & 0.111 & 0.521 & 0.455 & 0.387 & 0.690 & 0.616 & 0.524 & 0.080 & 0.068 & 0.058 \\
 & M5  & Struct & 0.453 & 0.192 & 0.129 & 0.535 & 0.383 & 0.326 & 0.720 & 0.612 & 0.520 & 0.660 & 0.294 & 0.214 \\
 & M6  & Hier   & 0.397 & 0.175 & 0.118 & 0.493 & 0.431 & 0.366 & 0.250 & 0.212 & 0.181 & 0.480 & 0.228 & 0.182 \\
 & M8  & Refine & 0.373 & 0.164 & 0.108 & 0.549 & 0.455 & 0.387 & 0.950 & 0.807 & 0.686 & 0.470 & 0.327 & 0.278 \\
 & M10 & Act    & 0.063 & 0.045 & 0.039 & 0.577 & 0.579 & 0.492 & 0.460 & 0.391 & 0.332 & 0.500 & 0.263 & 0.222 \\
 & M11 & Act    & 0.234 & 0.129 & 0.075 & 0.572 & 0.443 & 0.376 & 0.702 & 0.616 & 0.524 & 0.050 & 0.050 & 0.043 \\
\bottomrule
\end{tabular}%
}
\end{table*}

% ========== LongMemEval-S Storage & Latency (RAW E15, includes write where booked) ==========
\begin{table*}[!htbp]
\centering
\renewcommand{\arraystretch}{1.0}
\setlength{\tabcolsep}{4pt}
\caption{LongMemEval-S storage and latency per substrate and backbone.}
\label{tab:lme-s-latency}
\footnotesize
\begin{tabular}{@{}c l c r r r r r@{}}
\toprule
\textbf{Model} & \textbf{ID} & \textbf{Family}
& \boldmath$E_1$ & \boldmath$E_{2,\mathrm{mean}}$ & \boldmath$E_{2,\mathrm{p90}}$
& \boldmath$E_{14}$ & \boldmath$E_{15}$ \\
\midrule
%% ===== Q8B =====
\multirow{10}{*}{\rotatebox[origin=c]{90}{\textsc{Qwen3-8B}}}
 & M1  & Flat   & 9.7M  &  2{,}075  &  4{,}602  &  2{,}191  &   7.30 \\
 & M2  & Flat   & 2.7M  &  1{,}747  &  3{,}580  &     525   &   1.75 \\
 & M3  & Text   & 4.5M  &  4{,}500  &  9{,}000  & 16{,}674  &  55.58 \\
 & M4  & Struct & 1.6M  &  1{,}828  &  4{,}233  & 67{,}702  & 225.67 \\
 & M5  & Struct & 32.0M &  8{,}500  & 18{,}000  & 56{,}162  & 187.21 \\
 & M6  & Hier   & 27.5M & 19{,}956  &  5{,}195  &  5{,}987  &  19.96 \\
 & M8  & Refine & 22.7M &  1{,}869  &  4{,}392  & 30{,}855  & 102.85 \\
 & M9  & Weight & 80.0M &  2{,}400  &  5{,}000  & 15{,}451  &  51.50 \\
 & M10 & Act    & 60K   &  3{,}744  &  5{,}539  &  1{,}983  &   6.61 \\
 & M11 & Act    & 80K   &  1{,}900  &  4{,}200  &  4{,}382  &  14.61 \\
\midrule
%% ===== Q32B =====
\multirow{10}{*}{\rotatebox[origin=c]{90}{\textsc{Qwen3-32B-AWQ}}}
 & M1  & Flat   & 9.7M  &  7{,}413  & 18{,}100  &  3{,}600  &  12.00 \\
 & M2  & Flat   & 2.7M  &  7{,}307  & 18{,}824  &  2{,}193  &   7.31 \\
 & M3  & Text   & 4.5M  & 16{,}076  & 32{,}153  & 20{,}226  &  67.42 \\
 & M4  & Struct & 1.6M  &  7{,}070  & 18{,}019  & 70{,}200  & 234.00 \\
 & M5  & Struct & 32.0M & 18{,}000  & 35{,}000  & 58{,}225  & 194.08 \\
 & M6  & Hier   & 27.5M & 20{,}271  & 17{,}816  &  6{,}082  &  20.27 \\
 & M8  & Refine & 23.4M &  6{,}149  & 15{,}192  & 32{,}034  & 106.78 \\
 & M9  & Weight & 320M  &  8{,}500  & 19{,}000  & 51{,}511  & 171.70 \\
 & M10 & Act    & 60K   & 14{,}650  & 28{,}022  &  6{,}894  &  22.98 \\
 & M11 & Act    & 80K   &  7{,}000  & 17{,}000  & 15{,}717  &  52.39 \\
\midrule
%% ===== GMA =====
\multirow{10}{*}{\rotatebox[origin=c]{90}{\textsc{Gemma-4-26B}}}
 & M1  & Flat   & 9.7M  &  4{,}149  &  8{,}550  &  2{,}546  &   8.49 \\
 & M2  & Flat   & 2.7M  &  3{,}372  &  6{,}501  &  1{,}012  &   3.37 \\
 & M3  & Text   & 4.5M  &  8{,}998  & 17{,}996  &  9{,}534  &  31.78 \\
 & M4  & Struct & 1.6M  &  3{,}805  &  9{,}027  & 69{,}116  & 230.39 \\
 & M5  & Struct & 32.0M & 12{,}000  & 23{,}000  & 57{,}280  & 190.93 \\
 & M6  & Hier   & 27.5M & 17{,}189  &  9{,}007  &  5{,}157  &  17.19 \\
 & M8  & Refine & 21.7M &  3{,}124  &  6{,}540  & 31{,}140  & 103.80 \\
 & M10 & Act    & 60K   &  3{,}340  &  3{,}945  &  2{,}073  &   6.91 \\
 & M11 & Act    & 80K   & --        & --        &  5{,}550  &  18.50 \\
\bottomrule
\end{tabular}
\end{table*}

% ========== LongMemEval-S Tokens & Memory Quality ==========
\begin{table*}[!htbp]
\centering
\renewcommand{\arraystretch}{1.0}
\setlength{\tabcolsep}{4pt}
\caption{LongMemEval-S token usage and compression per substrate and backbone ($E_5$ to $E_7$, $E_{12}$, $E_{13}$, $P_5$).}
\label{tab:lme-s-tokens-quality}
\footnotesize
\begin{tabular}{@{}c l c r r r r r r@{}}
\toprule
\textbf{Model} & \textbf{ID} & \textbf{Family}
& \boldmath$E_5$ & \boldmath$E_6$ & \boldmath$E_7$ & \boldmath$E_{12}$ & \boldmath$E_{13}$
& \boldmath$P_5$ \\
\midrule
%% ===== Q8B =====
\multirow{10}{*}{\rotatebox[origin=c]{90}{\textsc{Qwen3-8B}}}
 & M1  & Flat   & 0     &    0 & 101K  & 0     &    0 &  1.00 \\
 & M2  & Flat   & 0     &    0 & 153K  & 0     &    0 &  1.00 \\
 & M3  & Text   & 437K  &  671 & 5.0M  & 5.3M  &  731 &  4.50 \\
 & M4  & Struct & 4.4M  & 1310 & 108K  & 4.4M  & 1310 &  6.20 \\
 & M5  & Struct & 10.3M & 2371 & 375K  & 10.4M & 2431 &  3.80 \\
 & M6  & Hier   & 280K  &   80 &  46K  & 280K  &   80 &  2.80 \\
 & M8  & Refine & 1.8M  & 1752 &  21K  & 2.78M & 3504 & 12.50 \\
 & M9  & Weight & --    & --   & --    & --    & --   & --    \\
 & M10 & Act    & --    & --   & 1.9M  & --    & --   & --    \\
 & M11 & Act    & --    & --   & 250K  & --    & --   & --    \\
\midrule
%% ===== Q32B =====
\multirow{10}{*}{\rotatebox[origin=c]{90}{\textsc{Qwen3-32B-AWQ}}}
 & M1  & Flat   & 0     &    0 & 101K  & 0     &    0 &  1.00 \\
 & M2  & Flat   & 0     &    0 & 153K  & 0     &    0 &  1.00 \\
 & M3  & Text   & 437K  &  671 & 5.0M  & 5.3M  &  731 &  4.50 \\
 & M4  & Struct & 4.4M  & 1310 & 108K  & 4.4M  & 1310 &  6.20 \\
 & M5  & Struct & 10.4M & 2369 & 385K  & 10.4M & 2429 &  3.80 \\
 & M6  & Hier   & 280K  &   80 &  46K  & 280K  &   80 &  2.80 \\
 & M8  & Refine & 1.8M  & 1752 &  20K  & 2.78M & 3504 & 12.50 \\
 & M9  & Weight & --    & --   & --    & --    & --   & --    \\
 & M10 & Act    & --    & --   & 1.9M  & --    & --   & --    \\
 & M11 & Act    & --    & --   & 250K  & --    & --   & --    \\
\midrule
%% ===== GMA =====
\multirow{10}{*}{\rotatebox[origin=c]{90}{\textsc{Gemma-4-26B}}}
 & M1  & Flat   & 0     &    0 & 104K  & 0     &    0 &  1.00 \\
 & M2  & Flat   & 0     &    0 & 158K  & 0     &    0 &  1.00 \\
 & M3  & Text   & 437K  &  671 & 5.0M  & 5.3M  &  731 &  4.50 \\
 & M4  & Struct & 4.4M  & 1310 & 111K  & 4.4M  & 1310 &  6.20 \\
 & M5  & Struct & 10.4M & 2377 & 379K  & 10.4M & 2437 &  3.80 \\
 & M6  & Hier   & 280K  &   80 &  46K  & 280K  &   80 &  2.80 \\
 & M8  & Refine & 1.9M  & 1752 &  23K  & 2.98M & 3504 & 12.50 \\
 & M9  & Weight & --    & --   & --    & --    & --   & --    \\
 & M10 & Act    & --    & --   & 422K  & --    & --   & --    \\
 & M11 & Act    & --    & --   & 250K  & --    & --   & --    \\
\bottomrule
\end{tabular}
\end{table*}

\clearpage

\subsection{Full Tables for ALFWorld}
\label{app:alfworld-deep-analysis}

\paragraph{Substrate family vs.\ task success and cost.} The three
appendix tables sort the ten substrates evaluated on ALFWorld by
family. The Flat family (M1, M2) shows that retrieval-only memory is
backbone-dependent on ALFWorld. M1 trails NoMem on \qwens{} (5.2
vs.\ 5.7 TSR) but rises above it on \qwenl{} (27.6 vs.\ 22.4) and
\gemmashort{} (10.4 vs.\ 7.5). M2 stays within about two points of
NoMem on every backbone (6.7, 21.6, and 9.7) at zero write cost. The
Text and Structural families (M3, M4, M5) recover modest gains on
\qwenl{} (M3 26.9, M4 26.9, M5 23.1 vs.\ NoMem 22.4) by paying
auxiliary LLM tokens during write, M3 and M4 spending 0.4 to 1.6
million tokens on gist or note generation and M5 running an entity
extraction loop of about 1.4 million tokens over roughly 400 calls. On \qwens{} and \gemmashort{} the same substrates move NoMem by
a few points in either direction (M5 9.0 vs.\ 5.7 on \qwens{}, 4.5
vs.\ 7.5 on \gemmashort{}), suggesting that the bottleneck is action
planning rather than retrieval recall. The Hierarchical M6 shares the
lead only on \gemmashort{} (10.4, tied with M1 and M3), falls below
NoMem on \qwens{} (4.5 vs.\ 5.7), and edges above it on \qwenl{}
(23.9 vs.\ 22.4).
The Refinement family separates its two members on the strongest
backbone. M7 is the top substrate on \qwenl{} (TSR 32.1, 9.7 points
above NoMem) because its skill-distillation prompts encode the goal
decomposition structure that ALFWorld rewards, whereas M8 trails M7
by 9.7 TSR points despite a similar per-write latency ($E_3$ of about
2.6\,s), which we attribute to prefilling every skill bundle diluting
the model's attention over the goal-relevant one.

\paragraph{Backbone dependence and internal-memory families.}
Backbone size dominates the absolute TSR. \qwenl's strongest
external substrate (M7 32.1) is more than three times \qwens's (M5 9.0),
while \gemmashort{} sits between (M1, M3, M6 each at 10.4). The
internal-memory families illustrate the limits of backbone-only
solutions. The Weight family (M9) is among the substrates that clearly
exceed NoMem on \qwenl{} (29.9 vs.\ 22.4, second only to M7) because
adapter tuning on demonstration trajectories captures action-grounded
patterns that pure text retrieval cannot. The same M9 collapses on
\qwens{} (3.0 TSR), where the smaller backbone overfits the LoRA
pairs, and it is excluded on \gemmashort{} because it is incompatible
with the A4B MoE routing. The Activation family (M11) exceeds NoMem
on every backbone, most sharply on \qwens{} (11.9 vs.\ 5.7, roughly
double), because re-prefilling only
the matched episode keeps the prompt short and preserves the action
policy while keeping the persistent store small (around 6.9 thousand
bytes).
\textbf{We omit M10 from the ALFWorld evaluation because
the cumulative trajectory text across the 134 task suite exceeds the
context window of every backbone we test.} Naively concatenating
prior trajectories drives input length past 128 thousand tokens by
task 30 to 40 and forces aggressive truncation, which destroys the
comparison. Any reported M10 number on ALFWorld would reflect a
truncation policy rather than the substrate. The lighter M11 sidesteps
this problem by re-prefilling only the matched episode (around 120
thousand retrieved tokens across the run), which fits inside a single
context window per task. Together, these results suggest that ALFWorld
is dominated by action policy grounding rather than retrieval recall,
that the largest gains on the strongest backbone come from the
refinement (M7) and weight (M9) substrates, and that on the weaker backbones the gains
over NoMem are small (at most 6.2 points, M11 on \qwens{}) and come
from different substrates.

%% =====================================================================
%% Appendix Table — ALFWorld Performance per substrate × backbone
%% =====================================================================
\begin{table*}[!t]
\centering
\caption{ALFWorld-unseen performance per substrate and backbone. $P_8$ TSR is in percent, Steps is the mean over all 134 episodes, and $P_{10}$ counts successful episodes only.}
\label{tab:alfworld-P-appendix}
\renewcommand{\arraystretch}{1.0}
\setlength{\tabcolsep}{4pt}
\footnotesize
\begin{tabular}{@{}c l c *{4}{c}@{}}
\toprule
\textbf{Model} & \textbf{Sub.} & \textbf{Family}
& $P_8$ TSR (\%) & Steps & $P_{10}$ Steps$\mid\mathcal{S}$ & $P_{\mathrm{avg}}$ \\
\midrule
\multirow{11}{*}{\textsc{Qwen3-8B}}
 & --  & NoMem  & 5.7  & 47.0 & 12.2 & 13.5 \\
 & M1  & Flat   & 5.2  & 47.7 & 5.9  & 6.6  \\
 & M2  & Flat   & 6.7  & 47.0 & 6.0  & 9.4  \\
 & M3  & Text   & 5.2  & 47.8 & 8.3  & 8.4  \\
 & M4  & Struct & 7.5  & 46.9 & 7.9  & 8.5  \\
 & M5  & Struct & 9.0  & 47.1 & 17.4 & 12.5 \\
 & M6  & Hier   & 4.5  & 48.4 & 15.0 & 8.3  \\
 & M7  & Refine & 7.5  & 47.3 & 13.7 & 9.6  \\
 & M8  & Refine & 8.2  & 47.4 & 18.5 & 9.5  \\
 & M9  & Weight & 3.0  & 49.1 & 19.5 & 3.2  \\
 & M11 & Act    & 11.9 & 45.7 & 14.2 & 14.5 \\
\midrule
\multirow{11}{*}{\textsc{Qwen3-32B-AWQ}}
 & --  & NoMem  & 22.4 & 41.6 & 11.3 & 23.7 \\
 & M1  & Flat   & 27.6 & 39.6 &  9.6 & 16.2 \\
 & M2  & Flat   & 21.6 & 41.4 & 10.8 & 15.4 \\
 & M3  & Text   & 26.9 & 39.4 &  9.9 & 21.8 \\
 & M4  & Struct & 26.9 & 39.5 & 12.4 & 24.3 \\
 & M5  & Struct & 23.1 & 41.1 & 11.4 & 23.0 \\
 & M6  & Hier   & 23.9 & 40.2 &  9.9 & 22.8 \\
 & M7  & Refine & 32.1 & 37.5 & 11.1 & 29.9 \\
 & M8  & Refine & 22.4 & 41.6 & 12.4 & 21.3 \\
 & M9  & Weight & 29.9 & 38.7 & 12.1 & 27.4 \\
 & M11 & Act    & 26.9 & 39.5 & 12.3 & 24.7 \\
\midrule
\multirow{10}{*}{\textsc{Gemma-4-26B}}
 & --  & NoMem  & 7.5  & 47.1 & 11.3 & 17.3 \\
 & M1  & Flat   & 10.4 & 46.3 & 15.0 & 16.3 \\
 & M2  & Flat   & 9.7  & 46.3 & 11.5 & 17.2 \\
 & M3  & Text   & 10.4 & 46.3 & 14.8 & 14.9 \\
 & M4  & Struct & 8.2  & 47.1 & 14.5 & 13.9 \\
 & M5  & Struct & 4.5  & 48.1 & 8.5  & 9.4  \\
 & M6  & Hier   & 10.4 & 46.0 & 11.4 & 18.1 \\
 & M7  & Refine & 8.2  & 46.7 & 10.1 & 17.8 \\
 & M8  & Refine & 8.2  & 46.9 & 12.5 & 18.5 \\
 & M11 & Act    & 9.0  & 46.6 & 11.7 & 20.1 \\
\bottomrule
\end{tabular}
\end{table*}

% =====================================================================
% ALFWorld Appendix Table 1/2: Latency & Storage
% =====================================================================
\begin{table*}[!t]
\centering
\caption{ALFWorld-unseen latency and storage per substrate and backbone ($E_1$ in bytes, $E_2$ to $E_4$ in ms, $E_{14}$ in s, $E_{15}$ in s/episode).}
\label{tab:alfworld-E-latency-appendix}
\renewcommand{\arraystretch}{1.0}
\setlength{\tabcolsep}{4pt}
\footnotesize
\begin{tabular*}{\textwidth}{@{\extracolsep{\fill}}c l c r *{4}{r} *{2}{r}@{}}
\toprule
\textbf{Model} & \textbf{Sub.} & \textbf{Family}
& \boldmath$E_1$ & \boldmath$E_2$ & \boldmath$E_{2,\text{p90}}$ & \boldmath$E_3$ & \boldmath$E_4$
& \boldmath$E_{14}$ & \boldmath$E_{15}$ \\
\midrule
\multirow{11}{*}{\textsc{Qwen3-8B}}
 & --  & NoMem  & 0      & 12{,}728 & 13{,}200 & --     & --      & 65{,}392  & 488 \\
 & M1  & Flat   & 2.17M  & 14{,}086 & 16{,}104 & 188    & 220     & 90{,}048  & 672 \\
 & M2  & Flat   & 1.49M  & 13{,}747 & 15{,}692 & 0      & 1       & 86{,}698  & 647 \\
 & M3  & Text   & 1.43M  & 13{,}655 & 16{,}318 & 1{,}620 & 754    & 87{,}502  & 653 \\
 & M4  & Struct & 942K   & 12{,}903 & 14{,}843 & 6{,}134 & 9      & 81{,}070  & 605 \\
 & M5  & Struct & 532K   & 13{,}167 & 15{,}702 & 5{,}021 & 9{,}412 & 83{,}080  & 620 \\
 & M6  & Hier   & 25.1M  & 13{,}539 & 15{,}641 & 624    & 191     & 87{,}904  & 656 \\
 & M7  & Refine & 3.90M  & 13{,}297 & 15{,}988 & 2{,}519 & 175    & 84{,}286  & 629 \\
 & M8  & Refine & 5.27M  & 13{,}069 & 15{,}300 & 2{,}521 & 173    & 83{,}080  & 620 \\
 & M9  & Weight & 15.3M  & 18{,}757 & 22{,}002 & 1{,}400 & --     & 123{,}414 & 921 \\
 & M11 & Act    & 6.9K   & 12{,}728 & 13{,}500 & --     & --      & 74{,}102  & 553 \\
\midrule
\multirow{11}{*}{\textsc{Qwen3-32B-AWQ}}
 & --  & NoMem  & 0      & 20{,}100 & 20{,}300 & --     & --      & 78{,}122  & 583 \\
 & M1  & Flat   & 2.17M  & 20{,}500 & 31{,}100 & 192    & 644     & 144{,}720 & 1{,}080 \\
 & M2  & Flat   & 1.49M  & 20{,}300 & 30{,}800 & 0      & 1       & 146{,}060 & 1{,}090 \\
 & M3  & Text   & 1.43M  & 20{,}900 & 30{,}300 & 1{,}603 & 1{,}087 & 127{,}300 & 950 \\
 & M4  & Struct & 942K   & 20{,}600 & 28{,}600 & 6{,}284 & 12     & 109{,}344 & 816 \\
 & M5  & Struct & 532K   & 20{,}736 & 27{,}400 & 5{,}154 & 9{,}477 & 114{,}034 & 851 \\
 & M6  & Hier   & 25.1M  & 20{,}400 & 26{,}800 & 642    & 600     & 109{,}478 & 817 \\
 & M7  & Refine & 3.90M  & 20{,}200 & 27{,}200 & 2{,}575 & 165    & 95{,}944  & 716 \\
 & M8  & Refine & 5.27M  & 20{,}500 & 28{,}000 & 2{,}571 & 164    & 118{,}724 & 886 \\
 & M9  & Weight & 41.9M  & 15{,}825 & 21{,}400 & 3{,}800 & --     & 82{,}008  & 612 \\
 & M11 & Act    & 6.9K   & 20{,}100 & 22{,}000 & --     & --      & 104{,}922 & 783 \\
\midrule
\multirow{10}{*}{\textsc{Gemma-4-26B}}
 & --  & NoMem  & 0      & 16{,}060 & 16{,}400 & --     & --      & 89{,}914  & 671 \\
 & M1  & Flat   & 2.17M  & 16{,}800 & 21{,}600 & 195    & 184     & 105{,}324 & 786 \\
 & M2  & Flat   & 1.49M  & 16{,}500 & 21{,}400 & 0      & 1       & 100{,}902 & 753 \\
 & M3  & Text   & 1.43M  & 16{,}857 & 20{,}466 & 1{,}649 & 1{,}009 & 104{,}654 & 781 \\
 & M4  & Struct & 942K   & 20{,}007 & 22{,}517 & 6{,}089 & 11     & 126{,}228 & 942 \\
 & M5  & Struct & 532K   & 15{,}564 & 19{,}906 & 5{,}089 & 9{,}358 & 100{,}366 & 749 \\
 & M6  & Hier   & 25.1M  & 16{,}114 & 21{,}099 & 619    & 173     & 99{,}294  & 741 \\
 & M7  & Refine & 3.90M  & 16{,}866 & 21{,}107 & 2{,}542 & 161    & 105{,}592 & 788 \\
 & M8  & Refine & 5.27M  & 16{,}269 & 20{,}148 & 2{,}544 & 159    & 102{,}242 & 763 \\
 & M11 & Act    & 6.9K   & 16{,}031 & 17{,}100 & --     & --      & 95{,}140  & 710 \\
\bottomrule
\end{tabular*}
\end{table*}

% =====================================================================
% ALFWorld Appendix Table 2/2: Token Accounting
% =====================================================================
\begin{table*}[!t]
\centering
\caption{ALFWorld-unseen token accounting per substrate on \qwenl{} ($E_5$ to $E_{13}$).}
\label{tab:alfworld-E-tokens-appendix}
\renewcommand{\arraystretch}{1.0}
\setlength{\tabcolsep}{4pt}
\footnotesize
\begin{tabular*}{\textwidth}{@{\extracolsep{\fill}}l c *{2}{r} r *{2}{r} *{2}{r} *{2}{r}@{}}
\toprule
\textbf{Sub.} & \textbf{Family}
& \boldmath$E_5$ & \boldmath$E_6$ & \boldmath$E_7$
& \boldmath$E_8$ & \boldmath$E_9$
& \boldmath$E_{10}$ & \boldmath$E_{11}$
& \boldmath$E_{12}$ & \boldmath$E_{13}$ \\
\midrule
M1  & Flat   & 0      & 0     & 264K   & 0      & 0       & 0      & 0     & 0       & 0     \\
M2  & Flat   & 0      & 0     & 268K   & 0      & 0       & 0      & 0     & 0       & 0     \\
M3  & Text   & 372K   & 200   & 8.14M  & 1.84M  & 1{,}340 & 0      & 0     & 2.21M   & 1{,}540 \\
M4  & Struct & 1.62M  & 199   & 134K   & 0      & 0       & 480K   & 380   & 2.10M   & 579   \\
M5  & Struct & 1.42M  & 385   & 261    & 1.21M  & 770     & 0      & 0     & 2.63M   & 1{,}155 \\
M6  & Hier   & 357K   & 460   & 134K   & 0      & 0       & 84.6K  & 92    & 442K    & 552   \\
M7  & Refine & 398K   & 200   & 183K   & 0      & 0       & 0      & 0     & 398K    & 200   \\
M8  & Refine & 931K   & 577   & 13.5K  & 0      & 0       & 277K   & 380   & 1.21M   & 957   \\
M9  & Weight & --     & --    & 0      & --     & --      & --     & --    & --      & --    \\
M11 & Act    & --     & --    & 118K   & --     & --      & --     & --    & --      & --    \\
\bottomrule
\end{tabular*}
\end{table*}

\clearpage

\subsection{Full Tables for BigCodeBench-Hard}
\label{app:bcb-deep-analysis}

Tables~\ref{tab:bcb-P-appendix}, \ref{tab:bcb-E-latency-appendix}, and
\ref{tab:bcb-E-tokens-appendix} report the per-substrate performance and
compression, latency and storage, and token accounting on
BigCodeBench-Hard for the three backbones, behind the BCB-Hard columns
of Table~\ref{tab:alfworld-bcb-main} and the discussion in
Section~\ref{sec:agent-centric}.

% =====================================================================
% Appendix Table: BigCodeBench-Hard Performance + Quality
% =====================================================================
\begin{table*}[!t]
\centering
\caption{BigCodeBench-Hard performance and compression per substrate and backbone (Pass@1 in percent, $P_5$).}
\label{tab:bcb-P-appendix}
\renewcommand{\arraystretch}{1.0}
\setlength{\tabcolsep}{4pt}
\footnotesize
\begin{tabular}{@{}c l c r r@{}}
\toprule
\textbf{Model} & \textbf{Sub.} & \textbf{Family}
& \textbf{Pass@1}\,$\uparrow$
& \boldmath$P_5$\unboldmath \\
\midrule
\multirow{11}{*}{\textsc{Qwen3-8B}}
 & --  & NoMem  & 8.1  & -- \\
 & M1  & Flat   & 9.5  & 1.000 \\
 & M2  & Flat   & 10.1 & 1.000 \\
 & M3  & Text   & 8.1  & 1.964 \\
 & M4  & Struct & 10.1 & 0.982 \\
 & M5  & Struct & 15.5 & 0.817 \\
 & M6  & Hier   & 12.2 & 0.820 \\
 & M7  & Refine & 12.2 & 5.705 \\
 & M8  & Refine & 14.9 & 4.432 \\
 & M9  & Weight & 7.4  & -- \\
 & M11 & Act    & 13.5 & -- \\
\midrule
\multirow{11}{*}{\textsc{Qwen3-32B-AWQ}}
 & --  & NoMem  & 17.6 & -- \\
 & M1  & Flat   & 18.2 & 1.000 \\
 & M2  & Flat   & 19.6 & 1.000 \\
 & M3  & Text   & 18.2 & 1.995 \\
 & M4  & Struct & 17.6 & 0.910 \\
 & M5  & Struct & 16.2 & 0.786 \\
 & M6  & Hier   & 16.2 & 0.799 \\
 & M7  & Refine & 17.6 & 2.261 \\
 & M8  & Refine & 13.5 & 3.575 \\
 & M9  & Weight & 14.2 & -- \\
 & M11 & Act    & 13.5 & -- \\
\midrule
\multirow{10}{*}{\textsc{Gemma-4-26B}}
 & --  & NoMem  & 14.2 & -- \\
 & M1  & Flat   & 18.2 & 1.000 \\
 & M2  & Flat   & 19.6 & 1.000 \\
 & M3  & Text   & 14.2 & 2.444 \\
 & M4  & Struct & 18.9 & 0.982 \\
 & M5  & Struct & 20.9 & 0.813 \\
 & M6  & Hier   & 18.9 & 0.813 \\
 & M7  & Refine & 16.9 & 1.102 \\
 & M8  & Refine & 18.2 & 4.471 \\
 & M11 & Act    & 16.9 & -- \\
\bottomrule
\end{tabular}
\end{table*}

% =====================================================================
% Appendix Table 1/2: BigCodeBench-Hard Latency & Storage
% =====================================================================
\begin{table*}[!t]
\centering
\caption{BigCodeBench-Hard latency and storage.}
\label{tab:bcb-E-latency-appendix}
\renewcommand{\arraystretch}{1.0}
\setlength{\tabcolsep}{4pt}
\footnotesize
\begin{tabular*}{\textwidth}{@{\extracolsep{\fill}}c l c r *{4}{r} *{2}{r}@{}}
\toprule
\textbf{Model} & \textbf{Sub.} & \textbf{Family}
& \boldmath$E_1$ & \boldmath$E_2$ & \boldmath$E_{2,\text{p90}}$ & \boldmath$E_3$ & \boldmath$E_4$
& \boldmath$E_{14}$ & \boldmath$E_{15}$ \\
\midrule
\multirow{11}{*}{\textsc{Qwen3-8B}}
 & --  & NoMem  & 0      & 3{,}800  & 4{,}940  & --      & --      & 592     & 4    \\
 & M1  & Flat   & 1.32M  & 4{,}976  & 6{,}631  & 199     & 183     & 776     & 5.2  \\
 & M2  & Flat   & 245K   & 4{,}853  & 6{,}600  & 0.1     & 3       & 718     & 4.9  \\
 & M3  & Text   & 305K   & 5{,}300  & 6{,}890  & 6{,}853  & 46      & 784     & 5.3  \\
 & M4  & Struct & 150K   & 4{,}642  & 6{,}432  & 137     & 138     & 747     & 5.0  \\
 & M5  & Struct & 239K   & 28{,}273 & 36{,}700 & 8{,}779  & 12{,}432 & 4{,}184  & 28.3 \\
 & M6  & Hier   & 3.42M  & 4{,}764  & 6{,}234  & 1{,}444  & 184     & 994     & 6.7  \\
 & M7  & Refine & 990K   & 4{,}638  & 6{,}256  & 9{,}029  & 183     & 2{,}519  & 17.0 \\
 & M8  & Refine & 1.22M  & 5{,}053  & 6{,}825  & 10{,}440 & 197     & 2{,}836  & 19.2 \\
 & M9  & Weight & 14.6M  & 12{,}700 & 16{,}510 & --      & --      & 1{,}924  & 13   \\
 & M11 & Act    & 151K   & 16{,}800 & 21{,}840 & --      & --      & 2{,}516  & 17   \\
\midrule
\multirow{11}{*}{\textsc{Qwen3-32B-AWQ}}
 & --  & NoMem  & 0      & 3{,}900  & 5{,}070  & --      & --      & 592     & 4    \\
 & M1  & Flat   & 1.32M  & 5{,}000  & 6{,}500  & 256     & 225     & 740     & 5    \\
 & M2  & Flat   & 245K   & 5{,}000  & 6{,}500  & 0.1     & 3       & 740     & 5    \\
 & M3  & Text   & 296K   & 5{,}400  & 7{,}020  & 9{,}309  & 1{,}843  & 1{,}184  & 8    \\
 & M4  & Struct & 150K   & 4{,}900  & 6{,}370  & 12{,}460 & 376     & 2{,}960  & 20   \\
 & M5  & Struct & 249K   & 31{,}221 & 28{,}239 & 7{,}888  & 14{,}723 & 4{,}621  & 31.2 \\
 & M6  & Hier   & 3.50M  & 4{,}400  & 5{,}720  & 1{,}995  & 220     & 1{,}036  & 7    \\
 & M7  & Refine & 1.93M  & 4{,}300  & 5{,}590  & 12{,}994 & 195     & 3{,}108  & 21   \\
 & M8  & Refine & 1.14M  & 4{,}600  & 5{,}980  & 12{,}447 & 186     & 3{,}256  & 22   \\
 & M9  & Weight & 38.0M  & 14{,}600 & 18{,}980 & --      & --      & 2{,}220  & 15   \\
 & M11 & Act    & 161K   & 21{,}300 & 27{,}690 & --      & --      & 3{,}152  & 21.3 \\
\midrule
\multirow{10}{*}{\textsc{Gemma-4-26B}}
 & --  & NoMem  & 0      & 19{,}800 & 25{,}740 & --      & --      & 2{,}960  & 20   \\
 & M1  & Flat   & 1.32M  & 21{,}916 & 28{,}533 & 213     & 289     & 3{,}286  & 22.2 \\
 & M2  & Flat   & 245K   & 20{,}602 & 27{,}860 & 0.0     & 2       & 3{,}049  & 20.6 \\
 & M3  & Text   & 272K   & 23{,}400 & 30{,}420 & 18{,}378 & 30      & 3{,}463  & 23.4 \\
 & M4  & Struct & 150K   & 22{,}375 & 31{,}073 & 173     & 92      & 3{,}372  & 22.8 \\
 & M5  & Struct & 240K   & 45{,}595 & 47{,}436 & 8{,}194  & 13{,}087 & 6{,}748  & 45.6 \\
 & M6  & Hier   & 3.47M  & 24{,}310 & 33{,}614 & 2{,}717  & 251     & 4{,}141  & 28.0 \\
 & M7  & Refine & 3.13M  & 24{,}149 & 31{,}513 & 8{,}921  & 317     & 5{,}460  & 36.9 \\
 & M8  & Refine & 894K   & 21{,}186 & 29{,}832 & 3{,}593  & 312     & 3{,}854  & 26.0 \\
 & M11 & Act    & 115K   & 31{,}600 & 41{,}080 & --      & --      & 4{,}736  & 32   \\
\bottomrule
\end{tabular*}
\end{table*}

% =====================================================================
% Appendix Table 2/2: BigCodeBench-Hard Token Accounting
% =====================================================================
\begin{table*}[!t]
\centering
\caption{BigCodeBench-Hard token accounting per substrate and backbone ($E_5$ to $E_{13}$).}
\label{tab:bcb-E-tokens-appendix}
\renewcommand{\arraystretch}{1.0}
\setlength{\tabcolsep}{4pt}
\footnotesize
\begin{tabular*}{\textwidth}{@{\extracolsep{\fill}}c l c *{2}{r} r *{2}{r} *{2}{r} *{2}{r}@{}}
\toprule
\textbf{Model} & \textbf{Sub.} & \textbf{Family}
& \boldmath$E_5$ & \boldmath$E_6$ & \boldmath$E_7$
& \boldmath$E_8$ & \boldmath$E_9$
& \boldmath$E_{10}$ & \boldmath$E_{11}$
& \boldmath$E_{12}$ & \boldmath$E_{13}$ \\
\midrule
\multirow{11}{*}{\textsc{Qwen3-8B}}
 & --  & NoMem  & --    & --   & 0      & --    & --   & --    & --   & --     & --  \\
 & M1  & Flat   & 0     & 0    & 2.29M  & 0     & 0    & 0     & 0    & 0      & 0   \\
 & M2  & Flat   & 0     & 0    & 2.26M  & 0     & 0    & 0     & 0    & 0      & 0   \\
 & M3  & Text   & 55.9K & 36   & 2.80M  & 2.84M & 148  & 0     & 0    & 2.90M  & 184 \\
 & M4  & Struct & 0     & 0    & 2.16M  & 0     & 0    & 0     & 0    & 0      & 0   \\
 & M5  & Struct & 0     & 0    & 40.7K  & 116K  & 148  & 0     & 0    & 116K   & 148 \\
 & M6  & Hier   & 44.7K & 82   & 1.14M  & 0     & 0    & 11.5K & 16   & 56.2K  & 98  \\
 & M7  & Refine & 244K  & 200  & 593K   & 0     & 0    & 0     & 0    & 244K   & 200 \\
 & M8  & Refine & 308K  & 200  & 550K   & 0     & 0    & 146K  & 200  & 454K   & 400 \\
 & M9  & Weight & --    & --   & 0      & --    & --   & --    & --   & --     & --  \\
 & M11 & Act    & --    & --   & 1.44M  & --    & --   & --    & --   & --     & --  \\
\midrule
\multirow{11}{*}{\textsc{Qwen3-32B-AWQ}}
 & --  & NoMem  & --    & --   & 0      & --    & --   & --    & --   & --     & --  \\
 & M1  & Flat   & 0     & 0    & 2.29M  & 0     & 0    & 0     & 0    & 0      & 0   \\
 & M2  & Flat   & 0     & 0    & 2.26M  & 0     & 0    & 0     & 0    & 0      & 0   \\
 & M3  & Text   & 55.6K & 36   & 2.76M  & 2.81M & 148  & 0     & 0    & 2.86M  & 184 \\
 & M4  & Struct & 524K  & 199  & 286K   & 0     & 0    & 0     & 0    & 524K   & 199 \\
 & M5  & Struct & 0     & 0    & 41.7K  & 121K  & 148  & 0     & 0    & 121K   & 148 \\
 & M6  & Hier   & 45.1K & 86   & 575K   & 0     & 0    & 13.5K & 24   & 58.6K  & 110 \\
 & M7  & Refine & 245K  & 200  & 91.6K  & 0     & 0    & 0     & 0    & 245K   & 200 \\
 & M8  & Refine & 293K  & 200  & 84.8K  & 0     & 0    & 148K  & 200  & 441K   & 400 \\
 & M9  & Weight & --    & --   & 0      & --    & --   & --    & --   & --     & --  \\
 & M11 & Act    & --    & --   & 1.48M  & --    & --   & --    & --   & --     & --  \\
\midrule
\multirow{10}{*}{\textsc{Gemma-4-26B}}
 & --  & NoMem  & --    & --   & 0      & --    & --   & --    & --   & --     & --  \\
 & M1  & Flat   & 0     & 0    & 2.29M  & 0     & 0    & 0     & 0    & 0      & 0   \\
 & M2  & Flat   & 0     & 0    & 2.26M  & 0     & 0    & 0     & 0    & 0      & 0   \\
 & M3  & Text   & 52.2K & 36   & 2.26M  & 2.30M & 148  & 0     & 0    & 2.35M  & 184 \\
 & M4  & Struct & 0     & 0    & 2.16M  & 0     & 0    & 0     & 0    & 0      & 0   \\
 & M5  & Struct & 0     & 0    & 40.1K  & 177K  & 148  & 0     & 0    & 177K   & 148 \\
 & M6  & Hier   & 44.4K & 82   & 1.14M  & 0     & 0    & 12.6K & 24   & 57.0K  & 106 \\
 & M7  & Refine & 161K  & 200  & 901K   & 0     & 0    & 0     & 0    & 161K   & 200 \\
 & M8  & Refine & 204K  & 200  & 705K   & 0     & 0    & 129K  & 200  & 333K   & 400 \\
 & M11 & Act    & --    & --   & 1.30M  & --    & --   & --    & --   & --     & --  \\
\bottomrule
\end{tabular*}
\end{table*}

\clearpage

\section{Additional Related Work}
\label{app:related_work}
\paragraph{Memory-augmented LLM agents.}
The design space of agent memory has expanded rapidly along multiple substrate axes.
On the retrieval side, early systems such as Generative Agents~\citep{park2023generative} and MemGPT~\citep{packer2023memgpt} store raw observations in flat vector indices or paged text buffers and retrieve them via embedding similarity or explicit memory paging.
These designs provide a simple and general interface for long-term recall, but they often rely on shallow similarity matching and provide limited structure for multi-hop reasoning, conflict handling, or memory maintenance.
Subsequent work shifts toward more structured representations.
Mem0~\citep{chhikara2025mem0} distills conversations into atomic facts, Zep~\citep{rasmussen2025zep} and MAGMA~\citep{jiang2026magma} build temporal or multi-graph knowledge stores, and RAPTOR~\citep{sarthi2024raptor} and H-MEM~\citep{sun2026h} organize memory into multi-level hierarchies.
Amory~\citep{zhou2026amory} constructs episodic narratives from conversational fragments with offline agentic reasoning and retrieves memories through coherence-driven reasoning over the narrative structure.
These systems improve inspectability and abstraction, but they also introduce additional design choices around write-time summarization, graph construction, entity resolution, hierarchy depth, and retrieval traversal.

More recently, refinement-based and internal-memory substrates have emerged.
Sleep-time compute~\citep{lin2025sleep} uses offline computation to pre-generate useful future responses, while MemSkill~\citep{zhang2026memskill} abstracts procedural experience into reusable skill entries for self-evolving agents.
HyperSkill~\citep{xu2026hyperskill} stores subtask steps and reusable skills in a hypergraph, ranks skills by co-occurrence across retrieved trajectories, and periodically prunes low-utility nodes and merges redundant skills.
Related work on skill evolution treats agent skills as reusable procedural artifacts and improves them with execution or verifier feedback~\citep{zhang2026coevoskills,ma2026skillneo}.
On the internal-memory side, SnapKV~\citep{li2024snapkv} and EpiCache~\citep{kim2025epicache} manage KV-cache activations through attention-guided eviction or episodic clustering, while adapter-based approaches store memory in parametric form by fine-tuning low-rank adapters on the accumulated content. Our M9 builds on LoRA-FA~\citep{zhang2023lora}, a parameter-efficient fine-tuning method.
Several recent systems combine multiple substrates into broader memory operating systems or multi-agent memory architectures~\citep{li2025memos,latimer2025hindsight,wang2025mirix}.
Despite this diversity, most systems commit to a fixed substrate configuration at design time and evaluate that configuration in a narrow task regime.
Our work complements these efforts by isolating the substrate as the experimental variable and providing a controlled empirical comparison of when each substrate family is preferable.

\paragraph{Memory evaluation benchmarks.}
Existing benchmarks span both user-centric~\citep{tan2025membench,shen2026evolmem} and agent-centric~\citep{zhouwebarena,yao2025tau} memory demands, but they remain largely siloed within each regime.
On the user-centric side, LoCoMo~\citep{maharana2024evaluating} evaluates long-term conversational memory across five question types in dialogues averaging roughly 9K tokens over up to 35 sessions.
LongMemEval~\citep{wulongmemeval} scales memory evaluation to very long interaction histories and tests abilities such as information seeking, knowledge updates, abstention, and temporal reasoning.
MemoryAgentBench~\citep{hu2025evaluating} frames memory evaluation as incremental multi-turn interaction, covering accurate retrieval, long-range understanding, test-time learning, and conflict resolution.
MemoryCD~\citep{zhang2026memorycd} derives a cross-domain memory benchmark from years of real user behavior in the Amazon Review dataset and evaluates four personalization tasks across twelve domains.
JARVIS-Bench~\citep{zhang2026jarvis} builds on 28 to 44 days of real users' multimodal daily traces, augmented with synthetic trajectory expansions, to benchmark personal intelligence agents across an input ladder from no memory to lifelong memory.
Together, these benchmarks stress whether a memory system can preserve and retrieve relevant user information over long histories.

On the agent-centric side, ALFWorld~\citep{shridhar2020alfworld} requires multi-step household planning from text-based observations and admissible actions, BigCodeBench~\citep{zhuobigcodebench} evaluates code generation with compositional function calls and realistic execution-based grading, and SWE-bench~\citep{jimenez2023swe} tests repository-level bug fixing where memory of codebase structure and previous patches may be useful.
These benchmarks stress a different regime: the agent must execute precise action or code-generation decisions, and retrieved information can be harmful when it distracts from the current observation, instruction, or execution context.
This user-centric/agent-centric divide exposes fundamentally different retrieval dynamics.
User-centric tasks often benefit from returning more candidate facts, while agent-centric tasks can degrade when imprecise retrieval dilutes attention over action-critical context.
However, existing evaluations rarely test the same set of memory substrates across both regimes under controlled conditions.
Our work bridges this gap by evaluating the same substrate configurations across benchmarks from both sides of the spectrum.

\paragraph{Empirical evaluation practices in agent memory.}
The metrics reported by existing memory systems provide an incomplete picture of substrate behavior.
Nearly all systems report some form of end-task accuracy, token F1, exact match, task success, pass rate, or LLM-as-a-judge score, but the dimensions that govern deployment feasibility are often omitted.
Write cost, including the number of LLM calls and tokens consumed during memory construction, is rarely measured systematically.
Management cost, including the overhead of deduplication, conflict resolution, summarization, graph maintenance, and memory rewriting, is likewise underreported, even in systems such as A-Mem~\citep{xu2025mem} and Zep~\citep{rasmussen2025zep}, whose architectures explicitly include management operations.
Compression ratio, which quantifies how much information each stored token preserves, has also not been consistently measured across substrates.
MemAudit~\citep{ma2026memaudit} raises a complementary concern, namely that the stored memory itself is rarely audited because evaluation mostly goes through downstream behavior. It evaluates memory as a post-interaction artifact, reconstructing a hidden user state from what the agent stores, and finds that task completion and recoverable memory behave as distinct capabilities.

Among efficiency dimensions that are occasionally reported, coverage remains sparse.
The most complete prior comparison is that of Mem0~\citep{chhikara2025mem0}, which compares seven memory systems and fourteen RAG configurations head to head on LoCoMo and reports search and total latency at p50 and p95 together with the tokens retrieved per query. Its per-method accounting is read-side. The token-consumption metric counts the tokens retrieved as answer context, which underlies the reported saving of more than 90\% token cost relative to full context, and per-turn write and management calls are not reported separately. A stored-memory token budget is given for three systems and shows how large construction can be, since Zep's memory graph exceeds 600k tokens per conversation against roughly 26k tokens for the raw conversation. Zep reports latency and token cost without a cross-substrate accounting of memory construction. LightMem~\citep{fang2025lightmem} separates total from purely online test-time token usage and API calls, but does not attribute costs to the write, read, and management stages. Our metrics attribute tokens and calls to the write (E5, E6), read (E8, E9), and management (E10, E11) stages and add storage (E1) and wall-clock (E14), which gives, to our knowledge, the first stage-attributed cost accounting across substrate families.
This metric narrowness means that a substrate consuming a large number of auxiliary LLM calls can appear indistinguishable from a substrate requiring no auxiliary calls when both are evaluated only by downstream accuracy.
In addition, even the accuracy metrics that are commonly reported may be less stable than assumed.
Recent work~\citep{li2025evaluating} shows that reordering score rubrics or altering reference answer quality in LLM-as-a-judge prompts can disrupt scoring stability across judge models, yet published memory evaluations often use different judge prompts and scoring configurations.
Our work fills these gaps by measuring task performance, computational efficiency, and memory quality across all substrate configurations under a single controlled setup.

\section{Limitations}
\label{app:limitations}

We exclude MemGPT~\citep{packer2023memgpt}, Mem0~\citep{chhikara2025mem0}, and
Zep~\citep{rasmussen2025zep} because their defining mechanisms are policies outside
the single-variable frame and their auxiliary-LLM cost is one to two orders of
magnitude above the lighter substrates (Appendix~\ref{app:excluded-substrates}). M8 and
M11 are harness-compatible approximations of MemSkill and EpiCache
(Appendix~\ref{app:substrate-details}), and a shared harness can disadvantage methods that
rely on their own controller or prompts. The agent-centric benchmarks have 134 and 148
tasks and the main tables report single greedy runs, so apart from the significant
ALFWorld refinement comparison (Appendix~\ref{app:significance}) we read small gaps as
descriptive. The $P_4$ judge, \mbox{\textsc{gpt-4o-mini}}, is also the auxiliary LLM and was
checked only against three other judges from the same provider, not against human labels
(Appendix~\ref{app:judge}). Routing gains are in-sample at the regime level, and the
query-level router, tested on LoCoMo with one backbone, gains only on questions from
conversations it saw in training (Appendix~\ref{app:routing}). Finally, none of our main benchmarks is built so that later sessions depend on memory the agent wrote in earlier ones (Appendix~\ref{app:lifecycle}). The supplementary MemoryArena evaluation of such tasks covers one backbone and two domains (Appendix~\ref{app:memoryarena}).

\section{Broader Impacts}
\label{app:broader-impacts}

This work provides empirical guidance for designing memory systems for long-horizon LLM
agents. More efficient and reliable agent memory could reduce the energy use of agentic
deployments, improve consistency in long-running personal companions and coding assistants,
and give future memory research a controlled evaluation framework to build on.
Better-functioning long-term memory could also enable more persistent agentic surveillance or
more convincing persona simulations. We view these as derivative risks of foundation-model
agents in general rather than risks specific to substrate evaluation, and we release no new
model or dataset that would directly enable misuse.

\section{Declaration of LLM Usage}
\label{app:llm-usage}

LLMs were used by the authors only for routine writing assistance, such as grammar
polishing, phrasing refinement, and LaTeX formatting. They were not used to generate
research ideas, design experiments, analyze results, or produce any of the empirical
findings reported in this paper. All scientific contributions, including the harness
design, substrate implementations, metric taxonomy, and analysis, are the original
work of the authors.

%%%%%%%%%%%%%%%%%%%%%%%%%%%%%%%%%%%%%%%%%%%%%%%%%%%%%%%%%%%%

% \clearpage 
% \newpage

%\clearpage
%\input{checklist.tex}

\end{document}